\documentclass{article}
\usepackage{iclr2027_conference,times}

\usepackage[utf8]{inputenc}
\usepackage[T1]{fontenc}
\usepackage{graphicx}
\usepackage{iftex}
\ifPDFTeX\pdfminorversion=7\fi
\ifXeTeX\AtBeginDocument{}\fi
\usepackage{microtype}
\usepackage{xcolor}
\usepackage{amsmath,amssymb,amsfonts}
\usepackage{booktabs}
\usepackage{multirow}
\usepackage{algorithm}
\usepackage{algpseudocode}
\usepackage{array}
\usepackage{tabularx}
\usepackage{capt-of}
\usepackage{placeins}
\usepackage{listings}
\usepackage{upquote}
\usepackage[hidelinks]{hyperref}
\usepackage{xurl}

\lstdefinestyle{appendixstyle}{
  basicstyle=\ttfamily\small,
  breaklines=true,
  breakatwhitespace=false,
  columns=fullflexible,
  keepspaces=true,
  showstringspaces=false,
  upquote=true,
  tabsize=2,
  frame=single,
  framerule=0.5pt,
  framesep=4pt,
  rulecolor=\color{black},
  backgroundcolor=\color{white}
}

\lstnewenvironment{appendixbox}[1]
  {\lstset{style=appendixstyle,title={#1}}}
  {}

\newcolumntype{L}[1]{>{\raggedright\arraybackslash}p{#1}}

\title{Can Broad Biomedical Knowledge be\\
Contextualized into Scenario-Grounded\\
Propositions?}
\author{
\hspace*{-8pt}\begin{tabular}{@{}l@{}}
\textbf{Qingyuan Zeng}$^{1,\ast}$, \textbf{Ziyang Chen}$^{1,\ast}$, \textbf{Pengxiang Cai}$^{1}$, \textbf{Zixin Guan}$^{3}$, \textbf{Anglin Liu}$^{1}$,\\
\textbf{Min Zhou}$^{1}$, \textbf{Ziqi Gong}$^{1}$, \textbf{Xinyao Lai}$^{1}$, \textbf{Jintai Chen}$^{1,2,\dagger}$\\
$^{1}$The Hong Kong University of Science and Technology (Guangzhou)\\
$^{2}$The Hong Kong University of Science and Technology\\
$^{3}$Guangzhou University of Chinese Medicine\\
\texttt{jintaichen@hkust-gz.edu.cn}
\end{tabular}%
}

\iclrfinalcopy

\hypersetup{
  pdftitle={Can Broad Biomedical Knowledge be Contextualized into Scenario-Grounded Propositions?},
  pdfauthor={Qingyuan Zeng, Ziyang Chen, Pengxiang Cai, Zixin Guan, Anglin Liu, Min Zhou, Ziqi Gong, Xinyao Lai, Jintai Chen},
  pdfkeywords={biomedical knowledge contextualization, multi-agent systems, subgroup discovery, LINCS L1000}
}

\begin{document}

\maketitle

\fancyhead{}
\renewcommand{\headrulewidth}{0pt}

\begingroup
\renewcommand{\thefootnote}{\fnsymbol{footnote}}
\footnotetext[1]{These authors contributed equally to this work.}
\footnotetext[2]{Corresponding author.}
\endgroup

\begin{abstract}
Biomedical discovery often requires reconciling broad biomedical knowledge with specific experimental or clinical data. While background knowledge suggests useful biological mechanisms, it is typically too general to map directly onto concrete dataset variables. Conversely, data-driven patterns are often dataset-specific and lack mechanistic insight, creating a gap between abstract principles and concrete evidence. We formulate the gap as \textit{knowledge contextualization}: transforming broad biomedical knowledge into evidence-supported, scenario-grounded propositions, with the goal of allowing domain experts to efficiently inspect, replay, and validate candidate hypotheses. To achieve this, we propose SCENE, a bi-level multi-agent framework that implements knowledge contextualization as an iterative search process. The upper level translates broad knowledge into search directions and maps them onto the dataset schema. The lower level then executes these grounded directions using multi-objective optimization to find concrete propositions that balance evidential strength with sufficient data support. A feedback loop between the two levels progressively refines these search directions. We evaluate SCENE in two distinct settings: discovering patient subgroups with heterogeneous treatment benefits in clinical trial scenarios, and identifying context-specific biological responses in LINCS L1000 studies. In clinical trial scenarios, SCENE outperforms existing baselines by discovering highly specific, strongly supported subgroups. Furthermore, SCENE discovers perturbational contexts with stronger target-response matching and higher strong-response rates in LINCS L1000 studies. These results demonstrate that SCENE effectively bridges the gap between broad knowledge and scenario-specific evidence, providing traceable and inspectable hypotheses for follow-up validation.
\end{abstract}

\section{Introduction}
Biomedical discovery often requires reconciling broad biomedical knowledge with specific experimental or clinical datasets. While background knowledge suggests useful biological mechanisms, it is typically too generic to directly map onto the variables and constraints of a concrete setting. Conversely, the evidence collected in a specific dataset is rich in context but rarely explains how it connects to reusable biomedical principles~\citep{zarin2011clinicaltrials,lamb2006connectivitymap,subramanian2017nextgeneration}. This mismatch creates a practical obstacle: it becomes difficult to prioritize which data patterns deserve follow-up, because statistically strong patterns may lack biological rationale, while plausible biomedical ideas may remain too abstract to test in the available data.

Existing methods address parts of this problem but typically approach it from a single direction, failing to bridge the gap between broad biomedical knowledge and concrete data. On the knowledge side, integration methods inject external knowledge into prompts, representations, or retrieval modules~\citep{wang2024fedmeki,xie2025hypkg}. However, this approach leaves the knowledge largely static; it does not specify how broad biomedical knowledge should be grounded under the specific variables, endpoints, and validity constraints of a given dataset. On the evidence side, data-driven algorithms, such as tabular models and subgroup discovery, search for statistical patterns within a single dataset~\citep{foster2011virtual,lipkovich2011sides,athey2016recursive,athey2019generalized,zhou2024curls,yang2026learning,lamb2006connectivitymap,subramanian2017nextgeneration}. These discovered patterns are often weakly tied to prior biomedical knowledge, making them difficult to validate or reuse as traceable candidate hypotheses. Consequently, there remains a lack of an explicit conversion mechanism that translates broad knowledge into data-grounded, scenario-specific propositions.

\begin{figure}[!htb]
    \centering
    \includegraphics[width=\linewidth]{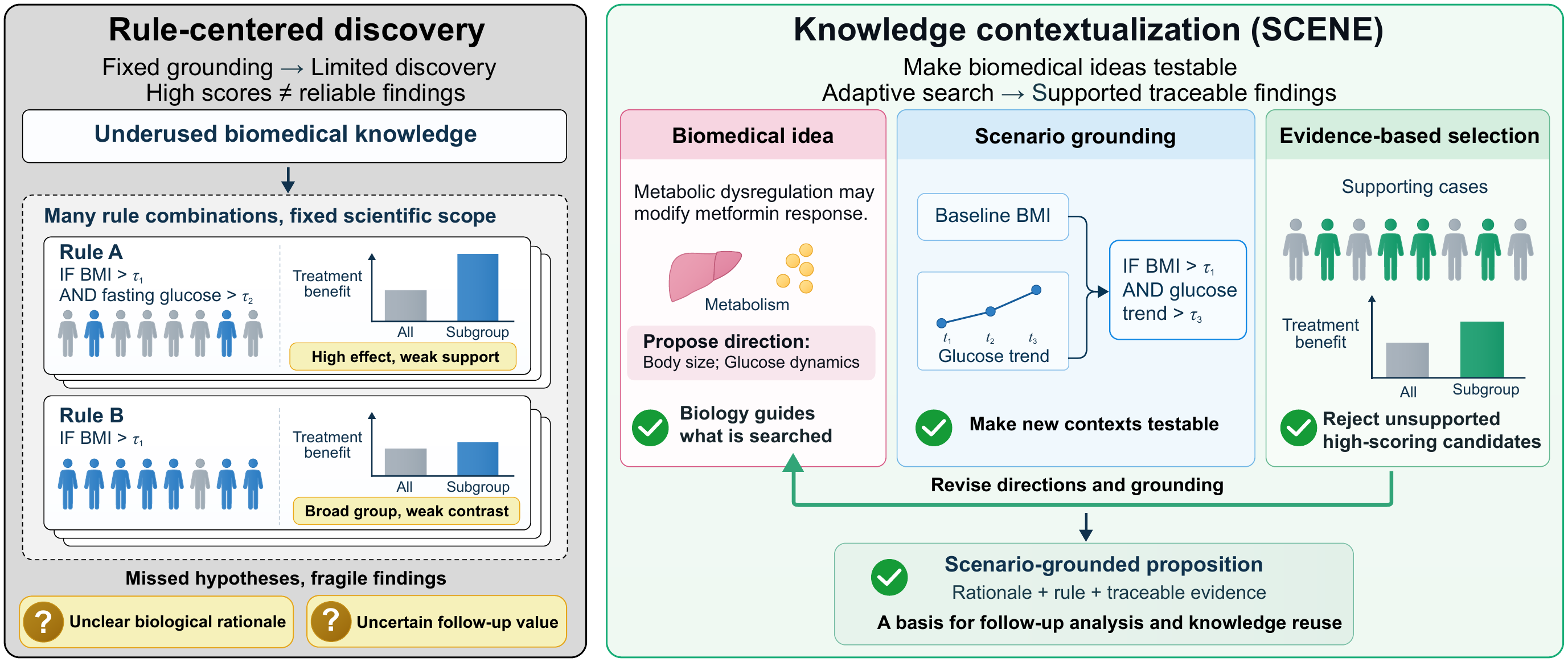}
    \caption{
From rule search to knowledge contextualization. Fixed grounding can leave relevant hypotheses unexplored, while high-scoring rules may lack adequate support. SCENE makes biomedical ideas testable, revises search directions and grounding using discovery evidence, and balances effect with support. Its scenario-grounded propositions link biomedical rationale, executable rules, and traceable evidence for follow-up validation and knowledge reuse. The clinical example is schematic. 
}
\label{fig:scenario_grounded_propositions}
\end{figure}

We therefore formulate this missing conversion step as \emph{knowledge contextualization}: transforming broad biomedical knowledge into evidence-supported, scenario-grounded propositions. The goal is to produce traceable candidate hypotheses that domain experts can efficiently inspect, replay, and validate in follow-up analysis~\citep{chandak2023primekg,moreau2013provdm}. This problem is challenging for three reasons. First, broad biomedical concepts do not directly match the concrete variables and measurements in a specific dataset. For example, a concept such as \textit{metabolic dysregulation} may need to be instantiated using measurements such as BMI, fasting glucose, or triglycerides, depending on the dataset schema. Second, the same biomedical principle requires different grounding across distinct settings; for example, it may manifest as a patient subgroup in a clinical trial, or as a context-specific cellular response in a perturbational assay~\citep{foster2011virtual,subramanian2017nextgeneration}. Third, even after mapping these concepts to concrete variables, the space of valid candidate propositions is immense, making exhaustive search computationally impractical~\citep{lipkovich2024modernhte,zhou2024curls}. Addressing these challenges requires an explicit contextualization process that can ground biomedical priors, evaluate them against available evidence, and return interpretable results; Figure~\ref{fig:scenario_grounded_propositions} contrasts rule-centered discovery with evidence-driven knowledge contextualization.

To address these challenges, we propose SCENE, a bi-level multi-agent framework that implements knowledge contextualization as an iterative search process~\citep{hong2023metagpt,wu2024autogen,zhang2025aflow}. At the upper level, SCENE translates broad biomedical knowledge into testable search directions and formulates a schema-grounding plan. At the lower level, it executes this plan using multi-objective optimization to find concrete propositions that balance evidential strength with sufficient data support~\citep{deb2002fast}. Crucially, the two levels form a bidirectional closed loop: the upper level constrains the search space based on biomedical priors, while the scenario evidence from the lower level guides the upper level to refine or pivot the search directions. By explicitly linking knowledge-guided generation with data-driven evaluation, SCENE provides a traceable mechanism to turn abstract knowledge into scenario-grounded propositions.

\begin{itemize}
    \item We formulate \emph{knowledge contextualization} as a standalone problem in biomedical discovery: transforming broad biomedical knowledge into evidence-supported, scenario-grounded propositions to produce traceable, inspectable candidate hypotheses.
    
    \item We propose SCENE, a bi-level multi-agent framework that forms a bidirectional closed loop between upper-level direction planning and lower-level multi-objective evolutionary search: the upper level guides the lower level by constraining and prioritizing its executable search space, while the lower level feeds scenario-specific evidence back to the upper level to refine subsequent directions.
    
    \item  We demonstrate SCENE's generality across two distinct biomedical settings: clinical-trial treatment-benefit subgroup discovery and LINCS L1000 perturbational context discovery. In clinical trials, SCENE outperforms existing baselines by discovering subgroups with higher risk reduction and stronger data support. In L1000 studies, it discovers perturbational contexts with stronger target-response matching and higher strong-response rates than evaluated baselines. We further show that SCENE propositions can improve downstream few-shot classification.
\end{itemize}

\section{Related Work}

\paragraph{Knowledge integration in biomedicine.}
A line of work incorporates external biomedical knowledge into machine learning systems through knowledge graphs, structured retrieval, or knowledge injection. Recent efforts such as HypKG~\citep{xie2025hypkg} contextualize biomedical knowledge graphs with patient-specific EHR information, while benchmarks such as FEDMEKI~\citep{wang2024fedmeki} study how medical knowledge can be injected into foundation models under privacy constraints. However, these approaches primarily operate at the level of representation learning or model training, aiming to improve downstream prediction or generation, rather than directly producing interpretable, scenario-grounded propositions in the current setting.

\paragraph{Single-scenario subgroup discovery and treatment effect heterogeneity.}
Another related literature studies treatment effect heterogeneity and subgroup discovery, especially in clinical trials. Classical methods such as Virtual Twins~\citep{foster2011virtual} and SIDES~\citep{lipkovich2011sides} search for treatment-responsive subgroups through predicted individual responses or recursive partitioning, while honest trees~\citep{athey2016recursive} and causal forests~\citep{athey2019generalized} extend this line through tree- or forest-based estimation of heterogeneous treatment effects. More recent approaches, such as CURLS~\citep{zhou2024curls} and Learning Subgroups with Maximum Treatment Effects without Causal Heuristics~\citep{yang2026learning} cast subgroup discovery as rule learning or structured optimization. Despite their differences, these methods share a common perspective: they treat the problem as data-only effect estimation, partitioning, or rule search within a given dataset. They neither use broad biomedical knowledge as an explicit hypothesis space, nor address how such knowledge can be converted into scenario-grounded propositions.

\paragraph{Our position.}
SCENE is not defined around subgroup discovery, treatment effect estimation, or any single downstream task. Instead, we study \emph{knowledge contextualization} itself: how broad biomedical knowledge can be proposed, grounded, tested, and refined in a concrete scenario. Under this view, subgroup rules in clinical trials are only one form of scenario-grounded proposition, alongside context-bounded findings in perturbational transcriptomic studies. SCENE uses scenario evidence to revise how knowledge is grounded and what is searched next, rather than only optimizing rules within a fixed feature set.

\section{Methodology}
\label{sec:method}

\subsection{Task and Overview}

For a biomedical scenario $s$, SCENE receives three inputs: biomedical prior knowledge $\mathcal{K}^{(s)}$, scenario evidence $\mathcal{D}^{(s)}=(X^{(s)},Y^{(s)},C^{(s)})$, and a machine-readable schema $\mathcal{S}^{(s)}$. Here $X^{(s)}$ denotes observable features, $Y^{(s)}$ denotes outcomes or response variables used for evidence evaluation, and $C^{(s)}$ denotes task context. The schema $\mathcal{S}^{(s)}$ specifies observable variables, admissible windows or operators, forbidden fields for candidate construction, and validity constraints. Biomedical prior knowledge $\mathcal{K}^{(s)}$ is treated as a finite set of pre-discovery records, such as statements derived from publications, trial records, or study resources. These records provide source-traceable starting points for search directions. Detailed record formats, manifest fields, and provenance requirements are given in Appendices~\ref{app:scenario_formalization} and~\ref{app:reporting_boundaries}.

The target output is a set of scenario-grounded propositions
\begin{equation}
\mathcal{P}^{(s)}=\{p_i\}_{i=1}^m,
\qquad
p_i=(d_i,q_i,\omega_i).
\end{equation}
Here $d_i$ is a search direction specifying the biomedical concept or relation to explore, $q_i$ is an executable rule or finding grounded in the current scenario, and $\omega_i$ is an evidence record containing effect, support, diagnostics, source links, and provenance. The grounded object $q_i$ may be a subgroup rule in clinical discovery or a cell--perturbagen--dose--time context in L1000. Its evidence record makes the proposition inspectable and replayable. Thus, the reportable unit is the entire proposition $(d_i,q_i,\omega_i)$, not a rule or finding alone.

SCENE implements knowledge contextualization as a closed-loop search process. The upper level translates prior knowledge and task context into bounded search directions and schema-grounding plans, which constrain and prioritize the executable search space. The lower level searches admissible candidate rules or findings with multi-objective optimization over effect and support, and returns scenario-specific feedback for later direction refinement. The final outputs are proposition-level artifacts that combine a direction, an executable grounded object, and an evidence record.

\begin{figure}[!htb]
    \centering
    \includegraphics[width=\linewidth]{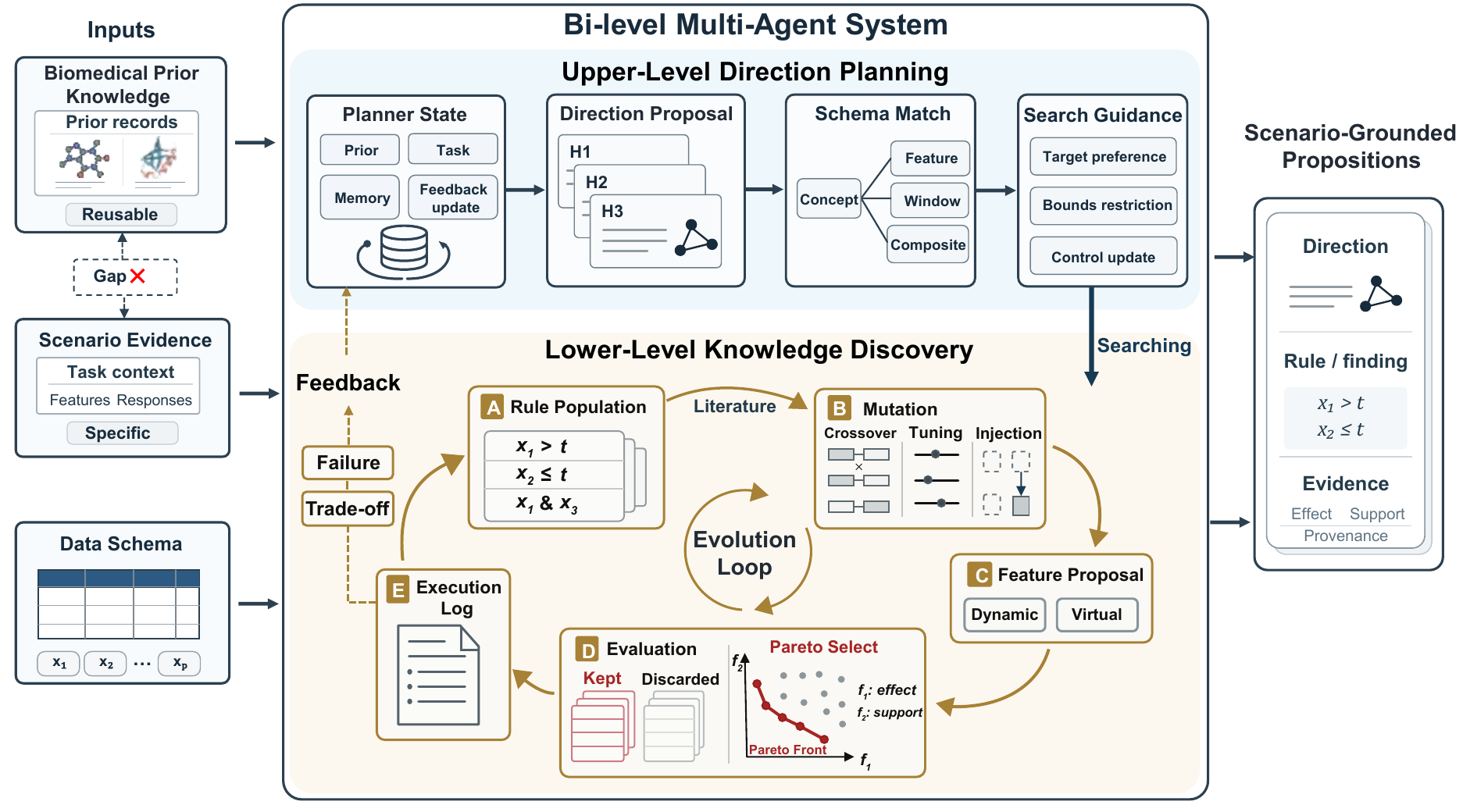}
    \caption{SCENE (Scenario-Contextualized Evidence and Knowledge Engine) artifact flow. Biomedical prior knowledge, scenario evidence, and a data schema define the discovery scenario. The upper level converts broad knowledge into bounded search directions and schema-grounding plans that constrain and prioritize the lower-level executable search space. The lower level searches candidate rules or findings with multi-objective effect/support optimization and returns scenario-specific feedback to refine later directions. Exported outputs are scenario-grounded propositions, each pairing a direction, an executable rule or finding, and an evidence record with provenance.}
    \label{fig:scene_overview}
\end{figure}

Figure~\ref{fig:scene_overview} summarizes the operational flow. Given an active direction at round $t$, the schema-grounding plan $\mathcal{G}_t$ induces an admissible candidate family $\mathcal{Q}_t$. A scenario adapter $\mathcal{A}^{(s)}$ specifies the executable candidate language, validity checks, support and effect functions, and reporting boundary. Details of the candidate grammar, grounding operators, reporting checks, and scenario-specific estimators are deferred to Appendices~\ref{app:grounding_operators}, \ref{app:reporting_boundaries}, and~\ref{app:scenario_estimators}. The round-level protocol is shown in Algorithm~\ref{alg:scene_protocol}.

\begin{algorithm}[t]
\caption{SCENE round-level search protocol}
\label{alg:scene_protocol}
\begin{algorithmic}[1]
\Require prior records $\mathcal{K}^{(s)}$, scenario evidence $\mathcal{D}^{(s)}$, schema $\mathcal{S}^{(s)}$, adapter $\mathcal{A}^{(s)}$, round budget $T$
\Ensure reportable propositions $\mathcal{P}^{(s)}$
\State initialize compact state $m_0$, feedback $f_0$, controls $\eta_1^0$, and $\mathcal{P}^{(s)}\gets\emptyset$
\For{$t \gets 1$ to $T$}
    \State $(d_t,\eta_t) \gets \Call{Plan}{\mathcal{K}^{(s)},C^{(s)},\mathcal{S}^{(s)},m_{t-1},f_{t-1},\eta_t^0}$ \Comment{propose a bounded search direction}
    \State $\mathcal{G}_t \gets \Call{Ground}{d_t,\mathcal{S}^{(s)},\eta_t}$ \Comment{map the direction to schema-visible objects}
    \State $(\mathcal{G}_t,\eta_t) \gets \Call{Validate}{\mathcal{G}_t,\eta_t,\mathcal{A}^{(s)}}$ \Comment{remove invalid or leakage-risk objects}
    \State $\mathcal{C}_{t,0} \gets \Call{Seed}{\mathcal{G}_t,\eta_t}$ \Comment{initialize candidate rules or findings}
    \State $(\mathcal{F}_t,\lambda_t) \gets \Call{Evolve}{d_t,\mathcal{G}_t,\mathcal{C}_{t,0},\mathcal{D}^{(s)},\mathcal{A}^{(s)},\eta_t}$ \Comment{search and retain a Pareto frontier}
    \State $(f_t,m_t,\bar{\eta}_{t+1}) \gets \Call{UpdateFeedback}{\mathcal{F}_t,\lambda_t,d_t,m_{t-1}}$ \Comment{update next-round search using discovery-visible logs}
    \State $\mathcal{P}^{(s)} \gets \mathcal{P}^{(s)} \cup \Call{Report}{\mathcal{F}_t,d_t,\mathcal{D}^{(s)},\mathcal{A}^{(s)}}$ \Comment{export propositions passing reporting checks}
    \State $\eta_{t+1}^0 \gets \bar{\eta}_{t+1}$
\EndFor
\State \Return $\mathcal{P}^{(s)}$
\end{algorithmic}
\end{algorithm}

\subsection{Upper-Level Direction Planning}

The upper level decides what should be searched next. It receives prior records, task context, schema summaries, and discovery-visible feedback from previous rounds. Its output is not a biomedical claim, but a bounded search direction together with search guidance:
\begin{equation}
(d_t,\eta_t)=
\mathrm{Plan}_{\theta}(\mathcal{K}^{(s)},C^{(s)},\mathcal{S}^{(s)},m_{t-1},f_{t-1},\eta_t^0),
\qquad
\eta_t=(\pi_t,H_t,\rho_t,\beta_t).
\end{equation}
Here $\pi_t$ is the operator mixture, $H_t$ is the generation budget, $\rho_t$ is the support floor, and $\beta_t$ contains bounded preferences such as target feature families, admissible ranges, avoid lists, or virtual/dynamic feature toggles. The symbol $\theta$ denotes fixed role instructions, decoding settings, controller bounds, and replay policy, all specified before the discovery run.

Operationally, \textsc{Plan} performs three functions. First, it selects or revises a search direction from prior knowledge, task context, and prior feedback. Second, it grounds the direction to schema-visible feature families, windows, or admissible composites. Third, it converts the grounded direction into search guidance for the lower level. All role outputs, including directions and grounding plans, are treated as structured proposals and must pass deterministic validation before entering executable search. Detailed role contracts and controller policies are given in Appendices~\ref{app:agent_roles} and~\ref{app:role_prompt_contracts}.

\subsection{Lower-Level Knowledge Discovery}

Conditioned on the active direction and its grounding plan, the lower level searches over an admissible candidate family $\mathcal{Q}_t$. Candidates are executable scenario-specific objects: subgroup rules in clinical settings and context-bounded findings in L1000 settings. The search starts from grounded seeds and diversity-oriented admissible candidates, then applies variation operators such as mutation, crossover, tuning, and injection. When raw observables are insufficient to instantiate a direction, virtual or dynamic features may be proposed, but they enter the search space only after deterministic construction, provenance recording, and validity checks.

Each candidate $q\in\mathcal{Q}_t$ is evaluated by the scenario adapter, which returns an effect objective $J_{\mathrm{eff}}^{(s)}(q)$, a support objective $J_{\mathrm{sup}}^{(s)}(q)$, and discovery-visible diagnostics $V^{(s)}(q)$. The effect objective measures the configured scenario contrast, such as a treatment-benefit contrast in clinical subgroup discovery or a target-response contrast in L1000 discovery. The support objective prevents the search from favoring highly specific but poorly supported fragments. These objectives define discovery-side ranking under the frozen split and manifest. Pareto dominance and frontier membership use only effect and support:
\begin{equation}
\mathcal{F}_t =
\left\{
q\in\mathcal{Q}_t :
\nexists q'\in\mathcal{Q}_t
\ \text{such that}
\begin{array}{l}
J_{\mathrm{eff}}^{(s)}(q')\ge J_{\mathrm{eff}}^{(s)}(q),\\
J_{\mathrm{sup}}^{(s)}(q')\ge J_{\mathrm{sup}}^{(s)}(q),\\
\text{and at least one inequality is strict}
\end{array}
\right\}.
\end{equation}
The retained frontier is both a candidate set and a feedback object. A coherent frontier indicates that the current direction admits repeated grounded instantiations, whereas empty, low-support, invalid, or redundant frontiers expose failure modes. Diagnostics can annotate instability or break exact ties after Pareto ranking and diversity filtering, but they cannot create or remove Pareto dominance. Evolutionary operators and default hyperparameters are detailed in Appendix~\ref{app:evolution_details}; executable grounding operators are described in Appendix~\ref{app:grounding_operators}.

\subsection{Closed-Loop Interaction and Reporting Boundary}

The upper and lower levels interact through validated artifacts and frontier logs rather than unconstrained dialogue. At each round, the upper level sends a direction, a grounding plan, and bounded search guidance to shape the lower-level search; the lower level returns a retained frontier $\mathcal{F}_t$ and an execution log $\lambda_t$ for the next planning step. The feedback update summarizes failure modes, support--effect trade-offs, alignment gaps, diversity, and diagnostic contrast:
\begin{equation}
(f_t,m_t,\bar{\eta}_{t+1})
=\Phi_{\mathrm{fb}}(\mathcal{F}_t,\lambda_t,d_t,m_{t-1}).
\end{equation}
These signals help later rounds preserve, narrow, broaden, or replace search directions.

Export is stricter than feedback. A candidate may influence later search if it is discovery-visible, but it becomes a reported proposition only if it lies on the retained frontier and passes validity, support, provenance, leakage, and diagnostic checks. Final held-out or test evidence is used only for post-discovery audit and is never fed back into the same run to revise directions, rules, findings, or search controls. SCENE therefore links validated knowledge grounding and effect/support search with evidence-driven revision, while preserving the direction, executable finding, and provenance in each reported proposition. Detailed split-lineage and reporting schemas are given in Appendix~\ref{app:reporting_boundaries}.

\FloatBarrier
\section{Experiments}
\label{sec:experiments}

\subsection{Experimental Setup}

\begin{table}[!htb]
\centering
\caption{Held-out clinical subgroup discovery. Frozen Best-1 rules are replayed over 50 paired splits. Task columns report held-out bad-outcome ARR, B/P denote breast-cancer/PCOS tasks and BL/TR denote baseline/trajectory views. P-ARR is population-adjusted ARR, U-Supp. is usable support, and Best-1 DC is train-to-holdout direction consistency. Higher is better for all displayed metrics.}
\label{tab:clinical_best1_benefit_revised}
\scriptsize
\setlength{\tabcolsep}{2.0pt}
\renewcommand{\arraystretch}{1.08}
\resizebox{\textwidth}{!}{%
\begin{tabular}{@{}lccccccccc@{}}
\toprule
\textbf{Method} &
\textbf{B-D-BL} &
\textbf{B-D-TR} &
\textbf{P-M-BL} &
\textbf{P-M-TR} &
\textbf{P-A-FPG} &
\textbf{P-A-AUC} &
\textbf{P-ARR $\uparrow$} &
\textbf{U-Supp. $\uparrow$} &
\textbf{Best-1 DC $\uparrow$} \\
\midrule
Virtual Twins
& 0.023
& -0.027
& 0.036
& 0.197
& 0.078
& 0.004
& 0.014
& 58.7
& 51.5 \\
SIDES
& 0.007
& 0.008
& 0.195
& -0.005
& 0.006
& -0.058
& 0.010
& 26.7
& \underline{74.3} \\
Causal Forest
& 0.066
& 0.005
& 0.104
& 0.035
& 0.060
& 0.130
& 0.016
& 56.3
& 51.4 \\
Honest Causal Tree
& 0.057
& \underline{0.037}
& 0.232
& 0.042
& 0.053
& 0.087
& 0.032
& \underline{70.0}
& 71.7 \\
RuAG  
& 0.004 
& 0.002 
& 0.117 
& 0.113 
& \underline{0.127}
& 0.008 
& 0.016 
& 33.3 
& 49.0 \\
CURLS
& 0.015
& 0.003
& 0.025
& \underline{0.210}
& 0.118
& 0.103
& 0.010
& 26.3
& 45.7 \\
MaxTE Subgroups
& \underline{0.076}
& 0.012
& \underline{0.234}
& 0.121
& 0.117
& \underline{0.132}
& \underline{0.054}
& 38.3
& 68.6 \\
SCENE (ours)
& \textbf{0.086}
& \textbf{0.095}
& \textbf{0.441}
& \textbf{0.439}
& \textbf{0.269}
& \textbf{0.270}
& \textbf{0.093}
& \textbf{84.7}
& \textbf{100.0} \\
\bottomrule
\end{tabular}%
}
\end{table}

\begin{figure}[!htb]
\centering
\begingroup
\newsavebox{\sceneLTableBody}
\sbox{\sceneLTableBody}{%
\footnotesize
\setlength{\tabcolsep}{2pt}%
\renewcommand{\arraystretch}{1.27}%
\resizebox{0.555\textwidth}{!}{\begin{tabular}{@{}lccccc@{}}
\toprule
\textbf{Method} & \textbf{Conn. $\uparrow$} & \textbf{Strong-Pos. $\uparrow$} & \textbf{AUPRC $\uparrow$} & \textbf{U-Supp. $\uparrow$} & \textbf{Gap $\downarrow$} \\
\midrule
MaxTE-Context & 0.104 & 18.5 & \textbf{0.334} & \textbf{78.8} & 0.228 \\
One-shot LLM & 0.107 & 24.4 & 0.253 & 38.3 & 0.129 \\
\midrule
Minimal & 0.166 & 27.6 & 0.260 & 33.5 & 0.111 \\
w/o Dynamic & 0.171 & 28.5 & 0.273 & 34.2 & 0.068 \\
w/o Pareto & 0.184 & 18.8 & 0.273 & 30.0 & 0.081 \\
w/o Upper & 0.177 & 35.1 & 0.271 & 29.5 & 0.106 \\
w/o Virtual & 0.170 & 24.2 & 0.275 & 34.3 & 0.087 \\
SCENE (ours) & \textbf{0.188} & \textbf{41.5} & 0.268 & 31.4 & \textbf{0.063} \\
\bottomrule
\end{tabular}}%
}
\noindent\begin{minipage}[t]{0.425\textwidth}
\vspace{0pt}
\centering
\includegraphics[height=\dimexpr\ht\sceneLTableBody+\dp\sceneLTableBody\relax,trim={1bp 0.7bp 0.8bp 0.7bp},clip]{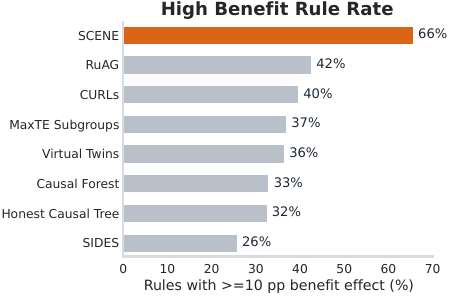}
\end{minipage}\hfill
\begin{minipage}[t]{0.555\textwidth}
\vspace{0pt}
\centering
\usebox{\sceneLTableBody}
\end{minipage}
\par\vspace{6pt}
\noindent\begin{minipage}[t]{0.425\textwidth}
\vspace{0pt}
\setlength{\abovecaptionskip}{0pt}
\setlength{\belowcaptionskip}{0pt}
\captionof{figure}{Fraction of exported rules with treatment-benefit effect at least 10 percentage points.}
\label{fig:exp-high-benefit}
\end{minipage}\hfill
\begin{minipage}[t]{0.555\textwidth}
\vspace{0pt}
\setlength{\abovecaptionskip}{0pt}
\setlength{\belowcaptionskip}{0pt}
\captionof{table}{L1000 baselines and component ablations. Conn. is connectivity margin, Strong-Pos. the strong-response rate, U-Supp. the usable-rule fraction (\%), and Gap the normalized train--holdout gap.}
\label{tab:l1000_ablation}
\end{minipage}
\endgroup
\end{figure}

\begin{figure}[!htb]
\centering
\includegraphics[height=100pt]{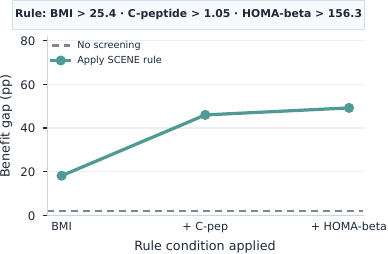}\hfill
\includegraphics[height=100pt]{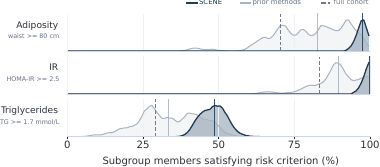}
\caption{Concrete clinical meaning of a SCENE subgroup rule.
(Left) The displayed SCENE rule increases the favorable outcome treatment--control gap.
(Right) Rule-level clinical-risk audit. Gray densities summarize valid top-ranked rules from prior methods replayed on the same cohort; blue densities summarize sensitivity-only local threshold perturbations around the SCENE rule.}
\label{fig:clinical-contextualized-case}
\end{figure}

We evaluate SCENE on two discovery settings: clinical subgroup discovery and
perturbational response discovery, and one downstream knowledge-reuse setting. The clinical benchmark uses patient-level tables linked to two registered trials, with NCT identifiers used only for registry and protocol linkage: 
NCT00174655 from Project Data Sphere and NCT02491333 from the Dryad release~\citep{projectdatasphere_access,green2015projectdatasphere,wen2021dryad,wen2022acupuncture}. We selected the trials by feasibility and coverage criteria: randomized arms, patient-level covariates, interpretable endpoints, sufficient support for repeated subgroup replay, and complementary biomedical settings. We compare against subgroup and heterogeneous treatment-effect baselines~\citep{foster2011virtual,lipkovich2011sides,athey2016recursive,athey2019generalized,zhou2024curls,yang2026learning}. We adapt \emph{RuAG}~\citep{zhang2025ruag} as an LLM-assisted rule-search baseline, combining LLM-based feature selection with Monte Carlo tree search for clinical subgroups. Across 50 paired splits, each run searches the training split, exports one prioritized rule, and replays it on held-out patients for six task frames: B-D-BL, B-D-TR, P-M-BL, P-M-TR, P-A-FPG, P-A-AUC (Appendix~\ref{app:clinical-task-definitions}). The primary clinical metric is held-out bad-outcome ARR (absolute risk
reduction, larger is better). P-ARR records population-adjusted benefit, U-Supp. records usable support, and Best-1 DC records train-to-holdout benefit-direction preservation over committed Best-1 rules, definitions are in Appendix~\ref{app:table1-implementation}. In the perturbational benchmark, based on CMap/LINCS L1000 perturbational signatures~\citep{lamb2006connectivitymap,subramanian2017nextgeneration,keenan2018lincs,koleti2018lincsportal}, each episode fixes a target mechanism and performs discovery. We report Conn. as the held-out connectivity margin between rule-covered signatures and background, Strong-Pos. as the strong response rate, AUPRC~\citep{davis2006prroc,saito2015prroc} as target mechanism retrieval, U-Supp. as usable held-out support, and Gap as the normalized train--holdout connectivity gap. \emph{MaxTE-Context} adapts MaxTE~\citep{yang2026learning} to mine L1000 context rules without biomedical priors; \emph{One-shot LLM} propose rules from priors and the schema without iterative refinement; Appendix~\ref{app:table2-implementation} gives the replay contract. Finally, the reuse setting compares downstream few-shot classification with C0 (few-shot row serialization), C1 (C0 plus generic knowledge context), and C2 (C1 plus SCENE knowledge cards), using final-test AUPRC, MacroF1~\citep{sokolova2009systematic}, and accuracy to measure ranking, class-balanced F1, and correctness. Held-out outcomes and connectivity are never used to select rules or revise prompts, downstream test labels are never used to choose SCENE cards.

\subsection{Clinical Trial Subgroup Discovery}

SCENE has the highest held-out ARR point estimate in all six frames (Table~\ref{tab:clinical_best1_benefit_revised}): on baseline static frames (B-D-BL, P-M-BL), it outperforms existing SOTA methods (0.086 vs.\ 0.076; 0.441 vs.\ 0.234) and retains its lead on trajectory-enhanced dynamic frames (B-D-TR, P-M-TR, P-A-FPG, P-A-AUC), indicating that the advantage is not tied to one endpoint or table view. SCENE also outperforms existing SOTA methods in aggregate diagnostics : with P-ARR of 0.093 and usable support of 84.7\%, versus baseline maxima of 0.054 and 70.0\%. Compared with the LLM-assisted \emph{RuAG} baseline, SCENE achieves higher ARR on all six frames, alongside higher P-ARR (0.093 vs.\ 0.016), usable support (84.7\% vs.\ 33.3\%), and direction consistency (100.0\% vs.\ 49.0\%). Thus, SCENE handles diverse clinical tables while discovering support-aware propositions.

\subsection{L1000 Perturbational Context Discovery} 

Table~\ref{tab:l1000_ablation} evaluates SCENE in the L1000 mechanism discovery setting by comparing \emph{MaxTE-Context}, \emph{One-shot LLM} and using component ablations to examine how different modules affect perturbational context discovery. SCENE achieves stronger connectivity and response concentration with a smaller train--holdout gap than both controls. Removing Pareto selection (w/o Pareto) primarily collapses the high-response tail. Removing upper-level planning (w/o Upper) produces a different failure mode: the strong positive rate falls and the gap increases, suggesting weaker transfer from discovery to the holdout set. Furthermore, removing virtual features (w/o Virtual) trades effect strength for coverage. These results indicate that the components are complementary rather than redundant. Specifically, upper-level direction planning stabilizes discovery across the boundary between training and holdout sets, Pareto selection preserves strong holdout responders, and virtual or dynamic grounding expands the search space without replacing the need for evidence-based frontier selection.

\subsection{Contextual Knowledge Reuse}

\begin{figure}[!htb]
\centering
\begin{minipage}[c][101pt][c]{0.38\textwidth}
\centering
\includegraphics[width=\linewidth]{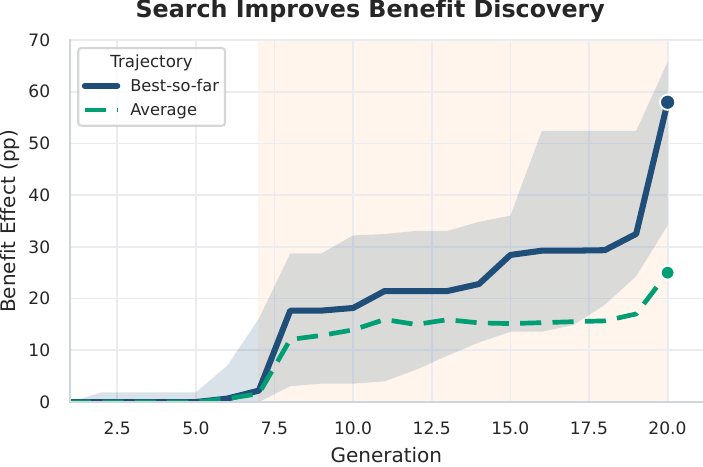}
\end{minipage}\hfill
\begin{minipage}[c][101pt][c]{0.60\textwidth}
\centering
\footnotesize
\setlength{\tabcolsep}{0pt}
\renewcommand{\arraystretch}{1.16}
\begin{tabular*}{\linewidth}{@{\extracolsep{\fill}}llccc@{}}
\toprule
\textbf{Scenario} & \textbf{Mode} & \textbf{AUPRC} & \textbf{MacroF1} & \textbf{ACC} \\
\midrule
\multirow{3}{*}{\textbf{Clinical Trial}}
& C0 & 0.5877 & 0.4408 & 0.4898 \\
& C1 & 0.6140 & 0.4745 & 0.5510 \\
& C2 & \textbf{0.6747} & \textbf{0.5291} & \textbf{0.6939} \\
\midrule
\multirow{3}{*}{\textbf{L1000}}
& C0 & 0.3582 & 0.5536 & 0.5600 \\
& C1 & 0.3725 & 0.5914 & 0.6100 \\
& C2 & \textbf{0.4113} & \textbf{0.6528} & \textbf{0.7200} \\
\bottomrule
\end{tabular*}
\end{minipage}
\par\vspace{6pt}
\noindent\begin{minipage}[t]{0.38\textwidth}
\vspace{0pt}
\setlength{\abovecaptionskip}{0pt}
\setlength{\belowcaptionskip}{0pt}
\captionof{figure}{Discovery-side SCENE search trajectory showing median best-so-far and mean candidate benefit across generations.}
\label{fig:exp-search-trajectory}
\end{minipage}\hfill
\begin{minipage}[t]{0.60\textwidth}
\vspace{0pt}
\setlength{\abovecaptionskip}{0pt}
\setlength{\belowcaptionskip}{0pt}
\captionof{table}{Few-shot reuse of SCENE propositions. C0: row-serialized few-shot baseline; C1: C0 + generic knowledge context; C2: C1 + validation-selected, row-conditioned SCENE cards. Metrics are means over 10 exemplar seeds.}
\label{tab:scenario_performance}
\end{minipage}
\end{figure}

Table~\ref{tab:scenario_performance} evaluates whether propositions discovered by SCENE can be reused as auxiliary knowledge for downstream few-shot tabular classification. We convert exported discovery propositions into a fixed pool of compact knowledge cards, select cards using downstream validation labels, and inject them into in-context prediction. Row serialization, exemplar budget, data split, and LLM decoding remain fixed across conditions. The comparison includes ordinary few-shot row serialization (C0), C0 plus generic scene context (C1), and C1 plus row-conditioned SCENE knowledge cards (C2). In the clinical scene, the downstream label is a row-level proxy for treatment-benefit evidence under the task endpoint; in the L1000 scene, the downstream label is the target-mechanism positive class defined for the selected perturbational task. C2 is best on all final test metrics in both scenarios, improving clinical accuracy from 0.5510 to 0.6939 and L1000 accuracy from 0.6100 to 0.7200 over C1. These results suggest that SCENE propositions provide reusable scenario-conditioned information beyond generic task descriptions, while serving as downstream utility evidence rather than independent biomedical validation. Seed-level summaries and prompt artifacts are audited in Appendix~\ref{app:table3-implementation}.

\begin{figure}[!htb]
\centering
\includegraphics[width=\linewidth]{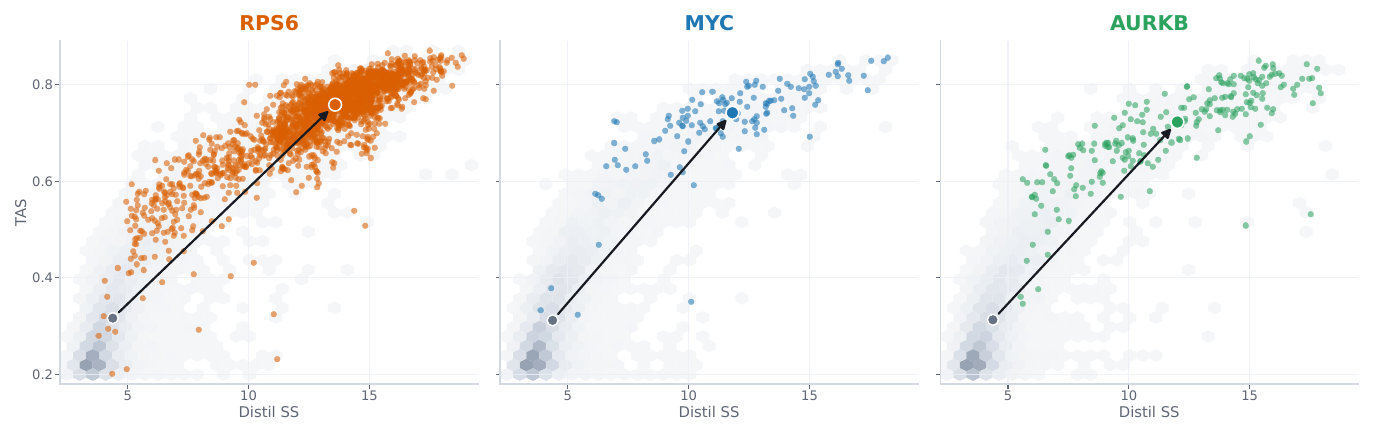}
\caption{Held-out L1000 rule replay on 3 target mechanisms. Gray hexbins summarize the held-out background, colored points denote signatures satisfying the selected rule, and arrows indicate the median shift from the background to rule-covered signatures.}
\label{fig:exp-l1000-replay}
\end{figure}

\subsection{Search, Replay, and Efficiency Diagnostics} 

\begin{figure}[!htb]
\centering
\begingroup
\setlength{\tabcolsep}{2pt}
\renewcommand{\arraystretch}{1.35}
\begin{minipage}[c][85pt][c]{0.40\textwidth}
\centering
  \includegraphics[width=\linewidth]{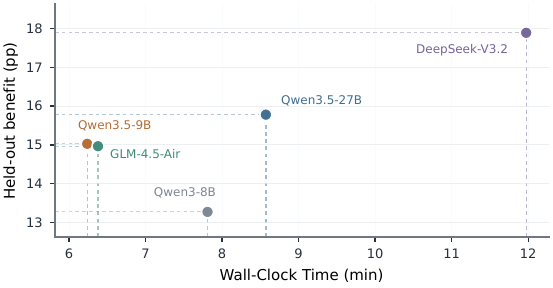}
\end{minipage}\hfill
\begin{minipage}[c][85pt][c]{0.58\textwidth}
\centering
  \resizebox{\linewidth}{!}{\begin{tabular}{@{}lcc@{}}
  \toprule
  Backend & Avg. run (min) & Avg. gen. (s) \\
  \midrule
  Qwen3.5-9B~\citep{qwen35_9b_hf} & \textbf{6.2} & \textbf{19.8} \\
  GLM-4.5-Air~\citep{zai2025glm45air} & 6.4 & 26.9 \\
  Qwen3-8B~\citep{qwen2025qwen3} & 7.8 & 25.8 \\
  Qwen3.5-27B~\citep{qwen35_27b_hf} & 8.6 & 31.2 \\
  DeepSeek-V3.2~\citep{deepseek2025v32} & 12.0 & 43.7 \\
  \bottomrule
  \end{tabular}}
\end{minipage}
\endgroup
\caption{Benefit--runtime trade-off and runtime cost across LLM backends. (Left) Each point is one third-party API backend under the same discovery protocol. (Right) Avg. run is mean wall-clock time per discovery run; Avg. gen. is mean wall-clock time per evolutionary generation within a run.}
\label{fig:model-runtime-benefit-cost}
\end{figure}

To complement the aggregate results, we further examine representative propositions, replay behavior, and efficiency diagnostics. Figure~\ref{fig:clinical-contextualized-case} shows that a metabolic P-A-AUC rule raises the favorable treatment--control gap and remains enriched for adiposity, insulin resistance, and elevated triglycerides relative to the rule distributions induced by prior methods in Table~\ref{tab:clinical_best1_benefit_revised}, suggesting a stable, task-relevant endocrine--metabolic phenotype rather than a clinically unstructured output. Figure~\ref{fig:exp-l1000-replay} replays exported rules for RPS6, MYC, and AURKB. Profiles consistently shift toward higher Distil SS and TAS (L1000 signature-strength and transcriptional-activity diagnostics~\citep{subramanian2017nextgeneration}), indicating that SCENE identifies coherent response contexts rather than isolated score gains. Figure~\ref{fig:exp-high-benefit} shows that $66\%$ of SCENE's reported clinical rules exceed a 10-percentage-point benefit threshold. Figure~\ref{fig:exp-search-trajectory} shows that both the median best-so-far and the average discovery-side benefit among candidates improve across generations. Together with Table~\ref{tab:scenario_performance}, these analyses support the view that SCENE outputs compact, scenario-grounded propositions that remain reusable outside the original discovery run. Figure~\ref{fig:model-runtime-benefit-cost} compares different API backends on the P-M-TR task, Qwen3.5-9B provides the most favorable runtime--benefit trade-off.

\subsection{Further Analysis}

Appendix~\ref{app:additional-experiments} provides post-hoc audits beyond aggregate benchmark scores. 
Appendix~\ref{app:s1-clinical-geometry} analyzes clinical benefit--support--variability geometry, showing that SCENE remains high-benefit, high-support, and low-variability. 
Appendix~\ref{app:s8-clinical-contextualization-advantage} compares reported rules on predefined clinical axes, showing stronger alignment with task-relevant phenotypes than prior-method rule. 
Appendix~\ref{app:l1000-proposition-inspectability} audits L1000 inspectability, showing improved response-quality diagnostics. 
Appendix~\ref{app:s2-holdout-reliability} evaluates replay reliability, showing that benefit directions are largely preserved and magnitude stability improves. 
Appendix~\ref{app:s3-rule-semantics} analyzes rule composition and provenance, showing that SCENE combines interpretable feature families rather than opaque rule strings. 
Appendix~\ref{app:s4-l1000-ablation} expands the L1000 ablation, showing complementary contributions from different components. 
Appendix~\ref{app:s5-feature-lifecycle} shows that generated features are validated before entering reported propositions. 
Appendix~\ref{app:s6-grounding-trace} audits grounding traces, showing a visible path from broad knowledge to executable propositions. It also examines planner consistency and report quality. Appendix~\ref{app:additional-controls} reports knowledge-guided controls and prior sensitivity. Together, these analyses support the view that SCENE outputs robust, inspectable, and provenance-aware scenario-grounded propositions rather than isolated high-scoring rules.



\FloatBarrier
\section{Conclusion}

In this work, we studied knowledge contextualization as a biomedical discovery problem: converting broad prior knowledge into evidence-supported, scenario-grounded propositions. We proposed SCENE, a bi-level multi-agent framework that couples direction planning, schema grounding, and multi-objective evidence search to produce traceable propositions rather than isolated data patterns. Compared with the evaluated baselines, SCENE achieves higher held-out ARR and aggregate usable-support rates in clinical subgroup discovery, higher connectivity margins and strong-response rates in LINCS L1000, and further improves downstream few-shot classification. These results suggest that broad biomedical knowledge can be operationalized as auditable, scenario-grounded hypotheses for follow-up validation.

\clearpage
\subsection*{AI use statement}
Generative AI tools were used to improve the presentation of author-written text, including grammar, phrasing, clarity, and readability. This writing assistance was limited to language polishing and did not alter the scientific content of the manuscript. Generative AI tools also assisted with literature search and the identification of relevant prior work. LLM use within SCENE and the experiments is described in Section~\ref{sec:method} and Appendix~\ref{app:method_details}, with two AI evaluators used for the exploratory report-quality assessment in Appendix~\ref{app:s6-grounding-trace}. All AI-assisted edits were reviewed and, where appropriate, revised by the authors.

\subsection*{Reproducibility statement}

To support reproducibility, we provide the concrete settings used in all reported experiments, including the scenario-specific data schemas, LLM role configurations, multi-objective evolutionary hyperparameters, threshold grids, and validation split boundaries. \textbf{The source code and processed datasets are included in the supplemental material submitted with this paper.} We will also release the code publicly upon acceptance. In addition, we give explicit definitions of the operational boundaries used by SCENE, such as the role instruction contracts, deterministic reporting predicates, and scenario-specific estimators, so that the bi-level knowledge contextualization procedure can be reproduced as faithfully as possible.

\bibliography{ref}
\bibliographystyle{iclr2027_conference}


\clearpage
\appendix

\section*{Overview}
In this appendix, we provide additional details and supplementary analyses for SCENE. The appendix is organized as follows. Appendix~\ref{app:method_details} presents extended methodological details, including scenario formalization, explicit LLM role contracts, the evolutionary grounding procedure, comprehensive experimental setups, and detailed implementation protocols for all main tables.  Appendix~\ref{app:additional-experiments} provides supplementary post-hoc audits (e.g., clinical effect-support geometry, L1000 inspectability, and generated feature lifecycles). Finally, Appendix~\ref{app:limitations} and Appendix~\ref{app:broader-impacts} discuss limitations and broader impacts, respectively.

\FloatBarrier
\section{Additional Method Details}
\label{app:method_details}

\subsection{Notation and Scenario Formalization}
\label{app:scenario_formalization}

This appendix expands the compressed methodological description in the main text. The main text defines SCENE at the level of the discovery protocol, while this section makes explicit the formal objects that are passed through the protocol. SCENE reports scenario-grounded, hypothesis-generating propositions with provenance and diagnostics; it does not report confirmed clinical recommendations, externally validated mechanisms, or claims certified by a language model.

For a biomedical scenario $s$, the scenario evidence is written as
\begin{equation}
\mathcal{D}^{(s)}=(X^{(s)},Y^{(s)},C^{(s)}),
\end{equation}
where $X^{(s)}$ denotes observable features, $Y^{(s)}$ denotes outcomes or response variables used for evidence evaluation, and $C^{(s)}$ denotes task context. Biomedical prior knowledge is represented as a finite manifest of source-documented records available before discovery:
\begin{equation}
\mathcal{K}^{(s)}=\{\kappa_a\}_{a=1}^{A_s},
\qquad
\kappa_a=(\mathrm{id}_a,u_a,\sigma_a,\chi_a,\psi_a).
\end{equation}
Here $A_s$ is the number of knowledge records, $\mathrm{id}_a$ is a stable record identifier, $u_a$ is a compact knowledge statement, $\sigma_a$ records the source descriptor, $\chi_a$ records the scenario scope, and $\psi_a$ records the transform that made the record available to SCENE. These records provide source-traceable search cues, but they are not treated as conclusions about the current dataset.

The exported output is contextualized knowledge, represented as a set of scenario-grounded propositions:
\begin{equation}
\mathcal{P}^{(s)}=\{p_i\}_{i=1}^{m},
\qquad
p_i=(d_i,q_i,\omega_i).
\end{equation}
Here $d_i$ is a bounded search direction, $q_i$ is an executable grounded rule or finding in the current scenario, and $\omega_i$ is an evidence record containing effect, support, diagnostics, source links, split information, and provenance. The reportable unit is therefore the entire proposition $p_i$, not a rule or finding alone. This convention also defines the output boundary used by Figure~\ref{fig:scene_overview}: grounded rules or findings and evidence records are components of one scenario-grounded proposition rather than two independent outputs.

\begin{table}[!htb]
\caption{Notation used by SCENE.}
\label{tab:notation}
\centering
\small
\setlength{\tabcolsep}{3pt}
\begin{tabularx}{\linewidth}{@{}L{2.6cm}X@{}}
\toprule
Symbol & Meaning \\
\midrule
$s$ & Biomedical scenario or task instance. \\
$\mathcal{D}^{(s)}$ & Scenario evidence, equal to $(X^{(s)},Y^{(s)},C^{(s)})$. \\
$X^{(s)}$ & Observable features or measurements available under the scenario schema. \\
$Y^{(s)}$ & Outcomes, responses, or configured evidence variables used for evaluation. \\
$C^{(s)}$ & Task context, including scenario description, endpoint definition, and discovery constraints. \\
$\mathcal{K}^{(s)}$ & Source-documented prior knowledge records available before discovery. \\
$\kappa_a$ & One prior record with identifier, statement, source descriptor, scope, and transform metadata. \\
$\mathcal{S}^{(s)}$ & Machine-readable schema with variable types, admissible windows, forbidden fields, missingness rules, and composition constraints. \\
$d_t,\eta_t$ & Active direction and bounded search guidance at round $t$. \\
$\mathcal{Z}_t,\mathcal{G}_t$ & Direction-conditioned vocabulary and grounding plan. \\
$q,\mathcal{Q}_t$ & Candidate rule/finding and admissible candidate family. \\
$J_{\mathrm{sup}}^{(s)},J_{\mathrm{eff}}^{(s)}$ & Scenario-defined support and effect objectives. \\
$\mathcal{F}_t,\lambda_t$ & Pareto frontier and execution log returned by the lower level. \\
$f_t,m_t$ & Feedback packet and compact working memory. \\
$\mathcal{P}^{(s)}$ & Exported set of scenario-grounded propositions. \\
$p_i=(d_i,q_i,\omega_i)$ & One reportable proposition: direction, executable grounded object, and evidence record. \\
$\omega_i$ & Evidence and provenance record attached to a reported proposition. \\
\bottomrule
\end{tabularx}
\end{table}

The main text keeps candidate search at the protocol level. Executably, the active direction and validated grounding plan induce a vocabulary $\mathcal{Z}_t$ and validity function $\Gamma_t^{(s)}$, giving
\begin{equation}
\mathcal{Q}_t=
\{q\in\mathcal{H}(\mathcal{Z}_t): |q|\le L,
\Gamma_t^{(s)}(q)=1,
J_{\mathrm{sup}}^{(s)}(q)\ge \rho_t\}.
\end{equation}
In the tabular instantiations used here, the default object in $\mathcal{H}(\mathcal{Z}_t)$ is a bounded conjunction
\begin{equation}
q(x)=\bigwedge_{j=1}^{\ell}\mathbb{I}[z_{k_j}\,\bowtie_j\,\tau_j],
\qquad \ell\le L,
\end{equation}
where literals are drawn from raw schema-visible observables or approved virtual/dynamic operators. In non-clinical settings, the same notation denotes an executable context-bounded finding rather than a clinical subgroup rule.

The scenario adapter used by Algorithm~\ref{alg:scene_protocol} can be written as
\begin{equation}
\mathcal{A}^{(s)}=
\big(\mathcal{O}^{(s)},\Gamma^{(s)},J_{\mathrm{sup}}^{(s)},J_{\mathrm{eff}}^{(s)},
\mathcal{T}^{(s)},\Omega^{(s)}\big),
\end{equation}
where $\mathcal{O}^{(s)}$ specifies observable families, $\Gamma^{(s)}$ validity, $J_{\mathrm{sup}}^{(s)}$ support, $J_{\mathrm{eff}}^{(s)}$ effect, $\mathcal{T}^{(s)}$ optional grounding operators, and $\Omega^{(s)}$ split, provenance, diagnostic, and output schemas.

\begin{table}[!htb]
\caption{Scenario formalization under the common SCENE adapter interface.}
\label{tab:scenario_formalization}
\centering
\small
\setlength{\tabcolsep}{2.4pt}
\begin{tabularx}{\linewidth}{@{}L{1.7cm}XX@{}}
\toprule
Component & Clinical-trial scenario & LINCS L1000 perturbational scenario \\
\midrule
Evidence & Subjects with admissible subgroup covariates, treatment arm, endpoint, and protocol context. Treatment and outcome fields are evaluation fields, not subgroup literals. & Perturbational profiles or signatures with cell, perturbagen, dose/time, response or connectivity score, and optional signature-derived attributes. \\
Candidate $q$ & Bounded subgroup conjunction over baseline or explicitly landmarked pre-decision summaries. & Bounded context conjunction over admissible perturbation, cell, dose/time, compound, or signature attributes. \\
Validity & Type-correct literals, no treatment/outcome/post-outcome leakage, admissible windows, nonempty arms, and support floor. & Compatible context literals, admissible dose/time windows, matched or recurrent admissible support, and the support floor. \\
Support & $\min\{n_{\mathrm{treat}}(q),n_{\mathrm{ctrl}}(q)\}$. & Number of matched or recurrent admissible evidence instances. \\
Effect & Configured within-subgroup treatment-benefit contrast. & Configured connectivity or response contrast, blocked when a valid blocking design is available. \\
Output & Scenario-grounded proposition containing a subgroup rule, evidence record, and provenance. & Scenario-grounded proposition containing a perturbational context finding, evidence record, and provenance. \\
\bottomrule
\end{tabularx}
\end{table}

\subsection{Upper-Level Controller and Role Artifacts}
\label{app:agent_roles}

The upper level is implemented through typed role artifacts rather than requiring physically separate models. The same language model may instantiate multiple roles, but each role consumes and emits a different typed object. This design is intended to make the contextualization process replayable and auditable: language-model calls may propose directions, grounding maps, or critiques, but they do not directly validate biomedical claims or bypass deterministic controllers.

The accepted information flow is
\begin{equation}
\begin{gathered}
(\mathcal{K}^{(s)},C^{(s)},m_{t-1},f_{t-1})
\xrightarrow{\text{direction planning}} (d_t,\eta_t)
\xrightarrow{\text{semantic grounding}} \mathcal{G}_t\\
\xrightarrow{\text{search}} (\mathcal{F}_t,\lambda_t)
\xrightarrow{\text{feedback}} (f_t,m_t,\bar{\eta}_{t+1}).
\end{gathered}
\end{equation}
The upper level emits search contracts, not final biomedical conclusions. These contracts define what should be grounded next, which schema-visible objects may be used, and how the lower-level search should be bounded.

\begin{table}[!htb]
\caption{Upper-level role artifacts and fail-closed behavior.}
\label{tab:artifact_schemas}
\centering
\small
\setlength{\tabcolsep}{2.5pt}
\begin{tabularx}{\linewidth}{@{}lXX@{}}
\toprule
Artifact & Required content & Validation or fallback \\
\midrule
Direction $d_t$ & Search intent, supporting knowledge IDs or \texttt{model\_\allowbreak suggested}, grounding cues, target preferences, rationale. & Unknown source IDs are rejected or marked \texttt{model\_\allowbreak suggested}; unsupported cues cannot be reported as prior-supported. \\
Guidance $\eta_t$ & Operator mixture, generation budget, support floor, bounds restrictions, avoid list, dynamic-feature toggles. & Numeric controls are clipped to manifest ranges; invalid toggles are ignored and logged. \\
Grounding plan $\mathcal{G}_t$ & Matched features/windows/composites, valid literal keys, invalid combinations, seed candidates, grounding rationale. & Unmaterialized or leakage-risk literals are dropped before search. \\
Feedback $f_t$ & Frontier summary, alignment gaps, failure modes, support-effect trade-offs, stability notes. & Missing diagnostics trigger conservative defaults. \\
Memory $m_t$ & Retained/rejected cues, recent directions, stable families, failure modes, next-round hint. & Memory is bounded and summarized; it is not converted into external prior knowledge without a new manifest record. \\
\bottomrule
\end{tabularx}
\end{table}

The \textsc{UpdateFeedback} call in Algorithm~\ref{alg:scene_protocol} is implemented by the deterministic priority rule in Algorithm~\ref{alg:frontier_feedback}. This rule maps invalidity, support failures, diversity collapse, diagnostic contrast, alignment gaps, and stagnation into bounded next-round actions.

\begin{algorithm}[t]
\caption{UpdateFeedback / FrontierFeedbackUpdate}
\label{alg:frontier_feedback}
\begin{algorithmic}[1]
\Require frontier $\mathcal{F}_t$, execution log $\lambda_t$, direction $d_t$, memory $m_{t-1}$, manifest thresholds $\tau$
\Ensure feedback $f_t$, memory $m_t$, bounded controls $\bar{\eta}_{t+1}$
\State compute \texttt{invalid\_rate}, \texttt{low\_support\_rate}, \texttt{diversity}, \texttt{diagnostic\_contrast}, \texttt{alignment\_gap}, and \texttt{stagnation} from $(\mathcal{F}_t,\lambda_t)$
\If{\texttt{leakage\_flag} or \texttt{schema\_violation} or \texttt{invalid\_rate} $> \tau_{\mathrm{invalid}}$}
    \State \texttt{action} $\gets$ \texttt{replace}
\ElsIf{$\mathcal{F}_t=\emptyset$}
    \State \texttt{action} $\gets$ \texttt{broaden} if one manifest-permitted retry remains; otherwise \texttt{replace}
\ElsIf{\texttt{low\_support\_rate} $> \tau_{\mathrm{low\_support}}$ or \texttt{alignment\_gap} is high}
    \State \texttt{action} $\gets$ \texttt{broaden}
\ElsIf{\texttt{diversity} $< \tau_{\mathrm{diversity}}$ or \texttt{stagnation} exceeds patience}
    \State \texttt{action} $\gets$ \texttt{narrow}
\ElsIf{\texttt{diagnostic\_contrast} is high and support is adequate}
    \State \texttt{action} $\gets$ \texttt{preserve} with conservative control deltas
\Else
    \State \texttt{action} $\gets$ \texttt{preserve}
\EndIf
\State construct $f_t$ with the action, frontier summary, failure modes, alignment gaps, and stability notes
\State update $m_t$ with retained/rejected cues, stable families, and recent failure modes
\State set $\bar{\eta}_{t+1}$ by clipping any control deltas to manifest bounds
\State \Return $(f_t,m_t,\bar{\eta}_{t+1})$
\end{algorithmic}
\end{algorithm}

The compact feedback statement in the main text is therefore instantiated by a deterministic priority order. Default thresholds are recorded in the manifest, e.g., diagnostic-contrast high/low cutoffs, diversity floor, invalid-rate ceiling, low-support ceiling, and upper-level stagnation patience. These values may be tuned per scenario before discovery, but not by final held-out or test metrics.

Language-model calls are validated as JSON-producing proposal functions. A strategist artifact contains \texttt{direction}, \texttt{knowledge\_ids}, \texttt{grounding\_cues}, \texttt{target\_preference}, \texttt{control\_adjustment}, and \texttt{rationale}. A proposer artifact contains \texttt{semantic\_matches}, \texttt{grounding\_map}, \texttt{candidate\_literals}, \texttt{invalid\_combinations}, \texttt{seed\_candidates}, and optional \texttt{feature\_proposals}. Critic/refiner artifacts contain frontier summaries, failure modes, stability notes, next-direction hints, memory updates, and bounded control deltas. Extra top-level fields, unknown feature identifiers, unregistered source IDs, non-finite numbers, and claims of validation success are rejected.

\subsection{Role Instruction Contracts}
\label{app:role_prompt_contracts}

SCENE specifies role instructions as contracts rather than as unconstrained prompts. Each language-model call is treated as a JSON-producing proposal function: the model may suggest a direction, grounding, critique, or refinement, but its output has no executable effect until parsed, checked against the frozen manifest, and validated by deterministic controllers. The controller supplies only discovery-visible context, including source records, schema summaries, frontier feedback, split-visibility flags, and bounded control ranges. It rejects malformed JSON, unknown source identifiers, unknown feature keys, forbidden fields, non-finite values, out-of-range controls, and unsupported biomedical assertions. The following compact contracts summarize the role interfaces used by SCENE; scenario-specific prompts instantiate the placeholders with manifest-visible context, but the executable boundary is the schema and controller rather than the wording of the prompt.

\begin{appendixbox}{Strategist contract}
role: strategist
input:
{
  "scenario_context": "<task and discovery-visible split context>",
  "knowledge_records": [{"id": "kappa_a", "cue": "..."}],
  "schema_summary": {"feature_families": ["..."], "forbidden": ["..."]},
  "memory": "<bounded prior-round summary>",
  "frontier_summary": "<recent frontier status and trade-offs>",
  "control_bounds": "<manifest ranges>"
}
instructions:
- Propose a search direction, not a biomedical conclusion.
- Use supplied knowledge IDs only; otherwise mark "model_suggested".
- Map concepts to schema-visible grounding cues or mark them unresolved.
- Propose a replacement or revised direction when frontier feedback indicates failure.
- Do not report effect size, statistical significance, or validation status.
output_json:
{
  "direction": "...",
  "status": "source_supported|model_suggested|feedback_derived",
  "knowledge_ids": ["kappa_a"],
  "grounding_cues": ["..."],
  "target_preference": {"prefer": ["..."], "avoid": ["..."]},
  "control_adjustment": [{"param": "...", "delta": 0.0, "reason": "..."}],
  "rationale": "short evidence-bounded rationale"
}
\end{appendixbox}

\begin{appendixbox}{Proposer-grounder contract}
role: proposer_grounder
input:
{
  "direction": "<selected direction>",
  "schema_summary": "<feature keys, types, windows, grammar>",
  "allowed_operators": ["mutation", "crossover", "tuning", "injection"],
  "forbidden_fields": ["..."],
  "feature_statistics": "<discovery-visible summaries only>"
}
instructions:
- Use only supplied feature keys, windows, and operators.
- Do not use treatment, outcome, post-outcome, or
  validation-derived fields as literals.
- Translate each cue into matched features, admissible windows,
  candidate literals, or an explicit invalid/unresolved entry.
- Do not claim effect, significance, or validation success.
output_json:
{
  "semantic_matches": [{"cue": "...", "features": ["..."], "windows": ["..."]}],
  "grounding_map": {"cue_id": {"literal_keys": ["..."], "status": "matched"}},
  "candidate_literals": [{"feature": "...", "op": ">|<=", "value_source": "grid"}],
  "seed_candidates": [["literal_key_a", "literal_key_b"]],
  "invalid_combinations": [{"cue": "...", "reason": "not_schema_visible"}],
  "feature_proposals": [{"request": "...", "reason": "..."}],
  "safety": {"uses_forbidden_field": false}
}
\end{appendixbox}

\begin{appendixbox}{Critic-refiner contract}
role: critic_refiner
input:
{
  "direction": "<current direction>",
  "frontier_summary": "<Pareto set, support/effect, diversity>",
  "execution_log": "<invalid, low-support, weak-alignment cases>",
  "current_controls": "<bounded runtime controls>"
}
instructions:
- Use only frontier and execution-log evidence; do not invent experience.
- Summarize failure_modes and stability_notes separately.
- Choose one next action: preserve, narrow, broaden, or replace.
- Keep control_delta bounded, conservative, and next-round only.
- Do not convert feedback into external prior knowledge.
output_json:
{
  "frontier_summary": {"status": "coherent|empty|unstable|low_support"},
  "failure_modes": ["..."],
  "stability_notes": ["..."],
  "action": "preserve|narrow|broaden|replace",
  "next_direction_hint": "...",
  "control_delta": [{"param": "...", "delta": 0.0, "reason": "..."}],
  "memory_update": "bounded summary"
}
\end{appendixbox}

When \texttt{feature\_proposals} are present, the scenario adapter may request one virtual or dynamic feature plan using only registered tables, allowed variables, admissible windows, and leakage-safe operators; invalid or unmaterialized plans are dropped and logged. If executable synthesis is used for an accepted feature plan, it is separately sandboxed and validated before generated features enter search.

\subsection{Role Artifact Templates and One-Round Trace}
\label{app:role_artifact_trace}

The following schematic artifact illustrates the role interface using a simplified metformin-response example. Field values are illustrative rather than experimental observations, and the object is intended to show the replay contract rather than measured support or effect. The example emphasizes that a reported output is a proposition-level object containing a direction, an executable grounded object, and an evidence record.

\begin{appendixbox}{Metformin-response example}
{
  "knowledge_record": {
    "id": "kappa_metabolic_01",
    "cue": "metabolic dysregulation may modify metformin response",
    "scope": "clinical subgroup discovery"
  },
  "direction": {
    "text": "search for an insulin-resistance or metabolic-risk subgroup",
    "knowledge_ids": ["kappa_metabolic_01"],
    "status": "source_supported"
  },
  "semantic_match": {
    "features": ["baseline_BMI", "fasting_glucose", "HOMA_IR_if_available"],
    "windows": ["baseline"],
    "dropped": ["post-treatment glucose"]
  },
  "search_guidance": {
    "support_floor": "manifest_min_balanced_support",
    "operators": ["mutation", "crossover", "tuning"],
    "avoid": ["treatment", "outcome", "post-outcome"]
  },
  "grounded_object": {
    "candidate_key": "baseline_BMI>tau_1 AND fasting_glucose>tau_2",
    "threshold_source": "predeclared discovery grid"
  },
  "frontier_log": {
    "frontier_status": "coherent",
    "failure_modes": ["HOMA_IR unavailable in this schema"]
  },
  "feedback": {
    "action": "preserve",
    "memory_update": "metabolic baseline features remain productive"
  },
  "reported_proposition": {
    "direction": "search for an insulin-resistance or metabolic-risk subgroup",
    "grounded_object": "baseline_BMI>tau_1 AND fasting_glucose>tau_2",
    "evidence_record": {
      "support": "...",
      "effect": "...",
      "source_id": "kappa_metabolic_01",
      "NoLeak": true,
      "split_id": "...",
      "diagnostics": "..."
    }
  }
}
\end{appendixbox}

A one-round trace is therefore an artifact chain rather than a transcript of agent conversation:
\begin{equation}
\kappa_a
\rightarrow d_t
\rightarrow (\mathcal{G}_t,\eta_t)
\rightarrow (\mathcal{F}_t,\lambda_t)
\rightarrow f_t
\rightarrow p_i .
\end{equation}
The knowledge record supplies a provenance-bearing cue; the direction states a bounded search intent; schema match and guidance translate that intent into executable schema objects; the frontier and log expose what the lower level could or could not ground; feedback updates memory and controls; and the final arrow emits a proposition-level object $p_i=(d_i,q_i,\omega_i)$ only if the candidate also satisfies the reporting predicate in Appendix~\ref{app:reporting_boundaries}. This trace is illustrative of the interface contract and does not encode a measured effect size, support count, or empirical success claim.

\subsection{Evolutionary Grounding Procedure}
\label{app:evolution_details}

The grounding loop executes the proposer's search space under the current direction. It maintains a population $\mathcal{C}_{t,g}$, an archive $\mathcal{R}_t$, and an append-only log $\lambda_t$. The elements of this population are executable candidates: clinical subgroup rules in clinical tasks and context-bounded findings in L1000 tasks. Pareto dominance uses only $(J_{\mathrm{eff}}^{(s)},J_{\mathrm{sup}}^{(s)})$; pseudo-validation can diagnose instability or break exact ties after Pareto rank and diversity, but it cannot create dominance.

\begin{algorithm}[t]
\caption{Direction-conditioned grounding at round $t$}
\label{alg:scene_grounding}
\begin{algorithmic}[1]
\Require direction $d_t$, contract $\mathcal{G}_t=(\mathcal{Z}_t,\Gamma_t^{(s)},\mathcal{C}_{t,0})$, evidence $\mathcal{D}^{(s)}$, adapter $\mathcal{A}^{(s)}$, controls $\eta_t=(\pi_t,H_t,\rho_t,\beta_t)$
\Ensure frontier $\mathcal{F}_t$ and execution log $\lambda_t$
\State freeze replay configuration $\xi_t$ \Comment{seed, split, adapter, grammar, controls}
\For{$g \gets 0$ to $H_t$}
    \State $\hat{\mathcal{C}}_{t,g} \gets \Call{Validate}{\mathcal{C}_{t,g},\Gamma_t^{(s)},\rho_t}$ \Comment{deduplicate and filter}
    \State $\mathcal{R}_{t,g} \gets \Call{Evaluate}{\hat{\mathcal{C}}_{t,g},\mathcal{D}^{(s)},\mathcal{A}^{(s)}}$ \Comment{effect/support diagnostics}
    \State update archive $\mathcal{R}_t$ and log $\lambda_t$ with kept, rejected, and invalid cases
    \If{$g < H_t$}
        \State $\mathcal{U}_{t,g} \gets \Call{Vary}{\hat{\mathcal{C}}_{t,g},d_t,\mathcal{G}_t,\pi_t}$ \Comment{mutation, crossover, tuning, injection}
        \State $\mathcal{C}_{t,g+1}\gets\Call{Select}{\hat{\mathcal{C}}_{t,g}\cup\mathcal{U}_{t,g},\mathcal{R}_t,\eta_t}$ \Comment{rank, diversity, key order}
    \EndIf
\EndFor
\State \Return $\mathcal{F}_t\leftarrow\operatorname{Pareto}(\mathcal{R}_t)$, $\lambda_t$
\end{algorithmic}
\end{algorithm}

\begin{table}[!htb]
\caption{Executable grounding controls recorded in $\xi_t$ and $\lambda_t$.}
\label{tab:evolution_controls}
\centering
\small
\setlength{\tabcolsep}{2.2pt}
\begin{tabularx}{\linewidth}{@{}L{1.8cm}XX@{}}
\toprule
Component & Operation & Logged state \\
\midrule
Initialization & Mix proposer seeds with admissible diversity-oriented candidates. & Seed source, validity status, duplicates, support failures. \\
Mutation & Add, drop, or modify literals, thresholds, bins, dose/time windows, or landmark windows. & Parent key, edited field, grid identifier, validity reason. \\
Crossover & Combine compatible partial structures from frontier or archive candidates. & Parent keys, compatibility checks, resulting semantic key. \\
Tuning & Move thresholds/windows along a predeclared grid around supported candidates. & Old/new values, grid step, support/effect change. \\
Injection & Introduce fresh candidates from grounding cues, feature proposals, or refiner hints. & Trigger, source cue, source record or \texttt{model\_\allowbreak suggested} tag. \\
Selection & Apply validity, deduplication, support floor, Pareto rank, diversity, deterministic key order. & Full ordering key and retained/rejected status. \\
Replay & Freeze seed stream, split ID, adapter version, candidate grammar, population size, archive cap. & Configuration hash and role-output hash when available. \\
\bottomrule
\end{tabularx}
\end{table}

\begin{table}[!htb]
\caption{Default executable values used when a scenario manifest does not specify stricter settings.}
\label{tab:evolution_defaults}
\centering
\small
\setlength{\tabcolsep}{4pt}
\begin{tabularx}{\linewidth}{@{}L{3.2cm}L{5cm}X@{}}
\toprule
Field & Default & Role in execution \\
\midrule
Population $N_t$ & Clinical: 240; L1000: 160; smoke: 40 & Candidate population and offspring budget per generation after clipping to $[40,320]$. \\
Generations $H_t$ & Clinical: 20; L1000: 14; smoke: 2 & Full runs are clipped to $[5,40]$; smoke mode is flagged as non-evaluation. \\
Rule length and support & $L=3$; clinical $\rho_t\ge30$ balanced-arm support; L1000 $\rho_t\ge60$ instances & Scenario adapters may raise floors; broadening cannot lower them below the manifest floor. \\
Operator mix $\pi_0$ & Mutation 0.60, crossover 0.25, injection 0.15 & Each coordinate is clipped to $[0.05,0.80]$ and renormalized. \\
Retry/archive & Retry cap $5N_t$; archive cap $\min(1000,5N_t)$ & Invalid, duplicate, or low-support offspring are retried; archives are pruned by Pareto rank, diversity, then key order. \\
Stagnation & 5 generations lower-level; 3 rounds upper-level & Triggers early stopping below or direction replacement above. \\
Controller thresholds & $\Delta_{\mathrm{high}}=0.65$, $\Delta_{\mathrm{low}}=0.25$, diversity floor 0.20, invalid-rate ceiling 0.50, low-support-rate ceiling 0.60 & Used for preserve/narrow/broaden/replace hints before final validation is visible. \\
\bottomrule
\end{tabularx}
\end{table}

Scenario manifests may set stricter floors or disable operators, but final validation metrics cannot alter search defaults for the same run.

\subsection{Grounding Operators}
\label{app:grounding_operators}

Some directions refer to concepts that are only partially visible in raw observables. This is the main operational challenge of knowledge contextualization: broad biomedical ideas are rarely expressed in the same vocabulary as the dataset schema. SCENE can therefore expand the grounding vocabulary from $\mathcal{Z}^{(0)}$ to
\begin{equation}
\mathcal{Z}_t=\mathcal{Z}^{(0)}\cup\mathcal{Z}^{(\mathrm{virt})}\cup\mathcal{Z}^{(\mathrm{dyn})}.
\end{equation}
Here $\mathcal{Z}^{(0)}$ contains raw schema-visible observables, $\mathcal{Z}^{(\mathrm{virt})}$ contains approved virtual composites, and $\mathcal{Z}^{(\mathrm{dyn})}$ contains approved dynamic or windowed summaries. The executable grammar is fixed by the scenario manifest before discovery:
\begin{equation}
\begin{aligned}
q &::= \ell_1\wedge\cdots\wedge \ell_r,\qquad 1\le r\le L,\\
\ell &::= z \bowtie \tau \mid z\in B \mid \operatorname{missing}(z)=b,\\
z &::= x_k \mid \operatorname{virt}_r(x_{a_1},\ldots,x_{a_m})
\mid \operatorname{dyn}_r(x_{a_1},\ldots,x_{a_m};w),\\
\bowtie &\in \{<,\le,>,\ge\}.
\end{aligned}
\end{equation}
Here $x_k$ must be a schema-visible observable, $B$ and $\tau$ must come from predeclared grids or manifest-approved category sets, $w$ is an admissible temporal or perturbational window, and every operator instance must return a typed feature with provenance. Mutation, crossover, tuning, and injection can edit only objects generated by this grammar; invalid literals are dropped before scoring and recorded in $\lambda_t$.

\begin{table}[!htb]
\caption{Examples of how broad biomedical cues can be grounded under different scenario schemas. These examples illustrate the grounding interface; they are not additional rules or reported findings.}
\label{tab:grounding_examples}
\centering
\small
\setlength{\tabcolsep}{3pt}
\begin{tabularx}{\linewidth}{@{}L{2.5cm}XXX@{}}
\toprule
Broad cue & Clinical-trial grounding & L1000 grounding & Gate \\
\midrule
Metabolic dysregulation & BMI, fasting glucose, triglycerides, HOMA-IR, insulin-related summaries when available. & Pathway or signature summaries related to metabolic programs when materialized from registered profiles. & Schema-visible variables only; no outcome or validation-derived fields. \\
Treatment-response heterogeneity & Baseline or landmarked pre-decision subgroup literals, with treatment and endpoint reserved for evaluation. & Context literals over cell, perturbagen, dose, time, and admissible signature attributes. & Clinical treatment/outcome fields cannot become literals. \\
Dynamic response context & Predeclared longitudinal or trajectory summaries within admissible windows. & Dose-time summaries or perturbational signature summaries within allowed windows. & Window must be declared before search and replayable on holdout. \\
Mechanistic assay context & Available only when represented by baseline, laboratory, pathology, or approved derived measurements. & Cell-line context, perturbagen class, pathway summaries, and target-compatible signature attributes. & Comparison pools and target-derived labels cannot be constructed post hoc. \\
\bottomrule
\end{tabularx}
\end{table}

Virtual operators derive scalar or categorical features from existing tables, such as thresholded composites, ratios, bins, or feature-family summaries. Dynamic operators summarize auxiliary longitudinal or perturbational traces, such as pre/post changes, event counts, window aggregates, dose-time summaries, or perturbational signature summaries. Both return typed features with construction provenance, missingness indicators, admissibility tags, and fail-closed behavior. A proposed feature that cannot be materialized from available columns, violates the schema, or uses forbidden response/validation information is excluded and logged. Thus, generated features expand the grounding space but do not relax the executable boundary.

\subsection{Reporting, Provenance, and Split Boundaries}
\label{app:reporting_boundaries}

A candidate may guide feedback even when it is not reportable, provided it was produced from discovery-visible information. Export is stricter than feedback: a candidate becomes a reported scenario-grounded proposition only if it has frontier membership, validity, support, provenance, leakage checks, and diagnostics. We write the reporting predicate as
\begin{equation}
\begin{aligned}
\operatorname{Report}(q,d_t,t)
={}&\mathbf{1}\{q\in\mathcal{F}_t,\,
\Gamma_t^{(s)}(q)=1,\\
&J_{\mathrm{sup}}^{(s)}(q)\ge\rho_t,\,
\operatorname{Prov}(q,d_t,t),\\
&\operatorname{NoLeak}^{(s)}(q),\,
\operatorname{Diag}(q,t)\}.
\end{aligned}
\end{equation}
When this predicate is satisfied, SCENE exports a proposition
\begin{equation}
p_i=(d_t,q,\omega_i),
\end{equation}
where $\omega_i$ stores the evidence and provenance fields. If the predicate fails, the candidate may remain in the execution log for feedback, but it is not exported as contextualized knowledge.

\begin{table}[!htb]
\caption{Field-level checks required before a candidate can be exported as a scenario-grounded proposition.}
\label{tab:report_fields}
\centering
\small
\setlength{\tabcolsep}{4pt}
\begin{tabularx}{\linewidth}{@{}L{2.2cm}XX@{}}
\toprule
Check & Required fields & Fail-closed behavior \\
\midrule
Provenance & Candidate, round, direction, manifest, schema, split, replay-config IDs; knowledge-record IDs or \texttt{model\_\allowbreak suggested}; grounding map; operator provenance. & Missing IDs, hashes, grounding entries, or operator provenance prevent export. \\
No leakage & Forbidden-field list, literal keys, operator input columns, split lineage, L1000 comparison-pool lineage. & Clinical treatment/outcome/post-outcome literals fail. L1000 response/connectivity-derived, validation-derived, comparison-pool, or post hoc block features fail. \\
Diagnostics & Effect/support, frontier membership, Pareto rank or archive status, diagnostic contrast, pseudo-validation status or missingness, blocking status where relevant, invalid/low-support counts. & Missing diagnostics prevent export, although the case may remain in $\lambda_t$ for feedback. \\
\bottomrule
\end{tabularx}
\end{table}

Split lineage is frozen before discovery. Source-documented priors may be consumed by strategist and proposer; within-run memory and logs may guide the same discovery episode; discovery-visible pseudo-validation may guide conservative bounded control update; final held-out or test archives are post-discovery audit artifacts only. Importing a completed validation summary into a later run requires a new frozen knowledge record with source run, split ID, timestamp, evidence-status label, and allowed consumer roles, so that validation-derived knowledge is explicit rather than silently recycled.

A run manifest records scenario ID, freeze time, schema hash, split ID, controller thresholds, control bounds, allowed feature families, forbidden fields, enabled operators, role-output hashes when available, adapter version, and final-validation boundary. Source-free suggestions may be logged as \texttt{model\_\allowbreak suggested} only for data-grounded search hypotheses; they cannot be cited as external biomedical knowledge or prior-supported mechanisms.

\paragraph{Minimal manifest fields.}
An executable run stores a compact JSON-like manifest before discovery:
\begin{appendixbox}{Manifest contract}
{
  "scenario_id": "clinical_trial_or_l1000_task",
  "freeze_time_utc": "...",
  "schema_hash": "sha256:...",
  "split_id": "discovery_split_seed",
  "knowledge_records": [{"record_id": "...", "source_type": "..."}],
  "controller_policy": {
    "action_priority": ["replace", "broaden", "narrow", "preserve"],
    "delta_high": 0.65,
    "diversity_floor": 0.20,
    "invalid_rate_ceiling": 0.50
  },
  "control_bounds": {
    "population_size": [40, 320],
    "generations": [5, 40],
    "operator_probability": [0.05, 0.80],
    "max_rule_len": 3
  },
  "final_validation_boundary": {
    "visible_to_discovery": false,
    "metrics_feed_back": false
  }
}
\end{appendixbox}

Actual manifests additionally store scenario-specific paths, hashes, replay seeds, adapter versions, enabled operators, forbidden fields, and role-output hashes when role outputs are cached.

\subsection{Scenario-Specific Estimators}
\label{app:scenario_estimators}

The scenario adapter defines the concrete effect and support functions used by lower-level search. The main text presents these functions abstractly as $J_{\mathrm{eff}}^{(s)}$ and $J_{\mathrm{sup}}^{(s)}$; this section records the two main instantiations used in the paper.

For clinical treatment-benefit discovery and a fixed admissible rule $q$, the configured effect is a descriptive bad-outcome reduction:
\begin{equation}
\begin{aligned}
J_{\mathrm{eff}}^{(\mathrm{trial})}(q)
={}&\Pr(Y_{\mathrm{bad}}=1\mid q(X)=1,T=0)\\
&-\Pr(Y_{\mathrm{bad}}=1\mid q(X)=1,T=1),\\
J_{\mathrm{sup}}^{(\mathrm{trial})}(q)
={}&\min\{n_1(q),n_0(q)\}.
\end{aligned}
\end{equation}
This sign convention matches the main clinical endpoint: larger values indicate a larger reduction in unfavorable outcomes under the treatment arm relative to the control arm. Because SCENE adaptively searches over many candidates and refines later directions using discovery feedback, reported subgroup effects are exploratory, selection-conditioned estimates. Unless a separate pre-registered or multiplicity-adjusted analysis is performed, SCENE does not report valid post-selection $p$-values, treatment-labeling claims, or external clinical recommendations. Rules using post-baseline summaries are marked as predictive or associational unless a valid landmark interpretation is declared by the scenario.

For LINCS L1000, the configured effect is a context-versus-background target-response or connectivity contrast. The adapter freezes blocking variables, comparison-pool eligibility, exclusion lists, threshold grids, and feature-construction rules before search. A candidate may determine selected-instance membership through its admissible literals, but it may not determine controls after observing its effect. Let $c_i$ be the response or connectivity score. For blocking strata $\mathcal{B}(q)$ with both selected and comparison evidence,
\begin{equation}
\begin{aligned}
\mathcal{M}_b(q)&=\{i\in\mathcal{M}(q):b(i)=b\},\\
\mathcal{M}^{0}_b(q)&=\{i\in\mathcal{M}^{0}(q):b(i)=b\},
\end{aligned}
\end{equation}
with weights $w_b(q)=|\mathcal{M}_b(q)|/\sum_{b'\in\mathcal{B}(q)}|\mathcal{M}_{b'}(q)|$. The blocked contrast is
\begin{equation}
J_{\mathrm{eff}}^{(\mathrm{l1000,blk})}(q)=
\sum_{b\in\mathcal{B}(q)}w_b(q)
\left(\bar{c}_{\mathcal{M}_b(q)}-\bar{c}_{\mathcal{M}^{0}_b(q)}\right).
\end{equation}
If no admissible blocking design is available, the adapter uses the predeclared unblocked fallback $\bar{c}_{\mathcal{M}(q)}-\bar{c}_{\mathcal{M}^{0}(q)}$ when support conditions are met and records \texttt{blocking\_status=unblocked}. L1000 findings are descriptive perturbational associations within the configured assay and matching design, not causal compound effects or validated mechanisms.

\subsection{Experiment Matrix and Reporting Contract}
\label{app:experiment-matrix}

This subsection links the method objects defined above to the three empirical reporting settings. It does not introduce additional search rules; detailed task construction and metric definitions are deferred to Appendices~\ref{app:clinical-task-definitions}--\ref{app:table3-implementation}. SCENE is evaluated in three settings: (i) clinical subgroup discovery, (ii) perturbational mechanism discovery, and (iii) downstream few-shot contextual knowledge reuse. These settings share the same high-level discovery backbone but differ in unit of analysis, target variable, candidate language, and evaluation functional.

\paragraph{Clinical subgroup discovery.}
The unit is one patient. Each run operates on one clinical task frame, one paired train--holdout split, and one discovery seed. Each method returns an ordered rule list, but only the first committed rule is used in the paper-facing best-1 benchmark. Holdout outcomes are never used to choose or reorder rules. Under the proposition notation above, the grounded object $q_i$ is a subgroup rule and $\omega_i$ stores the held-out replay evidence, support, diagnostics, and provenance.

\paragraph{Perturbational mechanism discovery.}
The unit is one L1000 signature instance. Each run fixes one target mechanism, one split, one search seed, and one ablation profile. Discovery is executed on the training split only; all reported perturbational metrics are computed by replaying frozen rules on the held-out split. Here $q_i$ is a context-bounded perturbational finding, and $\omega_i$ records target-response evidence, support, replay diagnostics, and split lineage.

\paragraph{Contextual knowledge reuse.}
The unit is one downstream tabular row, either a patient or a signature instance. Each run fixes one scene, one outer split, one few-shot exemplar set, one prompt condition, and one deterministic LLM decoding policy. Discovered propositions are converted into compact knowledge cards only from train-side discovery and selected only on validation data. Test rows are used only once for final prediction and metric computation. This setting evaluates whether proposition-level outputs can be reused as auxiliary context; it is not an independent validation of every biomedical mechanism encoded in a card.

Across all three settings, we maintain strict train/validation/test separation: held-out outcomes, held-out connectivity, and final downstream labels are not used to revise prompts, re-select rules, or change knowledge-card content within the same evaluation episode.

\subsection{Clinical Task Definitions}
\label{app:clinical-task-definitions}
Table~\ref{tab:clinical-task-definitions} expands the short clinical task headers used in Table~\ref{tab:clinical_best1_benefit_revised}. The six tasks are constructed from two de-identified patient-level trial tables associated with the registered trials in the table; ClinicalTrials.gov provides identifiers and protocol context, not patient-level modeling rows~\citep{nct00174655,nct02491333,zarin2011clinicaltrials}. The trials were selected by pre-analysis feasibility and coverage criteria: randomized arms, patient-level covariates, interpretable treatment/endpoints, enough support for repeated replay, and coverage of more than one biomedical setting. No trial was added or removed after inspecting SCENE held-out performance. The NCT00174655 oncology table is accessed through Project Data Sphere~\citep{projectdatasphere_access,green2015projectdatasphere}; the NCT02491333 PCOS table is derived from the Dryad release of Wen et al.~\citep{wen2021dryad,wen2022acupuncture}. These are treatment-benefit subgroup-discovery tasks: each asks whether a prioritized subgroup rule identifies patients for whom treatment reduces a task-specific unfavorable clinical outcome, rather than estimating a global average treatment effect.

\begin{table}[!htb]
\caption{Clinical task definitions used in the treatment-benefit subgroup-discovery benchmark. BL/TR denote baseline/static and trajectory-enhanced feature views.}
\label{tab:clinical-task-definitions}
\centering
\small
\setlength{\tabcolsep}{4pt} %
\begin{tabularx}{\linewidth}{@{}lXXX@{}}
\toprule
Header & Trial and clinical setting & Treatment question and feature view & Response or unfavorable-outcome definition \\
\midrule
B-D-BL & NCT00174655; node-positive operable breast cancer. & Doxorubicin-containing chemotherapy-schedule comparison using baseline/static covariates. & Bad outcome is recurrence or death by the preprocessed disease-free-survival (DFS) status. \\
B-D-TR & NCT00174655; node-positive operable breast cancer. & Same doxorubicin-schedule comparison using trajectory-enhanced features. & Bad outcome is recurrence or death by the preprocessed DFS status. \\
P-M-BL & NCT02491333; polycystic ovary syndrome (PCOS) under sham acupuncture. & Metformin versus placebo using baseline/static covariates. & Good response is month-4 HOMA-IR reduction of at least 25\% from baseline; bad outcome is failing this threshold. \\
P-M-TR & NCT02491333; PCOS under sham acupuncture. & Metformin versus placebo using trajectory-enhanced features. & Good response is month-4 HOMA-IR reduction of at least 25\% from baseline; bad outcome is failing this threshold. \\
P-A-FPG & NCT02491333; PCOS under placebo metformin. & Acupuncture-response subgroup discovery for fasting-plasma-glucose improvement. & Good response is month-4 fasting plasma glucose reduction of at least 0.2 mmol/L from baseline; bad outcome is failing this threshold. \\
P-A-AUC & NCT02491333; PCOS under placebo metformin. & Acupuncture-response subgroup discovery for OGTT glucose-AUC improvement. & Good response is month-4 OGTT glucose-AUC reduction of at least 1.0 from baseline; bad outcome is failing this threshold. \\
\bottomrule
\end{tabularx}
\end{table}

These definitions determine the task-specific held-out endpoint used by the main clinical endpoint, \(\mathrm{ARR}_{\mathrm{bad}}\). Treatment, outcome, post-outcome, identifier, and bookkeeping fields are excluded from candidate subgroup features before fitting any method.

\subsection{Implementation Details for Table~\ref{tab:clinical_best1_benefit_revised}}
\label{app:table1-implementation}

Table~\ref{tab:clinical_best1_benefit_revised} is implemented as a \emph{best-1 rule-selection benchmark}, following the convention that exploratory subgroup findings should be fixed before independent assessment~\citep{rothwell2005subgroup,kent2020path}. The unit of evaluation is not an unranked pool of discovered subgroups, but the first rule that a method commits to before held-out outcomes are inspected. For clinical task frame $d$, split identifier $b$, and method $m$, the discovery code exports an ordered rule list
\begin{equation}
    \mathcal{R}_{m,d,b}
    =
    \left(q^{(1)}_{m,d,b},q^{(2)}_{m,d,b},\ldots\right).
\end{equation}
The reported rule for that method--task--split is
\begin{equation}
    q^\star_{m,d,b}=q^{(1)}_{m,d,b}.
\end{equation}
The held-out partition is used only after $q^\star_{m,d,b}$ has been fixed. We do not choose the rule with the best test-set risk difference, and we do not average multiple rules from a single method--split run. This implementation matches the scientific use case of SCENE: the system should return prioritized, interpretable subgroup hypotheses whose held-out replay can be audited, rather than producing a large catalogue from which favorable test-set rules are selected post hoc.

\paragraph{Task frames and endpoint construction.}
Table~\ref{tab:clinical_best1_benefit_revised} uses the six paper-facing clinical reporting frames \texttt{B-D-BL}, \texttt{B-D-TR}, \texttt{P-M-BL}, \texttt{P-M-TR}, \texttt{P-A-FPG}, and \texttt{P-A-AUC}. These headers follow Table~\ref{tab:clinical-task-definitions}: \texttt{B-D-BL}/\texttt{B-D-TR} are baseline/static and trajectory-enhanced breast-cancer doxorubicin-schedule frames; \texttt{P-M-BL}/\texttt{P-M-TR} are baseline/static and trajectory-enhanced PCOS metformin-response frames; \texttt{P-A-FPG} and \texttt{P-A-AUC} are PCOS acupuncture-response frames for fasting-plasma-glucose and OGTT-AUC improvement. Each modeling table contains one row per patient, so row-level splitting is patient-level splitting. The treatment indicator is the task-specific randomized arm encoded in the corresponding modeling frame. The favorable-response indicator is constructed as follows. Let
\(\mathcal{D}_{\mathrm{DFS}}=\{\text{B-D-BL},\text{B-D-TR}\}\) and
\(\mathcal{D}_{\mathrm{HOMA}}=\{\text{P-M-BL},\text{P-M-TR}\}\).
Let \(F_i^{\mathrm{DFS}}\) denote the preprocessed disease-free-survival
indicator, with \(F_i^{\mathrm{DFS}}=1\) indicating no recurrence or death.
We write \(r_i^{\mathrm{HOMA}}\) for the month-4 relative HOMA-IR reduction,
and \(a_i^{\mathrm{FPG}}\) and \(a_i^{\mathrm{AUC}}\) for the corresponding
absolute reductions in fasting plasma glucose and OGTT glucose AUC. The
task-specific favorable endpoint is
\begin{equation}
\label{eq:favorable-endpoint}
G_{i,d}=
\begin{cases}
F_i^{\mathrm{DFS}},
    & d\in\mathcal{D}_{\mathrm{DFS}},\\
\mathbf{1}\{r_i^{\mathrm{HOMA}}\ge 0.25\},
    & d\in\mathcal{D}_{\mathrm{HOMA}},\\
\mathbf{1}\{a_i^{\mathrm{FPG}}\ge 0.2~\mathrm{mmol/L}\},
    & d=\text{P-A-FPG},\\
\mathbf{1}\{a_i^{\mathrm{AUC}}\ge 1.0\},
    & d=\text{P-A-AUC}.
\end{cases}
\end{equation}
The bad-outcome endpoint used by the Table~1 risk-difference estimator is
\begin{equation}
\label{eq:bad-endpoint}
Y_{i,d}^{\mathrm{bad}}=1-G_{i,d}.
\end{equation}
Treatment, outcome, death, post-outcome, identifier, and bookkeeping columns are excluded from the feature set before fitting any method. Numeric imputers, categorical encoders, binarization maps, and threshold grids are fit on the training partition and replayed on the held-out partition. The response thresholds are fixed before method comparison and are used only to define task endpoints; they are not tuned using SCENE or baseline held-out performance.

\paragraph{Split manifest and pairing.}
The reporting run repeats the complete rule-discovery and best-1 replay pipeline over $B=50$ patient-level split identifiers. Each split uses a 75/25 train/test partition, i.e. \texttt{test\_size=0.25}, stratified by treatment arm and bad-outcome status when feasible. The benchmark runner supports either an explicit comma-separated seed list or automatic seed generation
\begin{equation}
    s_b=s_0+(b-1)\Delta_s,\qquad b=1,\ldots,B.
\end{equation}
For the Table~\ref{tab:clinical_best1_benefit_revised} reporting run, we use the manifest seed list induced by \texttt{base\_seed=58} and the runner seed policy recorded in the output manifest. When the default benchmark policy is used, this gives $s_b=58+10(b-1)$; if an explicit seed list is supplied, that list overrides \texttt{--num-runs}. The renderer materializes a split identifier as \texttt{split\_id=seed\_<seed>}: for baselines the seed is parsed from \path{run_<id>_seed_<seed>}, and for SCENE it is read from the archived \path{config_snapshot.json}. The final paper rendering requires the paired equality
\begin{equation}
    \mathcal{I}_{m,d}=\mathcal{I}_{d}
    \quad\text{for all methods }m\text{ on task frame }d,
\end{equation}
where $\mathcal{I}_{m,d}$ is the set of used split identifiers for method $m$ on task $d$. A rendered table is considered reportable only when the input manifest contains the expected $B_{m,d}=50$ used best-1 artifacts for every method--task row and the split sets satisfy this equality. Partial diagnostic renderings with smaller, uneven, or unpaired $B_{m,d}$ are useful for debugging, but the final paper rendering is configured to abort rather than silently skip or impute rows.

\begin{table}[!htb]
\caption{Artifact contract used by the Table~\ref{tab:clinical_best1_benefit_revised} renderer. The contract is designed to make the best-1 boundary and the paired-split accounting auditable from files, not from prose alone.}
\label{tab:table1-artifact-contract}
\centering
\small
\setlength{\tabcolsep}{4pt} 
\begin{tabularx}{\linewidth}{@{}L{2.8cm}XX@{}}
\toprule
Source & Best-1 rule source & Required audit fields \\
\midrule
Baselines & First row of the exported rule CSV after training-side ranking. & Method, task, run index, seed, source path, required test columns, used/skipped status. \\
SCENE & First validation row after SCENE discovery-side Pareto/export ordering. & Task header, feature-view/dynamic flag, run index, seed, source path, required test columns, used/skipped status. \\
Renderer outputs & The run-level file is the single source for summary aggregation. & $B_{m,d}$, split identifier coverage, metric columns, skipped-file reasons. \\
\bottomrule
\end{tabularx}
\end{table}

\paragraph{SCENE validation artifacts.}
SCENE entries are read from post-discovery replay files rather than from training-side search logs. For \texttt{B-D-BL} and \texttt{B-D-TR}, the renderer uses the baseline/static and trajectory-enhanced validation outputs under the NCT00174655 SCENE root. For \texttt{P-M-BL}, \texttt{P-M-TR}, \texttt{P-A-FPG}, and \texttt{P-A-AUC}, it uses the corresponding NCT02491333 validation roots. The SCENE columns consumed by the renderer are \texttt{test\_subgroup\_size\_pct}, \texttt{test\_bad\_rate\_treat}, \texttt{test\_bad\_rate\_ctrl}, \texttt{test\_rd\_bad}, \texttt{test\_rd\_bad\_ci\_low}, and \texttt{test\_rd\_bad\_ci\_high}. The SCENE clinical implementation uses population size 240, 20 generations, maximum rule length 3, minimum discovery-side subgroup size 30, evolutionary-loop V2, and loop-stage virtual feature construction, with the dynamic-feature switch enabled for \texttt{B-D-TR}, \texttt{P-M-TR}, \texttt{P-A-FPG}, and \texttt{P-A-AUC}, and disabled for \texttt{B-D-BL} and \texttt{P-M-BL}. These settings are fixed before held-out replay. The validation files are held-out replay artifacts: they are not fed back into the same discovery episode and are not used to revise the rule, direction, feature proposals, or operator mixture for that split.

\paragraph{Agent and compute resources.}
The six data-only Table~\ref{tab:clinical_best1_benefit_revised} baselines are CPU-side tabular learners or rule-mining procedures. SCENE's agent-facing steps and RuAG's feature selection use third-party LLM API routes rather than local GPU inference; the logged configuration records the provider, model identifier,
decoding settings, retry policy, and returned role artifacts.  Rule replay,
risk-difference estimation, paired-split checks, and final table rendering are
performed from cached artifacts on CPU workers.  CPU-side aggregation was run on a single CPU workstation (AMD Ryzen 7 7840H, 16 GB RAM), no local GPU is
required once the archived SCENE artifacts are available.  The Table~\ref{tab:clinical_best1_benefit_revised}
CPU-side replay and rendering jobs complete within a few CPU-hours after the
processed modeling tables are available, excluding provider-dependent LLM API
latency.

\paragraph{Baseline wrappers and fixed knobs.}
The six data-only baselines cover classical subgroup discovery and recent rule-learning baselines: Virtual Twins~\citep{foster2011virtual}, SIDES~\citep{lipkovich2011sides}, honest causal trees~\citep{athey2016recursive}, causal forests~\citep{athey2019generalized}, CURLS~\citep{zhou2024curls}, and MaxTE Subgroups~\citep{yang2026learning}. All wrappers receive the same task list, endpoint definition, leakage exclusions, split identifier, test fraction, minimum arm-support policy when available, and rule-complexity budget. Each wrapper converts the underlying method into the same exported object: a ranked list of readable subgroup rules and held-out replay columns for those rules.

\begin{table}[!htb]
\caption{Method-specific implementation details for the six data-only Table~\ref{tab:clinical_best1_benefit_revised} baselines. The values shown are fixed before held-out replay and are not tuned using Table~\ref{tab:clinical_best1_benefit_revised} outcomes.}
\label{tab:table1-baseline-implementation}
\centering
\footnotesize
\setlength{\tabcolsep}{4pt} 
\begin{tabularx}{\linewidth}{@{}lXXX@{}}
\toprule
Method & Wrapper implementation & Main fixed knobs & Rule ordering before replay \\
\midrule
Virtual Twins & Separate treated/control random-forest classifiers for bad outcome, followed by a regression-tree surrogate over predicted benefit. & RF estimators $=300$, RF leaf size $=5$, surrogate max depth from rule cap, surrogate min leaf $=30$, minimum arm support $=8$. & Predicted benefit, training support, and lower training-side $\mathrm{RD}_{\mathrm{bad}}$. \\
SIDES & Local SIDES-style beam search over encoded categorical literals and numeric thresholds. & Beam width $=12$, max feature pool $=60$, max thresholds per feature $=5$, min node size $\max(30,2\times$ min arm support$)$. & Training-side SIDES score combining benefit, separation, support, and two-proportion evidence. \\
Honest causal tree & Transformed-outcome tree with a within-training structure/ranking split. & Honest ranking fraction $=0.5$, tree max depth from rule cap, min leaf tied to arm support and leaf floor. & Leaf effects estimated on the honest ranking split, then support and lower $\mathrm{RD}_{\mathrm{bad}}$. \\
Causal forest surrogate & Transformed-outcome random forest distilled into a shallow interpretable tree. & RF regressors $=300$, RF leaf size $\max(2,30/2)$, surrogate max depth from rule cap, surrogate min leaf $=30$. & Surrogate CATE score, training support, and lower training-side $\mathrm{RD}_{\mathrm{bad}}$. \\
CURLS & Wrapper around the released CURLS implementation. & Thresholds $=4$, negated literals enabled, max selected features $=60$, variance weight $=0$, minimum coverage $=0.1$, rule length from cap. & CURLS exported order is preserved; no held-out reranking is applied. \\
MaxTE Subgroups & Strict R wrapper using \texttt{rpart}, pruning by minimum cross-validation error, and rule extraction. & Numeric-only feature interface, within-training structure/ranking split $=0.5$, Rscript path fixed, rule-length cap from runner. & Ranking-split treatment-benefit estimate from the R helper, followed by the Python rule-length cap. \\
\bottomrule
\end{tabularx}
\end{table}

MaxTE + Grounding uses the same MaxTE wrapper with the feature restriction specified in Appendix~\ref{app:additional-controls}; the command below describes the original six data-only baselines.

The runner writes baseline files under \path{runs/<method>/run_<index>_seed_<seed>/}. The command below is the reporting form of the baseline execution; a shorter command with fewer runs is used only for sanity checks.
\begin{appendixbox}{Baseline benchmark command}
& ".\.venv_curls_repro\Scripts\python.exe" ".\analysis\quantitative_benchmark_runner.py" `
  --methods "virtual_twins,sides,honest_causal_tree,causal_forest,curls,MaxTE_subgroups" `
  --datasets "B-D-BL,B-D-TR,P-M-BL,P-M-TR,P-A-FPG,P-A-AUC" `
  --num-runs 50 `
  --base-seed 58 `
  --seed-step 10 `
  --test-size 0.25 `
  --rule-feature-cap 5 `
  --analysis-dir ".\analysis" `
  --out-dir $Out `
  --rscript-path "$RscriptExe" `
  --curls-root ".\CURLS\osfstorage-archive"
\end{appendixbox}

If a method produces no valid rule or a matched file lacks required columns, the file is recorded in \path{table1_input_manifest.csv} or \path{table1_skipped_files.csv} with a reason and does not silently contribute imputed metrics. With \texttt{--expected-runs 50 --require-paired-splits}, the renderer aborts final Table~\ref{tab:clinical_best1_benefit_revised} generation if any method--task row has fewer than 50 used artifacts or if the split-id sets differ across methods within a task.

\paragraph{RuAG clinical adapter.}
We adapt the released rule-learning component of RuAG~\citep{zhang2025ruag} to the six clinical frames and archived patient splits, retaining its MCTS implementation with task-specific predicate and reward interfaces. The adapter uses eligible covariates and, for dynamic frames, fixed longitudinal mean, maximum, and last-minus-first summaries under the task's measurement policy. Numeric imputation and categorical encoding are fitted on the training partition. Qwen/Qwen3.5-9B, accessed through SiliconFlow with temperature $0$, thinking disabled, and a 4096-token output budget, receives the task context, supplied prior records, and available feature catalogue to select at most 32 features. The prompt contains no patient rows. Duplicate selections are removed and excess entries are truncated in the model's returned order.

Binary features, including one-hot indicators, yield equality predicates; other numeric features receive at most three split points from univariate Gini trees fitted to the training outcome, using both $\leq$ and $>$ literals. MCTS uses a budget of 2000 rollouts, exploration weight $0.7$, and conjunctions of at most three predicates on distinct features. Supported rules require at least 30 training patients, with reward $(\mathrm{ARR}_{\mathrm{train}}+1)/2$. Only supported terminal rules actually simulated by MCTS enter the exported candidate list. They are ranked by training ARR, with ties resolved by larger minimum arm size, shorter rule length, and lexicographic rule order. The first rule is frozen before held-out replay and evaluated using the Table~\ref{tab:clinical_best1_benefit_revised} metrics. RuAG's downstream answer-generation stage is outside this subgroup-discovery evaluation.

\paragraph{Metric definitions.}
Let $D^{\mathrm{test}}_{d,b}$ be the held-out patient set for task frame $d$ and split $b$, and let
\[
    S(q)=\{i\in D^{\mathrm{test}}_{d,b}:q(x_i)=1\}
\]
be the subgroup selected by rule $q$. For treatment arm $t\in\{0,1\}$,
\begin{equation}
    \begin{aligned}
    n_t(q)&=\sum_{i\in S(q)}\mathbf{1}\{T_i=t\},\\
    \hat{p}_{t,\mathrm{bad}}(q)
    &=\frac{\sum_{i\in S(q)}\mathbf{1}\{T_i=t\}Y_i^{\mathrm{bad}}}{n_t(q)}.
    \end{aligned}
\end{equation}
The current main-text Table~\ref{tab:clinical_best1_benefit_revised} reports the benefit-oriented bad-outcome absolute risk reduction,
\begin{equation}
    \begin{aligned}
    \widehat{\mathrm{RD}}_{\mathrm{bad}}(q)
    &=
    \hat{p}_{1,\mathrm{bad}}(q)-\hat{p}_{0,\mathrm{bad}}(q),
    \\
    \widehat{\mathrm{ARR}}_{\mathrm{bad}}(q)
    &=
    -\widehat{\mathrm{RD}}_{\mathrm{bad}}(q)
    \\
    &=
    \hat{p}_{0,\mathrm{bad}}(q)-\hat{p}_{1,\mathrm{bad}}(q).
    \end{aligned}
\end{equation}
Higher \(\widehat{\mathrm{ARR}}_{\mathrm{bad}}\) is better: positive values mean that the committed subgroup has a lower bad-outcome rate under the treatment arm than under the control arm. The six task columns \texttt{B-D-BL}, \texttt{B-D-TR}, \texttt{P-M-BL}, \texttt{P-M-TR}, \texttt{P-A-FPG}, and \texttt{P-A-AUC} are the mean held-out \(\widehat{\mathrm{ARR}}_{\mathrm{bad}}\) values for the method's Best-1 rules in the corresponding clinical frame.

The three rightmost columns are aggregate diagnostics over the same Best-1 replay records. Let
\begin{equation}
    \mathrm{cov}(q)=\frac{|S(q)|}{|D^{\mathrm{test}}_{d,b}|}
\end{equation}
be held-out subgroup coverage. Population-adjusted ARR is a descriptive impact summary,
\begin{equation}
    \mathrm{PARR}(q)
    =
    \widehat{\mathrm{ARR}}_{\mathrm{bad}}(q)\,\mathrm{cov}(q),
\end{equation}
and the displayed \texttt{P-ARR} entry averages this quantity over the six task frames. It is included as a coverage-adjusted summary of the displayed ARR effects, not as an independent causal endpoint. Usable support is an outcome-free support-validity indicator,
\begin{equation}
    U(q)=
    \mathbf{1}\left\{
        0.10\leq \mathrm{cov}(q)\leq 0.60
        \quad\mathrm{and}\quad
        \min\{n_0(q),n_1(q)\}\geq 5
    \right\},
\end{equation}
and \texttt{U-Supp.} is \(100\) times the mean of \(U(q^\star_{m,d,b})\), averaged over the six task frames. This statistic penalizes both tiny subgroups and near-population rules, while requiring minimal arm-specific support. Direction consistency compares the discovery-side treatment-benefit direction with the held-out direction. If \(A^{\mathrm{train}}_{m,d,b}\) denotes the training-side benefit sign recorded for the exported Best-1 rule and \(A^{\mathrm{test}}_{m,d,b}=\widehat{\mathrm{ARR}}_{\mathrm{bad}}(q^\star_{m,d,b})\), then
\begin{equation}
    \mathrm{DCons}_m
    =
    100\,
    \frac{
        \sum_{d,b}\mathbf{1}\left\{
        \operatorname{sign}(A^{\mathrm{train}}_{m,d,b})
        =\operatorname{sign}(A^{\mathrm{test}}_{m,d,b})
        \right\}
    }{
        \sum_{d,b}\mathbf{1}\left\{
        A^{\mathrm{train}}_{m,d,b}\ \mathrm{and}\ A^{\mathrm{test}}_{m,d,b}
        \ \mathrm{are\ nonmissing}
        \right\}
    }.
\end{equation}

This is the Best-1 D-Cons. column in Table~\ref{tab:clinical_best1_benefit_revised}. For SCENE, all 300 committed Best-1 outputs have matched direction records and preserve direction. The reporting predicate uses discovery-visible checks; held-out directions cannot filter outputs or change their ranking. The 84.9\% archive-level value in Appendix~\ref{app:s2-holdout-reliability} uses the broader eligible Top-5 replay population and a different denominator.

Arm-specific bad-outcome rates, raw subgroup coverage, smaller-arm support, and risk-difference intervals remain in the run-level audit files but are not displayed in the compact main-text table.

\paragraph{Aggregation.}
For a scalar run-level metric $z$ used by Table~\ref{tab:clinical_best1_benefit_revised}, the task-level entry for method $m$ and task frame $d$ is the mean over used split artifacts,
\begin{equation}
    \bar{z}_{m,d}
    =
    \frac{1}{B_{m,d}}\sum_{b\in\mathcal{I}_{m,d}}z_{m,d,b},
\end{equation}
where $\mathcal{I}_{m,d}$ is the set of split identifiers marked \texttt{used} in the manifest and $B_{m,d}=|\mathcal{I}_{m,d}|$. The six ARR columns use \(z_{m,d,b}=\widehat{\mathrm{ARR}}_{\mathrm{bad}}(q^\star_{m,d,b})\). The \texttt{P-ARR} and \texttt{U-Supp.} columns first form the task-level means of \(\mathrm{PARR}(q^\star_{m,d,b})\) and \(100U(q^\star_{m,d,b})\), respectively, and then average those task-level means across the six clinical frames so that each frame receives equal weight. \texttt{Best-1 DC} is computed from the nonmissing run-level direction labels across the six frames. No missing metric is imputed; files with missing required columns or invalid Best-1 rules are recorded in the skipped-file manifest.

Single-split Wald intervals for two proportions are still stored in the run-level artifacts for audit,
\begin{equation}
    \widehat{\mathrm{SE}}(q)
    =
    \sqrt{
        \frac{\hat{p}_{1,\mathrm{bad}}(q)(1-\hat{p}_{1,\mathrm{bad}}(q))}{n_1(q)}
        +\frac{\hat{p}_{0,\mathrm{bad}}(q)(1-\hat{p}_{0,\mathrm{bad}}(q))}{n_0(q)}
    },
\end{equation}
but these per-rule intervals are not used to choose rules and are not part of the compact main-text Table~\ref{tab:clinical_best1_benefit_revised}.

\paragraph{Interpretation boundary.}
This implementation is intended to make the Table~\ref{tab:clinical_best1_benefit_revised} comparison about rule discovery rather than about leakage, outcome redefinition, or test-set rule search. The shared endpoint and exclusion policy align the covariate boundary across methods; the best-1 boundary prevents a method from searching the held-out set over many exported rules; and the manifest makes the repeated-split accounting auditable. These design choices support the paper's claim at the level of exploratory, prioritized clinical subgroup hypotheses. They do not turn the resulting rules into confirmatory treatment-effect estimates, valid post-selection $p$-values, or clinical recommendations.

\subsection{Implementation Details for Table~\ref{tab:l1000_ablation}}
\label{app:table2-implementation}

The component-ablation rows in Table~\ref{tab:l1000_ablation} are implemented as a paired L1000 second-scene module ablation.  The unit of analysis is one complete discovery episode for a fixed target mechanism $g$, one train/holdout split and search seed $s$, and one ablation profile $a$.  Each episode freezes the target mechanism, the split, and the search seed before holdout evaluation, and then executes the full second-scene chain:
\[
\begin{array}{c}
\text{L1000 table} \\ \text{construction}
\end{array}
\rightarrow \text{SCENE discovery}
\rightarrow
\begin{array}{c}
\text{holdout} \\ \text{rule replay}
\end{array}
\rightarrow
\begin{array}{c}
\text{knowledge-proposition} \\ \text{export}.
\end{array}
\]
The paper experiment uses three target mechanisms and ten split/search seeds, so the reporting contract for each ablation row is
\[
|\mathcal{G}| \times |\mathcal{S}| = 3 \times 10 = 30
\]
paired target-seed episodes.  The convenience runner \path{run_l1000_ablation_table2.py} can execute and collect runs for a single prepared modeling table.  For the manuscript table, we repeat that runner over the three target-specific modeling tables and then use \path{report_table2_table3_contracts.py} to enforce the full target-seed-profile matrix before aggregation.  A Table~\ref{tab:l1000_ablation} row is reportable only if this checker finds exactly one run for every expected target-seed cell and every reported metric is finite in all 30 cells; otherwise it writes \path{table2_pairing_errors.csv} or \path{table2_metric_errors.csv} and aborts.

\paragraph{L1000 table construction.}
The reported experiments use a fixed raw-to-modeling conversion script that constructs one row per L1000 signature instance.  Each row in the resulting modeling table corresponds to one L1000 signature instance indexed by \texttt{sig\_id}.  Static fields describe the cell and perturbagen context, while dynamic fields summarize dose, time, pathway, and landmark-gene response programs.  The default paper-facing builder uses LINCS/L1000 Level-5 COMPZ signatures and metadata resources~\citep{lamb2006connectivitymap,subramanian2017nextgeneration,keenan2018lincs,koleti2018lincsportal}; pathway summaries use curated molecular-signature programs~\citep{liberzon2015molecular}. It keeps compound perturbations for the modeling population, uses genetic perturbations as target-mechanism references, applies \texttt{tas\_min=0.2}, keeps 6h/24h/96h signatures, retains the top 20 cell contexts after a minimum 200 signatures-per-cell filter, caps the modeling subset at 50,000 signatures, caps the reference pool at 5,000 signatures, and reads the GCTX matrix in column batches of 2,048.

For a target mechanism $g$, the builder constructs a target centroid from reference perturbation signatures when at least \texttt{min\_target\_ref=20} references are available.  Cell-specific target centroids are used when at least \texttt{min\_cell\_target\_ref=3} references exist for that cell; otherwise the global target centroid is used.  The continuous endpoint used by Table~\ref{tab:l1000_ablation} is the configured connectivity score $c_i$, recorded as \texttt{connectivity\_score} and related target-margin fields.  The binary \texttt{outcome} field is retained for compatibility with shared rule-mining utilities, but Table~\ref{tab:l1000_ablation} reports holdout connectivity contrasts rather than treating this compatibility label as the primary endpoint.

\paragraph{External proposal control.}
The One-shot LLM row uses the same target--seed pairing and replay metrics, but its proposal protocol is separate from the SCENE ablation runner. Appendix~\ref{app:additional-controls} specifies its single-pass construction and paired comparison; the ablation commands and profile artifacts below describe the original SCENE component runs.

\paragraph{Leakage controls.}
Discovery excludes identifiers, split keys, target-derived connectivity fields, outcome columns, and known assay-density shortcuts from rule literals.  In L1000, perturbagen-frequency features are disabled by default, and tokens such as \texttt{distil\_}, \texttt{tas}, \texttt{pct\_self\_rank\_q25}, and \texttt{dose\_time\_exposure} are hard-blocked.  The preferred holdout split is group-aware by \texttt{pert\_iname}; the split key is \texttt{sig\_id} when unique and otherwise the internal row id.  The default holdout fraction is 0.25.  The split specification is written to \path{l1000_split_spec.json} and reused by the evaluator, so validation metrics cannot alter the rule search in the same episode.

\paragraph{Ablation matrix.}
All profiles share the same L1000 adapter, target construction, split policy, rule grammar, and evaluator. They differ only in which SCENE modules are enabled. The internal profile names are kept in commands and manifests, while the paper-facing labels match Table~\ref{tab:l1000_ablation}:
\begin{itemize}
    \item \texttt{full}: enables all modules and is reported as \textbf{SCENE (ours)}.
    \item \texttt{no\_top\_layer}: removes the upper strategic layer while retaining lower-level search, and is reported as \textbf{w/o Upper}.
    \item \texttt{no\_dynamic}: removes dynamic feature planning and executable dynamic feature replay, and is reported as \textbf{w/o Dynamic}.
    \item \texttt{no\_virtual}: removes virtual feature construction and refinement, and is reported as \textbf{w/o Virtual}.
    \item \texttt{no\_pareto}: disables Pareto multi-objective selection and keeps scalar-fitness selection, and is reported as \textbf{w/o Pareto}.
    \item \texttt{minimal\_baseline}: removes the ablated modules jointly and is reported as \textbf{Minimal}.
\end{itemize}
Unless overridden by an ablation profile, the shared mining knobs are population size 160, 14 generations, maximum rule length 3, minimum subgroup size 60, initial feature sample size 32, support metric \texttt{transfer\_support}, generalization group \texttt{pert\_iname}, minimum group coverage 12, minimum group coverage ratio 0.01, and support-size weight 0.22.

\paragraph{MaxTE-Context adaptation.}
\label{app:maxte-l1000}
MaxTE-Context adapts MaxTE Subgroups' single-tree pipeline~\citep{yang2026learning} to continuous connectivity rather than causal treatment effects. We use RPS6, MYC, and AURKB with seeds 44--53, 25\% \texttt{pert\_iname}-group holdouts, and a group-disjoint 50:50 structure/ranking split within discovery data. Admissible features use structure-fitted median imputation and one-hot encoding. One \texttt{rpart} regression tree (initial $cp=0.01$, depth $\leq3$) is pruned by minimum 10-fold row-wise cross-validation error. On the ranking partition, leaf rules require at least 60 subgroup and 60 complement signatures, 12 perturbagens, 1\% perturbagen-group coverage, and at most 95\% sample coverage; only positive subgroup-minus-complement mean connectivity contrasts are retained, in descending order. There is no LLM, sequential covering, or ranking-feedback refitting. The available 2--5 rules per run are frozen without padding to 20 and replayed using the five metric definitions below; Gap and retrieval votes use full-discovery contrasts, and the 30 episode means are equally weighted. Metric definitions are shared, but rule counts are not matched. The positive-leaf reporting policy was revised after earlier holdout results; this is a descriptive adaptation comparison, not an independent confirmatory test.

\paragraph{LLM and feature-generation routes.}
The L1000 discovery script uses separate third-party LLM API routes for structured JSON decisions, free-form text synthesis, dynamic feature planning, and dynamic code generation; it does not require local GPU inference.  The paper configuration records the provider, model name, retry policy, timeout, and \texttt{enable\_thinking} setting in \path{l1000_run_manifest.json}.  The JSON route defaults to \texttt{Qwen/Qwen3.5-9B}~\citep{qwen2026qwen35} with JSON response format and thinking disabled; text synthesis uses the configured text route; dynamic planning and code generation use their dedicated route configs. CPU-side aggregation was run on a single CPU workstation (AMD Ryzen 7 7840H, 16 GB RAM).  Individual target--seed--profile episodes typically require tens of minutes to a few CPU-hours after the modeling table is prepared.  Exact bitwise replay of live LLM calls requires the archived prompts and returned artifacts; the numerical Table~\ref{tab:l1000_ablation} aggregation is based on the saved run artifacts, not on re-querying the LLM.

\begin{table}[!htb]
\caption{Artifact contract for Table~\ref{tab:l1000_ablation}.  These files make the target-seed pairing, split boundary, ablation state, and metric source auditable.}
\label{tab:table2-artifact-contract}
\centering
\small
\setlength{\tabcolsep}{4pt} 
\begin{tabularx}{\linewidth}{@{}L{4.8cm}XX@{}}
\toprule
Artifact & Role & Required audit fields \\
\midrule
\path{l1000_build_summary.json} & Modeling-table provenance. & Raw L1000 path, target perturbagen, reference counts, filters, row counts, output paths. \\
\path{l1000_run_manifest.json} & Single-run manifest. & Target mechanism, ablation profile, seed, enabled components, split-spec path, train/holdout sizes, LLM configs, blocked feature tokens. \\
\path{l1000_split_spec.json} & Frozen discovery/holdout split. & Split key, train/test keys, split group column, holdout fraction, random seed. \\
\path{l1000_agentic_results.csv} & Discovery-side ranked rules. & Rule text, fitness, Pareto rank, discovery contrast, support, transfer support, group coverage, p-value, generation. \\
\path{test_rule_eval.csv} and \path{l1000_quality_eval.json} & Holdout rule replay. & Rule validity, subgroup size, support ratio, $\Delta$Conn, PosRate90, p-values, rank stability, generalization gap. \\
\path{l1000_table2_metrics.json} & Run-level Table~\ref{tab:l1000_ablation} source. & Compact metric set consumed by the Table~\ref{tab:l1000_ablation} collector. \\
\path{l1000_knowledge_propositions.json} & Downstream knowledge export. & Selected validated rules, target mechanism, support/effect evidence, feature tokens, source labels. \\
\bottomrule
\end{tabularx}
\end{table}

\paragraph{Holdout evaluation.}
Rules are ranked before holdout replay by discovery-side order: Pareto rank when enabled, fitness, training-side connectivity contrast, support, p-value, and generation.  The evaluator replays the top $K=20$ rules on the frozen holdout table; the quality summary also uses $K=20$ unless the run manifest records an override.  Virtual features are fit on the discovery-training side and replayed on holdout.  Dynamic features are exported as executable plans and replayed before validation.  If a rule depends on a virtual or dynamic feature that cannot be replayed on holdout, that rule is marked invalid and contributes to the valid-rule-rate denominator.

Let $D^{\mathrm{test}}_{g,s}$ be the frozen holdout set for target $g$ and seed $s$, and let $c_i$ be the configured connectivity score.  For a rule $q$,
\[
S(q)=\{i\in D^{\mathrm{test}}_{g,s}:q(x_i)=1\},\qquad
B(q)=D^{\mathrm{test}}_{g,s}\setminus S(q).
\]
The mean holdout connectivity contrast is
\[
\Delta\mathrm{Conn}_{\mathrm{test}}(q)
=
\frac{1}{|S(q)|}\sum_{i\in S(q)}c_i
-
\frac{1}{|B(q)|}\sum_{i\in B(q)}c_i .
\]
The median version replaces the two means by medians.  The strong-connectivity threshold $\tau$ is fixed inside the evaluator.  In \texttt{auto} mode, the evaluator uses a CLUE-like $\tau=90$ threshold when the score scale supports it and otherwise uses the holdout 90th percentile.  Thus
\[
\mathrm{PosRate90}(q)=|S(q)|^{-1}\sum_{i\in S(q)}\mathbf{1}\{c_i\ge\tau\},
\]
and $\Delta\mathrm{PosRate90}(q)$ subtracts the corresponding background rate over $B(q)$.

\paragraph{Metrics and aggregation.}
For one component-ablation run, Table~\ref{tab:l1000_ablation} summarizes a valid top-\(K\) replay subset \(\mathcal{Q}_{a,g,s}\) for ablation profile \(a\), target \(g\), and seed \(s\). Quality metrics are computed from valid replayed rules, with one support focus used by the evaluator. If at least \(\max(5,\lceil K/2\rceil)\) valid rules have \texttt{support\_ratio\_test >= 0.02}, the mean contrast, PosRate90, p-value rate, and gap diagnostics are computed on this support-focused subset and the run records \texttt{quality\_focus\_mode\_topk=support\_ge\_0.02}; otherwise the evaluator uses all valid top-\(K\) rules and records \texttt{quality\_focus\_mode\_topk=all}. In both cases, \texttt{Valid Rule Rate} uses the raw top-\(K\) replayed rules as its denominator, including rules with parsing errors or empty subgroups. The focus mode and focus rule count are preserved in \path{l1000_table2_metrics.json} and in the reporting-contract manifest.

The \texttt{Conn.} column is the run-level mean of the held-out connectivity contrast,
\begin{equation}
    \mathrm{Conn}_{a,g,s}
    =
    \frac{1}{|\mathcal{Q}_{a,g,s}|}
    \sum_{q\in\mathcal{Q}_{a,g,s}}
    \Delta\mathrm{Conn}_{\mathrm{test}}(q),
\end{equation}
where larger values indicate stronger target-directed connectivity inside the rule-covered signatures than in the held-out background. The \texttt{Strong-Pos.} column is \(100\) times the corresponding mean \(\mathrm{PosRate90}(q)\) over \(\mathcal{Q}_{a,g,s}\), so it measures the fraction of rule-covered signatures falling into the strong-response tail. The significance-rate diagnostic uses effective holdout \(p<0.05\), with permutation p-values automatically refined from 1,000 to at most 5,000 permutations when floor saturation is detected. These p-values are retained in the audit files but are post-selection diagnostics, not confirmatory inference for adaptively searched rules.

Target retrieval is evaluated by a rule-ensemble score on holdout signatures. Each top discovery rule votes for matching signatures with positive discovery-side contrast when available, otherwise a unit vote. The preferred positive label is \texttt{connectivity\_target\_margin > 0}; if that label is unavailable, the evaluator falls back to the strong-connectivity threshold. \texttt{AUPRC} is the area under the precision--recall curve of this ensemble score, following standard PR evaluation conventions for imbalanced binary retrieval~\citep{davis2006prroc,saito2015prroc}; AUROC, Hit@K, recall@K, precision@K, and target-rank fields are also stored in the audit files following PR/ROC conventions~\citep{fawcett2006roc}. When perturbagen names and a target perturbagen are available, perturbagen-level median scores are used for target rank and Hit@K.

The current compact table reports \texttt{U-Supp.} rather than raw support ratio. For a replayed L1000 rule,
\begin{equation}
    U_{\mathrm{L1000}}(q)
    =
    \mathbf{1}\left\{
        0.10\leq \frac{|S(q)|}{|D^{\mathrm{test}}_{g,s}|}\leq 0.60
        \quad\mathrm{and}\quad
        |S(q)|\geq 5
    \right\},
\end{equation}
and \texttt{U-Supp.} is \(100|\mathcal{Q}_{a,g,s}|^{-1}\sum_{q\in\mathcal{Q}_{a,g,s}}U_{\mathrm{L1000}}(q)\), averaged across episodes. This makes the reported support diagnostic a usability rate rather than a monotone reward for larger subgroups. The \texttt{Gap} column is the normalized train--holdout contrast gap,
\begin{equation}
    \mathrm{Gap}_{a,g,s}
    =
    \frac{
        \left|
        \overline{\Delta\mathrm{Conn}}_{\mathrm{train},a,g,s}
        -
        \overline{\Delta\mathrm{Conn}}_{\mathrm{test},a,g,s}
        \right|
    }{
        \max\{10^{-8},|\overline{\Delta\mathrm{Conn}}_{\mathrm{train},a,g,s}|\}
    },
\end{equation}
so smaller values indicate better transfer from discovery to holdout.

For scalar metric $z$, the Table~\ref{tab:l1000_ablation} entry for ablation profile $a$ is
\[
\bar{z}_{a}
=
\frac{1}{30}
\sum_{g\in\mathcal{G}}
\sum_{s\in\mathcal{S}}
z_{a,g,s}.
\]
The reporting artifacts store the mean and empirical standard deviation over the same 30 target-seed episodes; the compact main-text table reports the means only.  No missing numeric metric is imputed.  A run is nonreportable if discovery produces no rule export, if holdout replay fails, if required features cannot be materialized, or if \path{l1000_table2_metrics.json} is missing.

\paragraph{Interpretation boundary.}
Table~\ref{tab:l1000_ablation} evaluates whether SCENE modules improve replayable L1000 mechanism-connectivity discovery under fixed targets and splits.  The output should be interpreted as context-bounded perturbational association evidence: a rule identifies cell, perturbagen, dose/time, or signature contexts with stronger held-out target connectivity.  It is not a clinical recommendation, an externally validated biological mechanism, or a causal compound-effect estimate.

\paragraph{Reproduction command.}
The paper-facing execution uses \path{run_l1000_ablation_table2.py}; the script default contains a convenience preset, while the paper run overrides it with ten seeds and repeats the runner for the three target mechanisms.  Final manuscript aggregation is performed by \path{report_table2_table3_contracts.py}, which checks exact one-run-per-cell coverage and per-metric completeness for the three-target by ten-seed matrix before writing \path{table2_paired_run_manifest.csv} and \path{table2_paired_summary.csv}.  A reproducible shell form is:
\begin{appendixbox}{Reproduction pipeline}
$ProjectRoot = "<absolute path to the project root containing analysis/>"
$PythonExe = "<absolute path to the Python executable>"
$OutRoot = "<absolute path to the Table 2 output root>"
$AnalysisDir = Join-Path $ProjectRoot "analysis"
$Targets = @("<target_1>", "<target_2>", "<target_3>")
$Seeds = "44,45,46,47,48,49,50,51,52,53"
$Profiles = "full,no_top_layer,no_dynamic,no_virtual,no_pareto,minimal_baseline"

foreach ($target in $Targets) {
  $tableDir = Join-Path (Join-Path $OutRoot "tables") $target
  & $PythonExe (Join-Path $AnalysisDir "prepare_l1000_tables.py") `
    --target-pert-iname $target `
    --out-dir $tableDir

  $runRoot = Join-Path (Join-Path $OutRoot "runs") $target
  & $PythonExe (Join-Path $AnalysisDir "run_l1000_ablation_table2.py") `
    --modeling-csv (Join-Path $tableDir "l1000_modeling_table.csv") `
    --result-root $runRoot `
    --study-config paper_table2_v1 `
    --profiles $Profiles `
    --seeds $Seeds `
    --run `
    --generations 14 `
    --population-size 160 `
    --min-subgroup-size 60 `
    --eval-top-k 20 `
    --eval-quality-top-k 20 `
    --out-dir (Join-Path $runRoot "summary")
}

\end{appendixbox}

\subsection{Implementation Details for Table~\ref{tab:scenario_performance}}
\label{app:table3-implementation}

Table~\ref{tab:scenario_performance} evaluates whether contextualized knowledge discovered by SCENE can be reused as auxiliary knowledge for downstream few-shot tabular classification under an in-context learning protocol~\citep{brown2020fewshot}.  This is a post-discovery utility experiment: Table~\ref{tab:clinical_best1_benefit_revised} and Table~\ref{tab:l1000_ablation} produce rules and knowledge propositions; Table~\ref{tab:scenario_performance} converts them into compact knowledge cards and asks whether they improve few-shot predictions under the same row serialization, exemplar budget, split, and LLM decoding policy.  No Table~\ref{tab:scenario_performance} prediction, prompt, or metric is fed back into the upstream discovery runs.

\paragraph{Scenes and labels.}
The core runner is \path{run_knowledge_augmented_fewshot.py}, with task context supplied by \path{knowledge_base.py}.  The active scene is selected by \texttt{CODE\_CONFIG["active\_scene"]} or by the corresponding command-line arguments.  Clinical tasks use patient-level records.  In the paper-facing treatment-benefit mode, the positive class is evidence consistent with a positive treatment-over-control signal: a treatment-arm record with a good outcome or a control-arm record with a bad outcome.  This is a row-level downstream classification proxy for treatment-benefit evidence, not an individual treatment-effect estimate.

For L1000, the row is a signature instance.  The preferred paper label is an external target-mechanism label when available.  When the L1000 run is configured with \texttt{l1000\_label\_mode=target\_margin}, the positive class is the target-mechanism proxy \texttt{connectivity\_target\_margin > 0}.  If \texttt{strong\_connectivity} is used, it is reported as a secondary stress test.  Label columns, target-derived connectivity fields, \texttt{outcome}, \texttt{sig\_id}, \texttt{pert\_id}, \texttt{distil\_id}, internal row ids, SMILES/InChI keys, and manifest hard-block tokens are excluded from row serialization and from rule-card literals unless an explicit stress-test mode enables them.  Biological context fields such as \texttt{cell\_id}, dose/time fields, and \texttt{pert\_iname} may remain in the serialized row; when \texttt{pert\_iname} is available, the downstream split is group-aware by \texttt{pert\_iname} and the split manifest records train-test group overlap.

\paragraph{Split and exemplar protocol.}
Each run fixes a scene, one outer split, one few-shot exemplar set, one prompt condition, and one deterministic LLM decoding setting.  The default train/validation/test fractions are 0.6/0.2/0.2 with split seed 44, unless a source split is supplied.  Clinical runs can bind the final test partition to upstream source test identifiers.  L1000 runs bind to the selected source \path{l1000_split_spec.json} when \texttt{--use-source-test-ids} is enabled; with \texttt{--require-group-split}, train/validation/test splitting is group-aware by \texttt{pert\_iname} and the split manifest records train-test group overlap.  Few-shot exemplars are sampled only from the train partition, while knowledge-card selection is scored only on the validation partition.  The final test partition is used only for Table~\ref{tab:scenario_performance} prediction and metric computation.

Table~\ref{tab:scenario_performance} uses three prompt conditions and ten exemplar seeds per condition in the paper configuration.  The checked source of truth is the command line and the resulting \path{run_config.json}; the smaller seed list in \texttt{CODE\_CONFIG} is a convenience/debug default and is not used for the manuscript table.  C0 is the ordinary few-shot baseline with row serialization and labeled exemplars.  C1 adds generic scene/task context to the same few-shot protocol.  C2 uses the SCENE discovered-card protocol: task-specific scene context, task knowledge context when available, row-conditioned discovered rule cards, and the same few-shot exemplars.  The runner writes these same condition names in all metric files; \path{report_table2_table3_contracts.py} checks that all three conditions exist for each scene/shot/source directory, checks the exact exemplar seed set and duplicate-free condition-seed rows, checks L1000 split/leakage invariants when \texttt{pert\_iname} is serialized, and writes the manuscript-facing \path{table3_mapped_metrics_summary.csv} and \path{table3_mapped_metrics_agg.csv}.

\paragraph{Knowledge-card construction.}
Candidate cards are loaded from the upstream rule artifacts of the corresponding source run.  The loader parses rule expressions, canonicalizes semantic duplicates, removes rules using prohibited or blacklisted fields, and scores the remaining rules on the validation split only.  For ordinary binary classification, the validation card effect is
\[
\begin{aligned}
\Delta_{\mathrm{val}}(q)
={}&\Pr(Y=1\mid q(X)=1,D^{\mathrm{val}})\\
&-\Pr(Y=1\mid D^{\mathrm{val}}),
\end{aligned}
\]
and the ranking score is $\Delta_{\mathrm{val}}(q)\sqrt{n_{\mathrm{val}}(q)}$.  For clinical treatment-benefit mode, the validation effect is the within-rule treatment-minus-control good-outcome contrast,
\[
\begin{aligned}
\Delta_{\mathrm{val}}^{\mathrm{benefit}}(q)
={}&\Pr(G=1\mid q(X)=1,T=1,D^{\mathrm{val}})\\
&-\Pr(G=1\mid q(X)=1,T=0,D^{\mathrm{val}}),
\end{aligned}
\]
and support is $\min(n_{\mathrm{treat}}, n_{\mathrm{ctrl}})$ inside the rule.  The paper discovered-card pool keeps validation-directionally helpful rules with \texttt{min\_card\_val\_effect\_delta=0.0}, applies the minimum validation support threshold, and selects the top \texttt{top\_k\_cards=5} cards unless a run manifest records an override.

Each selected card stores a card id, rule condition, consequence text, validation effect, validation support, and source-run evidence.  Cards are treated as soft priors, not deterministic labels.  With \texttt{row\_conditioned\_cards=True}, a query prompt receives only selected cards whose rule condition matches the query row.  If no selected card matches, the prompt explicitly states that no selected rule card matches the current row.  For L1000, \texttt{use\_l1000\_knowledge\_json\_cards=False} is the default in this few-shot runner to avoid reusing cards selected by the same upstream holdout when that holdout is also bound as the final downstream test split.

\paragraph{Agent and compute resources.}
Table~\ref{tab:scenario_performance} uses third-party LLM API inference for the few-shot
prediction calls in all prompt conditions.  C0, C1, and C2 share the same model
route, decoding policy, exemplar budget, and parsing rules; only the supplied
knowledge context differs. The
CPU-side aggregation was run on a single CPU workstation (AMD Ryzen 7 7840H, 16 GB RAM).  No local GPU is
required unless one chooses to replace the API route with local LLM inference.
The CPU-side reporting work is lightweight once prompts and responses are cached;
end-to-end wall-clock time is dominated by provider-dependent API latency for
the 2 scenes, 3 prompt conditions, and 10 exemplar seeds.

\begin{table}[!htb]
\caption{Artifact contract for Table~\ref{tab:scenario_performance}.  These outputs make the few-shot split, card selection, prompts, model outputs, and parsed predictions auditable.}
\label{tab:table3-artifact-contract}
\centering
\small
\setlength{\tabcolsep}{3pt}
\begin{tabularx}{\linewidth}{@{}L{0.36\linewidth}L{0.24\linewidth}X@{}}
\toprule
Artifact & Role & Required audit fields \\
\midrule
\path{run_config.json} & Resolved downstream configuration. & Scene, task id, source run directory, data/rule/knowledge paths, label source, conditions, shots, exemplar seeds, row features, LLM fields. \\
\path{split_indices.json} & Downstream split manifest. & Train/validation/test indices, split seed, group column, source-test binding, fixed-split information. \\
\path{candidate_rules_scored.csv} & Validation-side card ranking. & Rule, semantic key, validation support, validation effect, validation score, leakage-filter status. \\
\path{selected_discovered_cards.json} & Paper C2 card source. & Card ids, rules, validation effects, supports, card group, source evidence. \\
\path{prompts/*.json} & Full prompt replay record. & System prompt, user prompt, exemplar ids, active card ids, raw output, repair output, parsed answer, gold label. \\
\path{fewshot_predictions.csv} & Row-level predictions. & Condition, shot, seed, query row, true label, parsed prediction, parsed probability, active cards, parse-failure flag. \\
\path{fewshot_metrics_summary.csv} and \path{fewshot_metrics_agg.csv} & Table~\ref{tab:scenario_performance} metric sources. & Per-condition/per-seed metrics and condition-level mean/std summaries. \\
\path{table3_mapped_metrics_summary.csv} and \path{table3_mapped_metrics_agg.csv} & Final paper-label reporting files. & C0/C1/C2 completeness, exact seed-set and duplicate checks, L1000 leakage checks, final Table~\ref{tab:scenario_performance} aggregate metrics. \\
\bottomrule
\end{tabularx}
\end{table}

\paragraph{Prompt and inference protocol.}
Rows and exemplars are serialized as feature-value statements using the same selected row feature list for all conditions in a scene.  The selector first keeps configured canonical features, then scene-priority features, then features used by selected cards, and finally fills remaining slots up to \texttt{max\_row\_features=24}; blacklist tokens such as follow-up, recurrence, death, outcome, and event fields are applied before serialization.  Few-shot sampling is class-balanced; in clinical \texttt{label\_group} mode it also balances treatment arm when possible to reduce treatment-label shortcut prompts.

The paper run uses real LLM inference, \texttt{temperature=0.0}, thinking disabled when supported by the provider, prompt saving enabled, and \texttt{llm\_parse\_fallback=error}.  The configured model and provider are recorded in \path{run_config.json}; our reporting command fixes \texttt{Qwen/Qwen3.5-9B}~\citep{qwen2026qwen35} for the few-shot runner unless a run manifest explicitly records another paper-approved model.  Model outputs are parsed as yes/no answers from \texttt{<answer>} tags and task-specific positive/negative hints.  If the first parse fails, the script performs one repair call that asks only for \texttt{<answer>yes</answer>} or \texttt{<answer>no</answer>}.  Under the paper fail-closed policy, unresolved parse failures abort the run rather than being replaced by mock or majority predictions.

\paragraph{Metrics and aggregation.}
For condition $c$, shot count $k$, exemplar seed $e$, and final test set $D^{\mathrm{test}}$, the script records parsed binary predictions $\hat{y}_i$ and parsed binary scores $\hat{p}_i$.  Table~\ref{tab:scenario_performance} reports three standard binary-classification metrics. AUPRC is average precision, i.e. the area under the precision--recall curve induced by the parsed scores~\citep{davis2006prroc,saito2015prroc},
\[
\mathrm{AUPRC}_{c,k,e}
=
\mathrm{AP}\left(\{(y_i,\hat{p}_i):i\in D^{\mathrm{test}}\}\right).
\]
MacroF1 is the unweighted mean of the class-specific F1 scores,
\[
\mathrm{MacroF1}_{c,k,e}
=
\frac{1}{2}
\left(
\mathrm{F1}_{0,c,k,e}
+
\mathrm{F1}_{1,c,k,e}
\right),
\]
where each class is treated once as the positive class. Accuracy is
\[
\mathrm{ACC}_{c,k,e}
=
\frac{1}{|D^{\mathrm{test}}|}
\sum_{i\in D^{\mathrm{test}}}\mathbf{1}\{\hat{y}_i=y_i\}.
\]
Because the prompted model is constrained to a yes/no answer, AUPRC has limited score resolution and is interpreted together with these thresholded metrics. Additional audit fields include AUROC, balanced accuracy, majority-class accuracy, accuracy minus majority accuracy, sensitivity, specificity, precision, test positive rate, predicted positive rate, and parse-failure count~\citep{fawcett2006roc}.  For scalar metric $z$,
\[
\bar{z}_{c,k}
=
\frac{1}{10}\sum_{e=1}^{10}z_{c,k,e},
\]
and \path{fewshot_metrics_agg.csv} reports the empirical standard deviation across the ten exemplar seeds.  The compact main-text Table~\ref{tab:scenario_performance} displays the means only.  The final Table~\ref{tab:scenario_performance} renderer should consume \path{table3_mapped_metrics_agg.csv}, not the runner's intermediate aggregate, so that condition completeness, seed identity, and leakage checks are always applied.

\paragraph{Failure handling and interpretation.}
A Table~\ref{tab:scenario_performance} run is nonreportable if the label has a single class, source split binding fails, the test split is empty, candidate rules cannot be parsed, all candidates are removed by leakage filters, no discovered card survives validation scoring for paper C2, the required real LLM is unavailable, or an answer cannot be parsed under the fail-closed policy.  If \texttt{max\_test\_samples} is used to control LLM cost, the subsampling mode, sampled class prevalence, and sampled row count are written to the metric files and run config.  Missing cards are not replaced with generic biomedical text.  Table~\ref{tab:scenario_performance} therefore measures the downstream prompt utility of SCENE-derived contextualized knowledge, not the independent biological or clinical truth of every card and not a causal validation of the upstream rules.

\paragraph{Reproduction command.}
The paper run invokes the three manuscript conditions directly:
\begin{appendixbox}{Reproduction pipeline}
$ProjectRoot = "<absolute path to the project root containing analysis/>"
$PythonExe = "<absolute path to the Python executable>"
$OutRoot = "<absolute path to the Table 3 output root>"
$env:PYTHONPATH = $ProjectRoot
$env:SILICONFLOW_API_KEY = "<provider API key>"
$env:SILICONFLOW_BASE_URL = "<OpenAI-compatible base URL>"

& $PythonExe -m analysis.run_knowledge_augmented_fewshot `
  --scene clinical `
  --task-id <clinical_task_id> `
  --source-run-dir "<absolute path to the Table 1 source run directory>" `
  --source-scenario both `
  --source-run-id run_01 `
  --conditions C0,C1,C2 `
  --shots 4 `
  --seeds 0,1,2,3,4,5,6,7,8,9 `
  --top-k-cards 5 `
  --min-card-val-effect-delta 0.0 `
  --row-conditioned-cards `
  --fewshot-sampling-mode label_group `
  --temperature 0.0 `
  --llm-model Qwen/Qwen3.5-9B `
  --no-llm-enable-thinking `
  --save-prompts `
  --strict-source-binding `
  --require-group-split `
  --require-real-llm `
  --llm-parse-fallback error `
  --out-dir (Join-Path $OutRoot "clinical")

& $PythonExe -m analysis.run_knowledge_augmented_fewshot `
  --scene l1000 `
  --task-id <l1000_task_id> `
  --source-run-dir "<absolute path to the Table 2 full-profile source run directory>" `
  --use-source-test-ids `
  --l1000-label-mode target_margin `
  --conditions C0,C1,C2 `
  --shots 4 `
  --seeds 0,1,2,3,4,5,6,7,8,9 `
  --top-k-cards 5 `
  --min-card-val-effect-delta 0.0 `
  --row-conditioned-cards `
  --temperature 0.0 `
  --llm-model Qwen/Qwen3.5-9B `
  --no-llm-enable-thinking `
  --save-prompts `
  --strict-source-binding `
  --require-group-split `
  --require-real-llm `
  --llm-parse-fallback error `
  --out-dir (Join-Path $OutRoot "l1000")

\end{appendixbox}

\subsection{LLM Backend Runtime Protocol}
\label{app:backend-runtime}
The backend-runtime comparison changes only the third-party LLM API backend while keeping the discovery scenario, split, seed, schema, population size, generation budget, support floor, and reporting predicate fixed. Total runtime is measured as wall-clock time for the full frozen discovery run, including API calls, parsing, controller validation, evolutionary scoring, and logging. Generation latency is the mean wall-clock time per evolutionary generation within that run. Because all backends are accessed through third-party APIs, these measurements are indicative runtime costs rather than hardware-normalized throughput; provider load and network conditions can change absolute values. The sweep is reported as a runtime sensitivity analysis and is not used to select rules or revise the main benchmark protocols. Figure~\ref{fig:model-runtime-benefit-cost} fixes the P-M-TR task, split, seed, schema, controller settings, and discovery budget, and changes only the third-party API backend. Held-out benefit is computed after discovery as bad-outcome ARR in percentage points, averaged over the top-10 discovery-ranked rules exported by each run; held-out outcomes are not used to rank or select these rules. Runtime includes API calls, parsing, controller validation, evolutionary scoring, and logging, and should be interpreted as indicative API wall-clock cost rather than hardware-normalized throughput.

\section{Additional Experiments and Analyses}
\label{app:additional-experiments}

This section provides implementation details and post-hoc evidence audits for SCENE. The post-hoc audits reuse completed clinical and L1000 outputs without additional LLM prompting or rule discovery, and no analysis involves holdout-driven model selection. The goal is to make the empirical evidence more legible along axes that are central to SCENE: effect--support geometry, contextualization of reported rules, L1000 inspectability, held-out replay reliability, semantic provenance, component behavior, generated-feature gating, grounding traces, planner consistency, report quality, the contribution of knowledge guidance, and sensitivity to explicit prior records. These audits should therefore be read as descriptive robustness and interpretation analyses rather than as new model-selection procedures.

For the clinical analyses, we use a benefit-oriented sign convention. The source summaries report $\mathrm{RD(Bad)}$ as treatment bad-outcome rate minus control bad-outcome rate. We therefore define
\begin{equation}
    \begin{aligned}
    \mathrm{Benefit}(R)
    &={}\Pr(Y_{\mathrm{bad}}=1 \mid R, T=0)\\
    &\quad-\Pr(Y_{\mathrm{bad}}=1 \mid R, T=1)\\
    &={}-\mathrm{RD(Bad)}.
    \end{aligned}
\end{equation}
Larger values indicate a larger reduction in bad outcomes under treatment. Unless otherwise stated, the clinical support axis denotes subgroup support, and the rule-level reliability analysis uses the smaller treatment-arm count inside the rule as the held-out support measure. This convention keeps the appendix aligned with the multi-objective evidence view used by SCENE: rules should have both a favorable scenario effect and enough support to be interpretable.

\subsection{Clinical Effect--Support--Uncertainty Geometry}
\label{app:s1-clinical-geometry}

The main clinical comparison table reports selected subgroup summaries, but a table alone does not show how treatment benefit is positioned relative to the support-validity criterion used by the primary benchmark. Raw support is difficult to interpret as a monotone objective: a rule covering nearly everyone can be uninformative, while a very small rule can be unstable. We therefore visualize the clinical comparison with the same usable-support diagnostic reported in Table~\ref{tab:clinical_best1_benefit_revised}, and then complements it with uncertainty and variability diagnostics.

\begin{figure}[!htb]
\centering
\includegraphics[width=\linewidth]{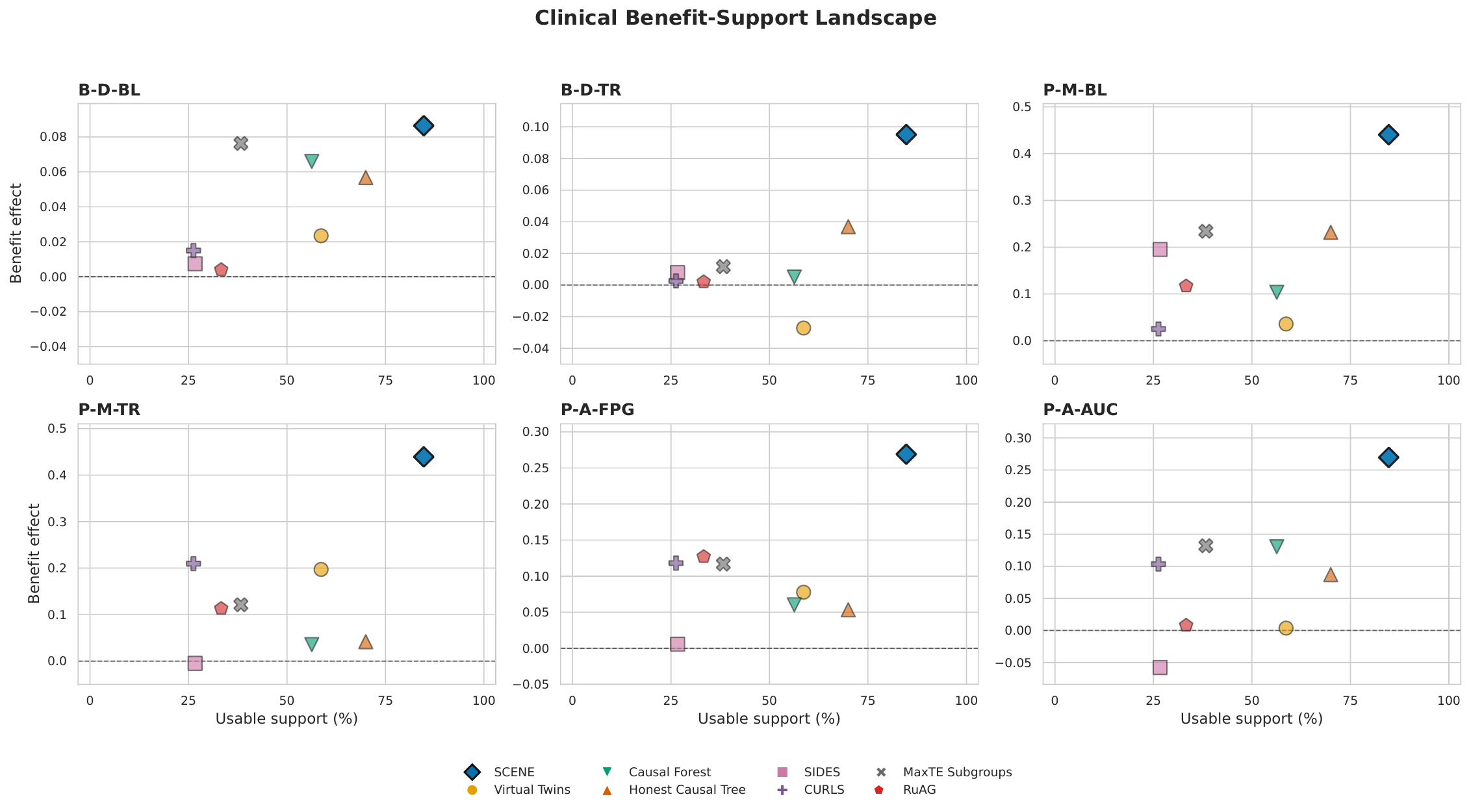}
\caption{Clinical benefit--support landscape. Each panel shows one clinical frame. The $x$-axis is usable-support value from Table~\ref{tab:clinical_best1_benefit_revised}, the $y$-axis is the corresponding benefit effect for that frame. Marker shape and color denote the method. SCENE is consistently separated on benefit while also retaining the highest usable support.}
\label{fig:s1a-benefit-support}
\end{figure}

For Figure~\ref{fig:s1a-benefit-support}, we use the method-level \emph{usable support} column reported in Table~\ref{tab:clinical_best1_benefit_revised}. Usable support measures how often a method's committed Best-1 rule has enough held-out coverage to be evaluated as a subgroup without degenerating into either a tiny fragment or a near-population rule. It is computed by first forming the task-level usability rate
\[
\begin{aligned}
u_{t,m}
={}&100\cdot \frac{1}{|\mathcal{R}_{t,m}|}
\sum_{r\in\mathcal{R}_{t,m}}\mathbb{I}\bigl\{
10\%\leq s_r\leq 60\%,\\
&\min(n_{r,\mathrm{treat}},n_{r,\mathrm{control}})\geq 5
\bigr\},
\end{aligned}
\]
where $\mathcal{R}_{t,m}$ is the set of Best-1 records available for method $m$ in clinical frame $t$, $s_r$ is the held-out subgroup coverage percentage for record $r$, and $n_{r,\mathrm{treat}}$ and $n_{r,\mathrm{control}}$ are the held-out treated and control counts inside the subgroup. The plotted support coordinate is then
\[
U_m=\frac{1}{6}\sum_{t=1}^{6} u_{t,m},
\]
which is the U-Supp. entry in Table~\ref{tab:clinical_best1_benefit_revised}. This binary usability flag penalizes both very small subgroups, which can be unstable, and near-population rules, which are weak subgroup discoveries, while also requiring minimal arm-specific support. The statistic is outcome-free: it does not use treatment effects or confidence intervals, and it asks only whether the committed Best-1 rule has a support profile suitable for held-out subgroup evaluation.

Across the 48 primary method--frame summaries used in this audit, the plotted benefit values reproduce the Table~1 ARR entries. SCENE has a median benefit effect of $0.269$, compared with $0.055$ for the non-SCENE baselines. On the Table~1 U-Supp. axis, SCENE reaches $84.7\%$, whereas the strongest non-SCENE baseline reaches $70.0\%$. At the per-frame level, SCENE has the largest benefit-effect point estimate in all six clinical frames. The correct interpretation is therefore not that SCENE always maximizes raw subgroup size, but that its high-benefit rules also satisfy the support-validity criterion used by the primary comparison.

Figure~\ref{fig:s1a-benefit-support} places SCENE and the clinical baselines into this benefit--usable-support landscape. Because the x-axis is the method-level Table~1 U-Supp. diagnostic, each method keeps the same support coordinate across panels, while the y-coordinate changes with the clinical frame's ARR. SCENE lies in the upper-right region across the six panels: it combines the largest frame-specific benefit values with the highest Table~1 usable support. This pattern is consistent with the role of SCENE's lower-level multi-objective search: it should not simply maximize subgroup size, nor should it collapse to the smallest high-effect subgroup.

The clinical comparison in Table~\ref{tab:clinical_best1_benefit_revised} reports Best-1 subgroup summaries for each method and task. The table is designed to make the primary comparison compact, but it does not directly show whether a high benefit estimate is accompanied by unstable behavior across repeated Best-1 records. We therefore add a complementary diagnostic that removes confidence intervals from the visualization and instead summarizes empirical variability of the Best-1 benefit effect. This audit is descriptive rather than inferential: it is not used for model selection, and it is not intended to replace the patient-level confidence intervals in the main quantitative comparison. Its purpose is to check whether the favorable clinical effect estimates of SCENE are accompanied by excessive run-to-run variability.

For this analysis, the benefit effect is the bad-outcome reduction,
\[
    \Delta = \mathrm{ControlBadRate} - \mathrm{TreatmentBadRate}.
\]
Larger values are better. The input records are aligned with the Table~\ref{tab:clinical_best1_benefit_revised} reporting protocol: when a method exports a ranked rule list, only the prioritized rule is used as its Best-1 rule; archived replay-audit records contribute one Best-1 replay record per row. For each method and clinical frame, we first compute a task-level center and a task-level variability statistic over these Best-1 records. We then average those task-level summaries across the six Table~\ref{tab:clinical_best1_benefit_revised} clinical frames. The resulting figures therefore show one point per method, not per-run points, and do not encode or reveal the number of available records for any method.

We use three variability definitions. The first uses the mean benefit as the center and the standard deviation (SD) as the variability statistic. The second uses the median benefit as the center and the median absolute deviation (MAD) around that median as a robust variability statistic. The third again uses the median benefit as the center but uses the interquartile range (IQR) as the variability statistic. In each plot, the horizontal axis is the benefit center, so moving right is better. The vertical axis is the empirical variability estimate, so moving upward is better because the axis is drawn with lower variability at the top. The upper-right region is therefore the favorable region: high benefit and low variability.

\begin{figure}[!htb]
\centering
\includegraphics[width=0.49\textwidth]{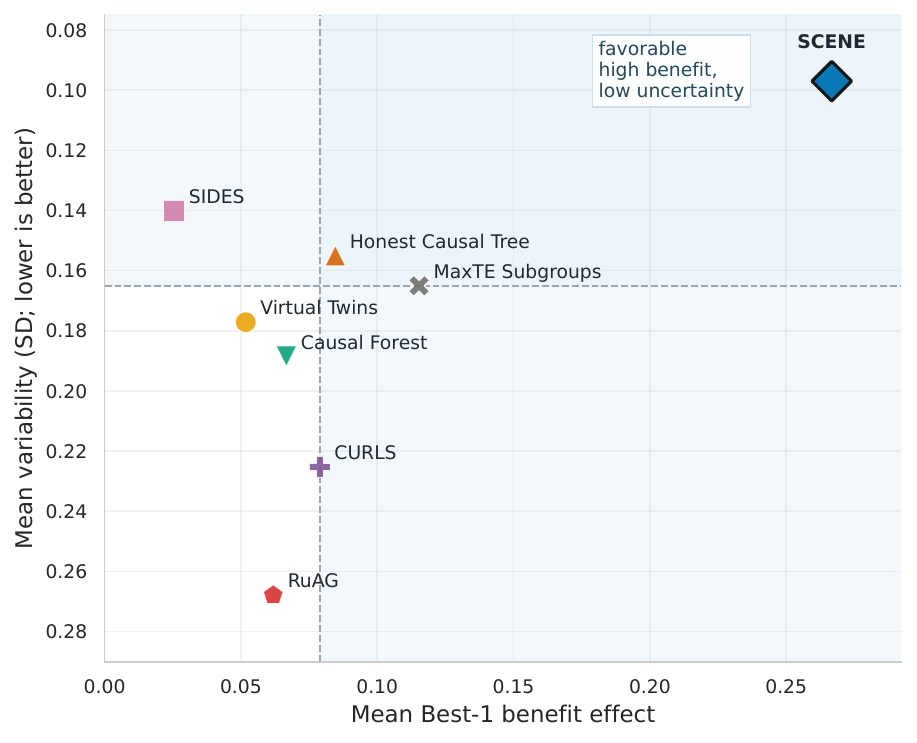}\hfill
\includegraphics[width=0.49\textwidth]{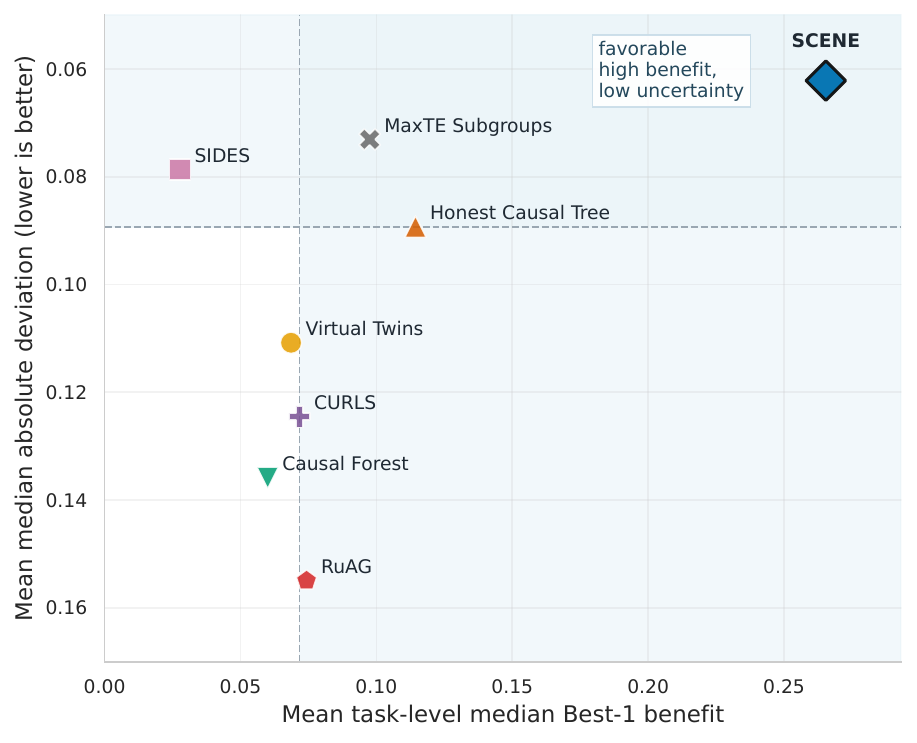}
\caption{Clinical benefit--variability profiles.
(Left) Mean benefit with task-level SD. (Right) Median benefit with task-level MAD. In both panels, moving right/up indicates higher benefit and lower empirical variability.}
\label{fig:s1b-mean-sd-profile}
\label{fig:s1b-median-mad-profile}
\end{figure}

Figure~\ref{fig:s1b-mean-sd-profile} gives the most conventional variability view. SCENE has the largest mean benefit effect ($0.267$). The same figure also shows that SCENE does not obtain this larger benefit by becoming more variable. Its average task-level SD is $0.097$, lower than every baseline in this audit. The closest baselines on variability are SIDES ($0.140$), Honest Causal Tree ($0.155$), and MaxTE Subgroups ($0.165$), but all three are substantially left of SCENE on the benefit axis. Conversely, CURLS and Causal Forest are not only lower-benefit but also more variable under this summary. Thus the mean-SD view supports the main table's interpretation: SCENE is not merely selecting a high-effect subgroup in a noisy manner; it is separated from the baselines in the favorable direction on both axes.

The median-MAD view in Figure~\ref{fig:s1b-median-mad-profile} asks whether the conclusion is driven by a small number of extreme Best-1 records. The answer is no. SCENE again has the largest center statistic, with a mean task-level median benefit of $0.265$. Its average MAD is $0.062$, which is also the smallest among all methods. The strongest baseline after applying the same robust summary is Honest Causal Tree by the stability-adjusted read-out, but its median benefit is only $0.114$ and its MAD is $0.089$. MaxTE Subgroups is close to Honest Causal Tree under this robust view, with median benefit $0.098$ and MAD $0.073$, but it remains far below SCENE in benefit. This is important because MAD is less sensitive to outlying records than SD. The fact that SCENE remains in the favorable upper-right region under MAD suggests that the result is not simply a mean-based artifact.

\begin{figure}[!htb]
\centering
\begin{minipage}[t]{0.48\textwidth}
    \vspace{0pt}
    \includegraphics[width=\linewidth]{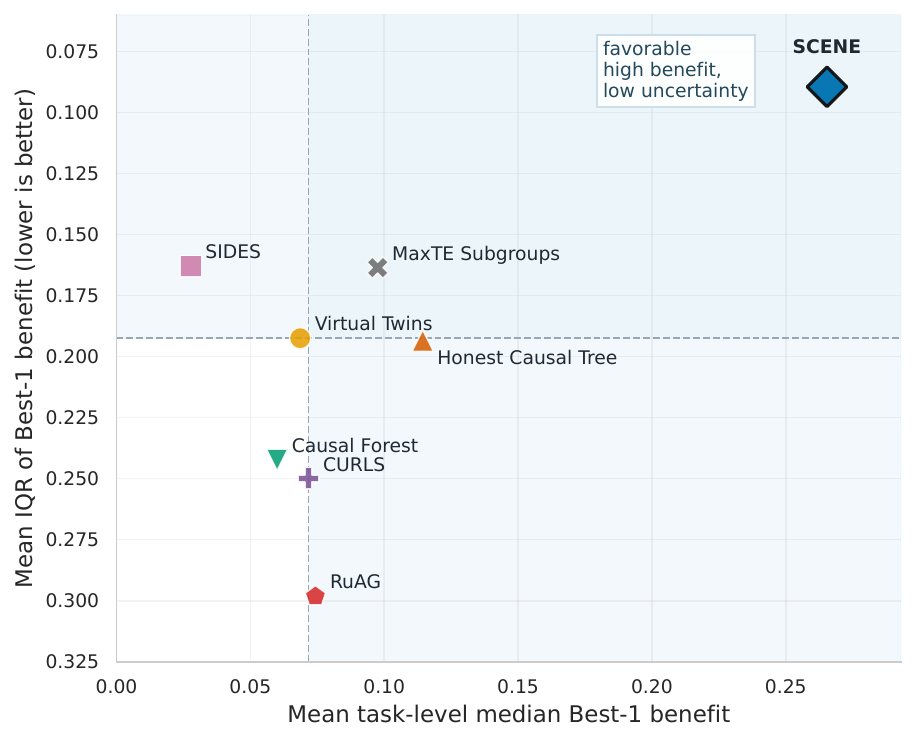}
\end{minipage}\hfill
\begin{minipage}[t]{0.50\textwidth}
    \vspace{0pt}
    \centering
    \small
    \setlength{\tabcolsep}{3pt}
    \renewcommand{\arraystretch}{1.45}
    \begin{tabular}{@{}c@{}}
    \textbf{Task-averaged geometry audit}\\[0.35em]
    \begin{tabular}{@{}lccc@{}}
    \toprule
    Read-out & SCENE & Best base. & Margin \\
    \midrule
    Benefit effect & 26.7 pp & 11.6 pp & +15.1 pp \\
    Effect/CI signal & 0.234 & 0.127 & +0.107 \\
    Benefit score & 1.000 & 0.430 & +0.570 \\
    Signal score & 0.904 & 0.541 & +0.363 \\
    Composite score & 0.976 & 0.520 & +0.309 \\
    \bottomrule
    \end{tabular}
    \end{tabular}
\end{minipage}
\caption{Robust clinical geometry and aggregate read-outs.
(Left) Median benefit with task-level IQR. (Right) Method-level averages over the six clinical frames; the margin compares SCENE with the strongest baseline for each read-out.}
\label{fig:s1b-median-iqr-profile}
\end{figure}

The compact audit table in Figure~\ref{fig:s1b-median-iqr-profile} uses the same task-level scorecard as the benefit--support landscape. The raw \emph{Benefit effect} is the bad-outcome reduction in percentage points. The \emph{Effect/CI signal} is the benefit effect divided by the width of the corresponding confidence interval, so it rewards effects that are large relative to their uncertainty band. The \emph{Benefit score} and \emph{Signal score} are task-wise min--max normalized versions of the benefit effect and Effect/CI signal, respectively. The \emph{Composite score} is a benefit-weighted scorecard read-out, computed as $0.50$ times the Benefit score plus $0.25$ times the normalized support score and $0.25$ times the Signal score. These score rows are therefore relative geometry diagnostics within each clinical frame, not new clinical endpoints.

Figure~\ref{fig:s1b-median-iqr-profile} uses IQR to focus on the middle half of the Best-1 records. This gives a second robust dispersion check that is different from MAD. The pattern remains stable. SCENE has a mean task-level median benefit of $0.265$ and an average IQR of $0.089$. The best baseline median benefit remains much smaller, and the baseline IQR values are all larger: SIDES and MaxTE Subgroups are around $0.163$--$0.164$, Honest Causal Tree and Virtual Twins are around $0.193$, and Causal Forest and CURLS are around $0.242$--$0.250$. The table next to the IQR panel adds a complementary task-averaged audit from the effect--support--uncertainty scorecard. SCENE has the highest average benefit effect, effect-to-CI signal, normalized benefit score, normalized signal score, and benefit-weighted composite score across the six clinical frames, with consistent margins over the strongest baseline for each read-out. Thus the IQR view and the compact table make distinct points: the figure audits empirical dispersion, while the table summarizes method-level separation on the clinical geometry scorecard without reusing the variability axes.

\begin{table}[!htb]
\caption{Summary of non-CI benefit--variability diagnostics. The values are method-level averages over the six Table 1 clinical frames. ``Adjusted'' is center minus variability and is reported only as a numerical read-out; it is not used as an additional figure axis.}
\label{tab:s1b-variability-summary}
\centering
\begingroup
\small
\setlength{\tabcolsep}{3pt}
\begin{tabular*}{\linewidth}{@{\extracolsep{\fill}}lcccc@{}}
\toprule
Diagnostic & SCENE Center & SCENE variability & SCENE adjusted & Best baseline adjusted \\
\midrule
Mean--SD & 0.267 & 0.097 & 0.170 & -0.053 \\
Median--MAD & 0.265 & 0.062 & 0.203 & 0.025 \\
Median--IQR & 0.265 & 0.089 & 0.176 & -0.066 \\
\bottomrule
\end{tabular*}
\endgroup
\end{table}

Table~\ref{tab:s1b-variability-summary} summarizes the same observation numerically. If one subtracts the variability estimate from the corresponding benefit center, SCENE remains the top method under all three non-CI diagnostics. The margin is largest in the mean-SD and median-IQR views, where all baseline adjusted values are negative. In the median-MAD view, the best baselines obtain small positive adjusted values, but they remain far below SCENE. This makes the qualitative conclusion robust to the choice of variability statistic.

These diagnostics should be interpreted carefully. SD, MAD, and IQR measure empirical variability across archived Best-1 records, not uncertainty in the formal statistical sense and not a replacement for held-out clinical validation. They are useful here because they avoid relying on potentially inconsistent CI endpoints from heterogeneous baseline output files and because they do not expose the number of replay records for each method. The consistent placement of SCENE across mean-SD, median-MAD, and median-IQR views provides a simple robustness check for the main clinical table: the method's advantage is not limited to a single confidence-interval rendering or a single dispersion statistic. Instead, SCENE remains the method with the highest clinical benefit and the lowest empirical variability across all three descriptive views.

\subsection{Clinical Contextualization Advantage over Prior Rule Vocabularies}
\label{app:s8-clinical-contextualization-advantage}

\begin{figure}[!htb]
\centering
\includegraphics[width=\linewidth]{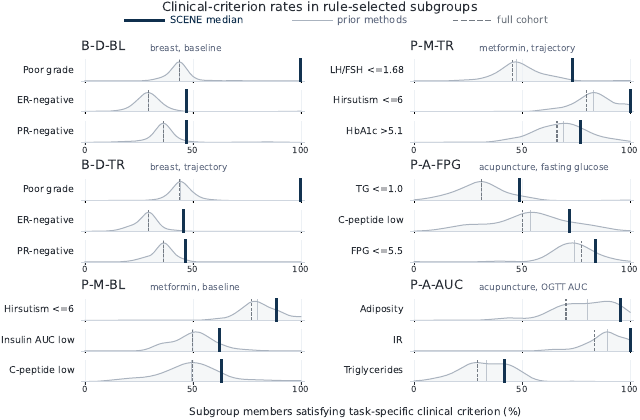}
\caption{Task-specific clinical-context audit of reported top rules.
Each task block contains three endpoint-relevant non-outcome clinical
criteria.  Dark vertical markers show the median across actual SCENE Best-1 rules
outputs of 50 runs; gray densities summarize replayable Best-1 rules of 50 runs from prior
methods; the dashed line is the full-cohort criterion rate.  The horizontal
axis is the percentage of subgroup members satisfying the displayed criterion.}
\label{fig:s8-scene-toprules-against-methods}
\end{figure}

Table~\ref{tab:clinical_best1_benefit_revised} evaluates whether each method's exported rule improves the held-out
clinical endpoint.  The present audit asks a different question: when a method
commits to its top-ranked rules, do the induced subgroups also concentrate on
clinically meaningful, task-relevant axes rather than arbitrary threshold
combinations? The prior-method reference distributions in this audit pool the six original data-only baselines. For each clinical task frame, we define three non-outcome
clinical criteria before plotting. The breast-cancer frames use standard
prognostic axes from the trial setting: poor histologic grade, estrogen
receptor negativity, and progesterone receptor negativity. The PCOS frames use
endpoint-matched endocrine, glycemic, lipid, and insulin-resistance axes.  For
metformin-response tasks these include hirsutism/gonadotropin phenotype and
insulin-exposure markers; for the fasting-glucose acupuncture task these
include fasting-glucose, triglyceride, and C-peptide criteria.  For the OGTT
AUC acupuncture task, we intentionally reuse the same three metabolic-risk
criteria shown in the main-text clinical-risk audit: central adiposity
(\(\mathrm{waist}\ge80\) cm), insulin resistance
(\(\mathrm{HOMA\mbox{-}IR}\ge2.5\)), and elevated triglycerides
(\(\mathrm{TG}\ge1.7\) mmol/L).  These criteria are used only for post-hoc
contextualization auditing and are never used to select or revise the rules.

Figure~\ref{fig:s8-scene-toprules-against-methods} compares actual SCENE Best-1 rule outputs with replayable Best-1 rules from prior methods for each task.  For
each task-specific clinical criterion, the gray density summarizes the
distribution over prior-method top-rule occurrences, while the dark vertical
marker reports the median over SCENE top-rule occurrences.  We visualize SCENE
by its median rather than by a kernel density because the number of SCENE
reported top-rule records is smaller than the overall prior-method rule set. The
dashed vertical line is the corresponding full-cohort rate. A rightward SCENE
median therefore means that SCENE's reported top rules more consistently
identify subgroups concentrated on that clinical axis.

Table~\ref{tab:s8-axis-context-matrix} expands the task-level summary into a full axis-level audit over the predefined non-outcome axes used in
Figure~\ref{fig:s8-scene-toprules-against-methods}.  This larger table is included for transparency rather
than for additional model selection: it shows the cohort rate, the prior-method
rule-vocabulary median, and the SCENE median for every predefined clinical
axis used in Figure~\ref{fig:s8-scene-toprules-against-methods}.  Reporting the
complete matrix makes the contextualization claim less dependent on a single
aggregate read-out.

\begin{table}[!htb]
\caption{Axis-level clinical contextualization matrix.
Values are percentages of subgroup members satisfying the displayed
non-outcome clinical criterion.  ``Prior'' is the median over replayable
top-ranked prior-method rules, and ``SCENE'' is the median over SCENE top-rule
outputs. The axes are fixed contextualization probes and are not used to
select or revise rules. This is a post-hoc contextualization audit, not causal
feature attribution.}
\label{tab:s8-axis-context-matrix}
\centering
\small
\begingroup
\setlength{\tabcolsep}{2.2pt}
\begin{tabular*}{\linewidth}{@{\extracolsep{\fill}}llccccc@{}}
\toprule
Frame & Clinical axis & Cohort & Prior & SCENE & SCENE--prior & SCENE--cohort \\
\midrule
B-D-BL & ER-negative & 29.2 & 29.3 & 46.9 & +17.6 & +17.6 \\
B-D-BL & PR-negative & 36.3 & 36.2 & 47.1 & +10.8 & +10.7 \\
B-D-BL & Poor grade & 43.7 & 43.7 & 100.0 & +56.3 & +56.3 \\
\midrule
B-D-TR & ER-negative & 29.2 & 29.5 & 45.6 & +16.0 & +16.3 \\
B-D-TR & PR-negative & 36.3 & 36.4 & 46.7 & +10.2 & +10.4 \\
B-D-TR & Poor grade & 43.7 & 44.3 & 100.0 & +55.7 & +56.3 \\
\midrule
P-M-BL & C-peptide low & 50.0 & 49.2 & 63.2 & +13.9 & +13.2 \\
P-M-BL & Hirsutism \(\le6\) & 77.3 & 80.0 & 88.7 & +8.7 & +11.4 \\
P-M-BL & Insulin AUC low & 50.0 & 50.0 & 62.3 & +12.3 & +12.3 \\
\midrule
P-M-TR & HbA1c \(>5.1\) & 66.2 & 69.0 & 77.1 & +8.1 & +11.0 \\
P-M-TR & Hirsutism \(\le6\) & 80.0 & 83.1 & 100.0 & +16.9 & +20.0 \\
P-M-TR & LH/FSH \(\le1.68\) & 45.6 & 47.2 & 73.2 & +25.9 & +27.5 \\
\midrule
P-A-FPG & C-peptide low & 50.3 & 54.1 & 71.7 & +17.7 & +21.5 \\
P-A-FPG & FPG \(\le5.5\) & 77.4 & 74.3 & 84.1 & +9.8 & +6.7 \\
P-A-FPG & TG \(\le1.0\) & 31.3 & 31.2 & 48.7 & +17.5 & +17.4 \\
\midrule
P-A-AUC & Adiposity & 70.3 & 80.2 & 95.6 & +15.5 & +25.3 \\
P-A-AUC & HOMA-IR \(\ge2.5\) & 83.3 & 89.4 & 100.0 & +10.6 & +16.7 \\
P-A-AUC & TG \(\ge1.7\) & 29.2 & 33.3 & 42.0 & +8.6 & +12.8 \\
\bottomrule
\end{tabular*}
\endgroup
\end{table}

Across all six clinical frames, the SCENE median is shifted toward clinically
interpretable task-specific phenotypes relative to the prior-method rule
vocabulary.  The expanded matrix makes the consistency of this shift visible:
all 18 predefined axes have positive SCENE--prior margins.  The smallest
margin is still positive (HbA1c in the metformin trajectory frame, $+8.1$
percentage points), while the largest margins occur on the breast-cancer
histologic-grade axis, where SCENE subgroups concentrate poor-grade tumors
much more strongly than the replayed prior-method rule vocabulary.  The PCOS
frames show the same pattern on different clinical axes: metformin tasks shift
toward endocrine and insulin-exposure phenotypes, whereas acupuncture glucose
and OGTT tasks shift toward glycemic, insulin-resistance, adiposity, and
triglyceride axes.  Thus the table adds information beyond the task-level
summary: it shows that the contextualization advantage is not carried by one
favorable axis or one disease frame.

This does not establish causal feature attribution or clinical treatment
guidance.  It supports the more limited claim that SCENE's reported rule
vocabulary is more contextualized: the rules that obtain held-out benefit also
map to recognizable clinical axes in the corresponding task frame, and this
mapping remains visible when the audit is expanded from task averages to the
individual clinical probes.

\subsection{L1000 Proposition Inspectability}
\label{app:l1000-proposition-inspectability}

\begin{figure}[!htb]
    \centering
    \includegraphics[width=\linewidth]{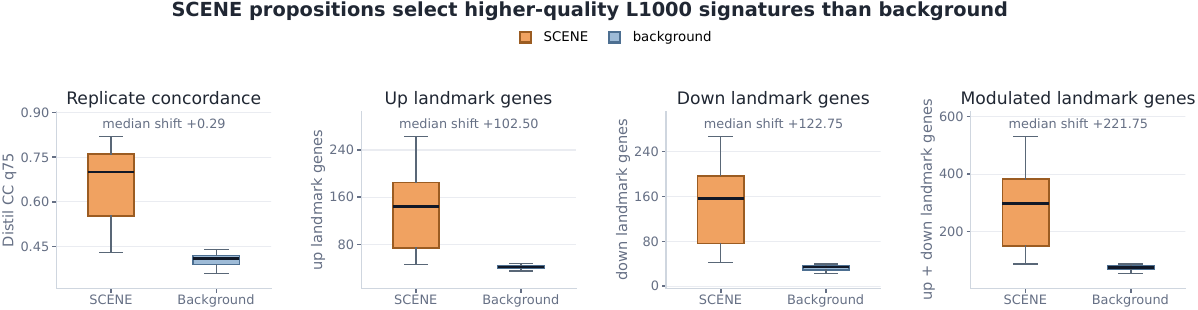}
    \caption{Matched L1000 replay audit.
    SCENE-selected signatures are compared with each proposition's held-out background using complementary L1000 audit metrics. Panels report replicate concordance, up-regulated landmark genes, down-regulated landmark genes, and total modulated landmark genes.}
    \label{fig:l1000-proposition-gain}
\end{figure}

\begin{figure}[!htb]
    \centering
    \includegraphics[width=\linewidth]{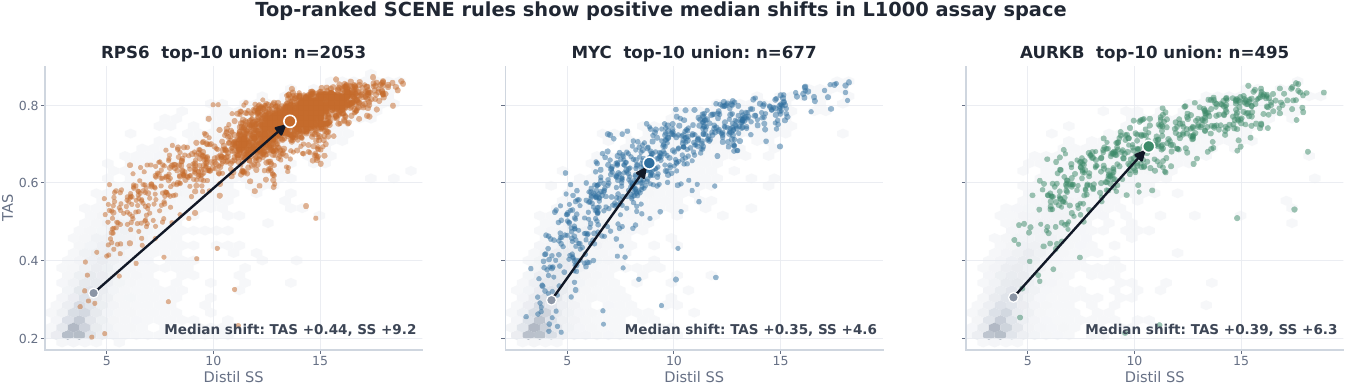}
    \caption{Union of Best-1 full-profile SCENE rules across ten target-seed runs.
    Colored points are signatures covered by at least one selected SCENE rule; gray hexagons show held-out density and arrows show median displacement.}
    \label{fig:l1000-top-rank-replay}
\end{figure}

SCENE is designed to return executable propositions rather than only a scalar score or an uninterpreted high-performing subset. In the L1000 setting, this means that a discovered rule should be replayable on held-out perturbational signatures and should identify a concrete region of the assay space that can be inspected by standard LINCS/L1000 quality and response measurements. This appendix evaluates that property from two complementary views: a proposition-level selected-versus-background audit and a target-level replay of top-ranked rule sets. Figure~\ref{fig:l1000-proposition-gain} reports the selected-versus-background audit, while Figure~\ref{fig:l1000-top-rank-replay} visualizes the replayed rule unions in L1000 assay space.

The selected-versus-background audit in Figure~\ref{fig:l1000-proposition-gain} tests whether exported propositions remain meaningful after held-out replay. The unit of the boxplot is a replayed proposition: for each rule, we compute the median metric value among selected signatures and the median value among its corresponding background signatures. The panels use audit metrics that do not duplicate the TAS--Distil SS replay axes: replicate concordance, up-regulated landmark genes, down-regulated landmark genes, and total modulated landmark genes. Across the propositions pooled over RPS6, MYC, and AURKB, median selected-background differences are positive in all four views: $+0.29$ Distil CC q75, $+102.5$ up-regulated landmark genes, $+122.8$ down-regulated landmark genes, and $+221.8$ total modulated landmark genes. The background boxes are narrow because each background is a large, overlapping held-out complement; this stability makes the selected-background separation more interpretable rather than less informative.

Figure~\ref{fig:l1000-top-rank-replay} complements the proposition-level audit by showing where the selected signatures lie in the assay space. For each target, we take the Best-1 full-profile rule from each of the ten Table~\ref{tab:l1000_ablation} target-seed runs and visualize the union of held-out signatures covered by at least one selected rule. The top-rule unions cover 2053 RPS6 signatures, 677 MYC signatures, and 495 AURKB signatures. They shift the covered-signature median by $+0.44$ TAS and $+9.2$ Distil SS for RPS6, $+0.35$ TAS and $+4.6$ Distil SS for MYC, and $+0.39$ TAS and $+6.3$ Distil SS for AURKB. Visually, these median-displacement arrows move toward the high-activity, high-strength region of the L1000 assay space. These positive shifts support the practical inspectability claim: SCENE outputs rules that can be replayed, audited, and interpreted as concrete perturbational contexts.

\subsection{Holdout Reliability and Support--Gap Diagnostics}
\label{app:s2-holdout-reliability}

A central failure mode of subgroup discovery is rule overfitting: a rule may have a strong discovery-side contrast but fail when replayed on held-out patients. This section therefore audits rule-level replay records instead of only reporting selected task-level summaries. This audit is especially important for SCENE because the lower-level search explores many candidate rules; a credible discovery system must expose where the archive is stable and where it is support-limited.

\begin{figure}[!htb]
\centering
\includegraphics[width=0.92\linewidth]{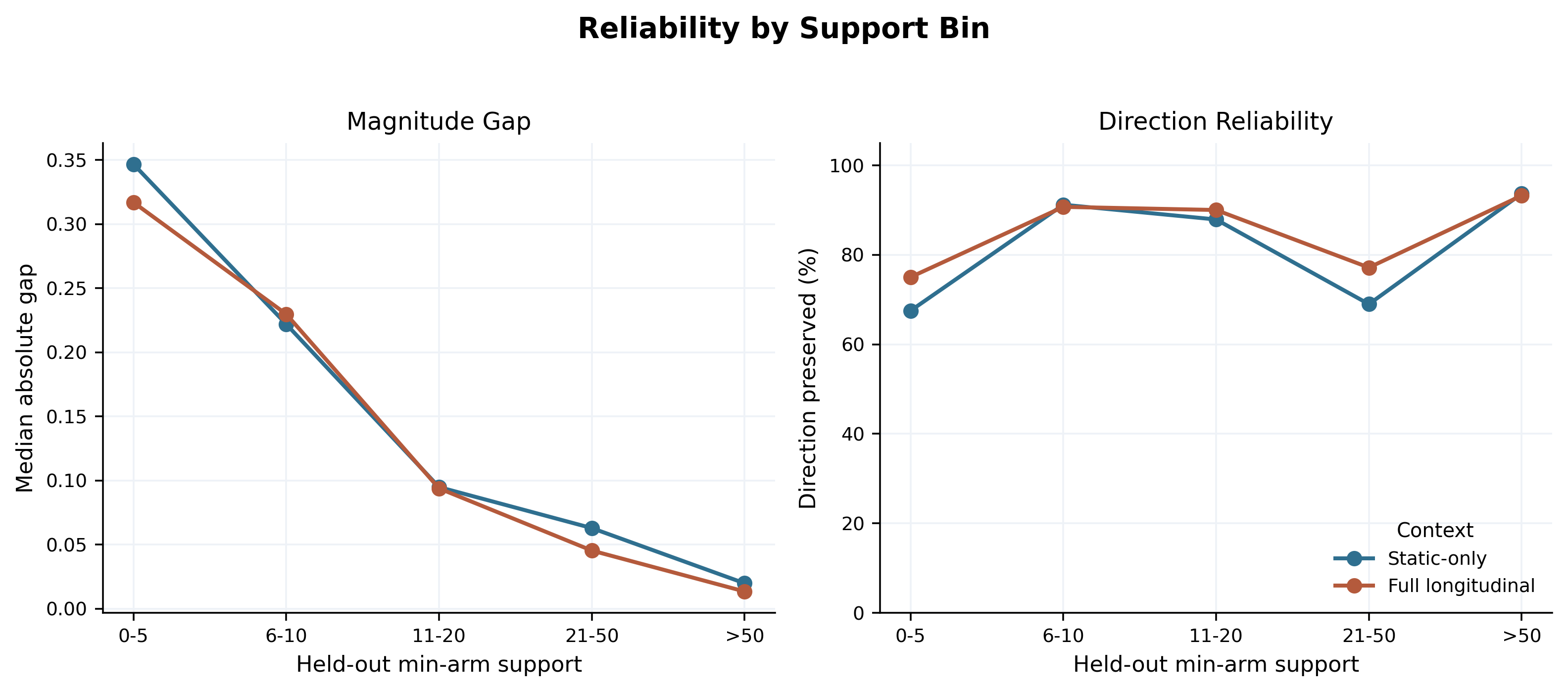}
\caption{Reliability by held-out support bin. The left panel shows the median absolute train--holdout gap by smaller-arm held-out support. The right panel shows the direction-preservation rate. Magnitude reliability improves sharply with support, while direction reliability is generally higher for supported rules but not perfectly monotonic across intermediate bins.}
\label{fig:s2d-support-bin}
\end{figure}

Figure~\ref{fig:s2d-support-bin} shows the held-out effect direction is preserved in $84.9\%$ of evaluations. Direction preservation is $86.4\%$ for full longitudinal contexts and $83.4\%$ for static-only contexts. The median absolute train--holdout gap is $0.071$, and the mean absolute gap is $0.175$. These values indicate that direction is more reliable than magnitude: many rules preserve whether treatment appears beneficial, but the exact effect size is often attenuated on held-out patients.

Table~\ref{tab:s2-support-floor-sensitivity} complements the binned figure from a different angle: it treats held-out smaller-arm support as a reporting boundary and asks how the replay audit changes under increasingly conservative minimum-support floors. The rows are cumulative thresholds. This table is therefore a sensitivity check on how much of the replay archive remains interpretable under stricter support requirements, whereas Figure~\ref{fig:s2d-support-bin} shows where the support--gap relationship is located. Unlike Table~\ref{tab:clinical_best1_benefit_revised} Best-1 DC, which is computed only over committed Best-1 rules, this audit includes the broader eligible replay archive, namely the Top-5 rules from each run.

\begin{table}[!htb]
\caption{Support-floor sensitivity of held-out replay reliability.
The table reports cumulative minimum smaller-arm support floors. Retained archive is shown as a percentage of eligible replay records. The gap is $|\Delta_{\mathrm{train}}-\Delta_{\mathrm{holdout}}|$ and is shown in percentage points.}
\label{tab:s2-support-floor-sensitivity}
\centering
\begingroup
\small
\setlength{\tabcolsep}{3pt}
\begin{tabular*}{\linewidth}{@{\extracolsep{\fill}}lcccc@{}}
\toprule
Support floor & Retained (\%) & Direction preserved (\%) & Held-out positive (\%) & Gap median/P75 (pp) \\
\midrule
All replayed & 100.0 & 84.9 & 91.3 & 7.1 / 17.5 \\
$m_{\min}\ge 10$ & 67.4 & 91.8 & 93.8 & 5.0 / 8.4 \\
$m_{\min}\ge 20$ & 47.9 & 92.0 & 94.2 & 3.9 / 6.9 \\
$m_{\min}\ge 50$ & 21.5 & 98.4 & 100.0 & 2.0 / 3.2 \\
\bottomrule
\end{tabular*}
\endgroup
\end{table}

The sensitivity pattern is useful because it separates two claims that can otherwise be conflated. Directional replay is already high when all eligible rules are considered, but magnitude agreement is support-sensitive. As the support floor becomes stricter, a smaller part of the archive remains, while direction preservation and positive held-out replay increase and the train--holdout gap decreases. This supports a conservative reporting practice: SCENE's replay archive should be interpreted with support-aware uncertainty, rather than treating every discovered rule as equally precise.

\subsection{Semantic Rule Composition and Feature Provenance}
\label{app:s3-rule-semantics}

SCENE is designed to produce scenario-grounded rules rather than opaque prediction scores. This section asks what kinds of clinical evidence the discovered rules actually use. We parse clinical rule text into semantic feature families and canonicalize rule predicates to reduce duplicate syntax. This analysis is a syntax and provenance audit, not a causal feature-importance analysis. A rule may contribute to multiple feature families when it combines predicates from multiple evidence sources.

\begin{figure}[!htb]
\centering
    \includegraphics[width=\linewidth]{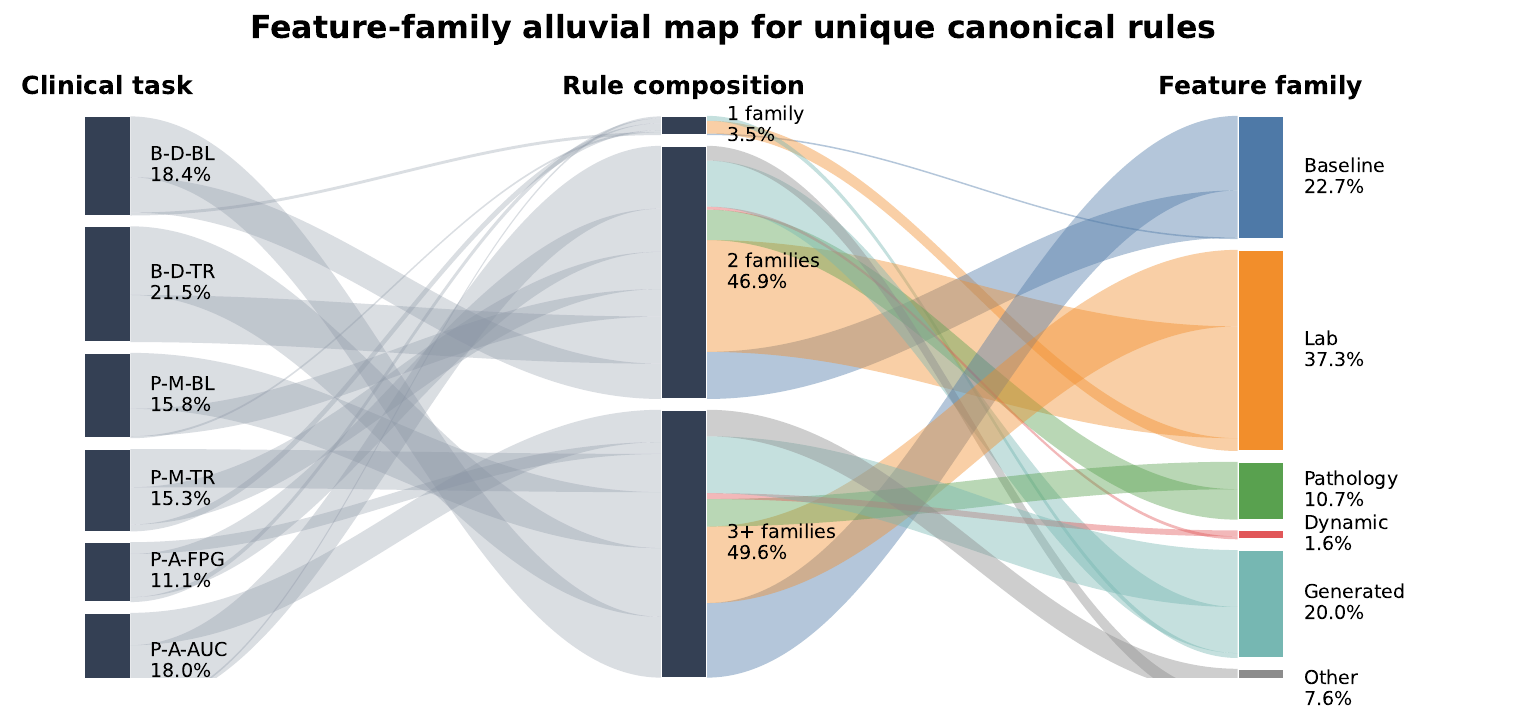}
\caption{Feature-family assignments from clinical rules. The alluvial map links the six clinical tasks, rule-composition breadth, and feature family. Percentages are computed over rule-family assignments derived from the SCENE selected rules used in Table~\ref{tab:clinical_best1_benefit_revised}; multi-family rules contribute once to each family they use, and rules appearing in multiple tasks contribute once to each observed task.}
\label{fig:s3a-alluvial}
\end{figure}

Figure~\ref{fig:s3a-alluvial} summarizes the SCENE selected rules used in Table~\ref{tab:clinical_best1_benefit_revised} as percentages over rule-family assignment flow units. The middle column now shows rule-composition breadth: $3.5\%$ of the displayed assignments come from single-family rules, $46.9\%$ from two-family rules, and $49.6\%$ from rules spanning three or more feature families. On the feature-family side, laboratory measurements are the largest component ($37.3\%$), followed by baseline clinical history ($22.7\%$), generated bins/formulas ($20.0\%$), tumor/pathology descriptors ($10.7\%$), other or unmapped features ($7.6\%$), and longitudinal dynamics ($1.6\%$). Dynamic predicates are therefore present but sparse; the appropriate interpretation is that longitudinal evidence acts as an additional context layer rather than dominating the rule vocabulary. Because multi-family rules contribute once to each used family, the displayed widths are assignment-weighted rather than unique-rule counts. Other includes unmapped or rare feature families.

Table~\ref{tab:s3-semantic-archetypes} adds a rule-internal evidence audit that is not visible in the alluvial map. The alluvial figure answers which feature families appear in the archive; the table asks whether recurring semantic archetypes with held-out replay evidence also retain positive clinical effect estimates. We report only aggregate read-outs for archetypes, not raw archive counts. The rows should be interpreted as syntax/provenance summaries of executable rules, not as causal attribution to individual feature families.

\begin{table}[!htb]
\caption{Held-out evidence among recurring semantic archetypes.
Rows summarize canonical rules in the displayed family combination for which held-out effects are available. Benefit is the bad-outcome reduction in percentage points; positive held-out is the share of those rules with benefit greater than zero.}
\label{tab:s3-semantic-archetypes}
\centering
\small
\begingroup
\setlength{\tabcolsep}{3.5pt}
\begin{tabularx}{\linewidth}{@{}>{\raggedright\arraybackslash}p{0.24\linewidth}>{\centering\arraybackslash}p{0.14\linewidth}>{\centering\arraybackslash}p{0.16\linewidth}>{\raggedright\arraybackslash}X@{}}
\toprule
Semantic archetype & Median benefit & Positive held-out & Interpretation \\
\midrule
Baseline + Lab & +19.0 pp & 94.7\% & Patient baseline context combined with direct clinical measurements. \\
Lab + Generated & +16.9 pp & 94.1\% & Measured variables augmented by derived scenario-specific summaries. \\
Baseline + Lab + Generated & +12.6 pp & 100.0\% & Three-source contextualization linking patient state, measurement, and generated summaries. \\
Lab only reference & +14.3 pp & 72.7\% & Single-family laboratory thresholds, included as a shortcut reference. \\
\bottomrule
\end{tabularx}
\endgroup
\end{table}

\subsection{L1000 Ablation and Holdout Evidence Diagnostics}
\label{app:s4-l1000-ablation}

The L1000 scenario evaluates whether the same SCENE framework can produce context-bounded target-response findings outside the clinical-trial setting. Here, the relevant outcome is not clinical efficacy but holdout connectivity evidence, target retrieval, support, significance, and generalization. The following diagnostics expand the main ablation table using the same paired L1000 outputs. They are descriptive because they summarize completed, frozen ablation artifacts and do not change the profile selection or replay protocol.

\begin{figure}[!htb]
\centering
\includegraphics[width=0.72\linewidth]{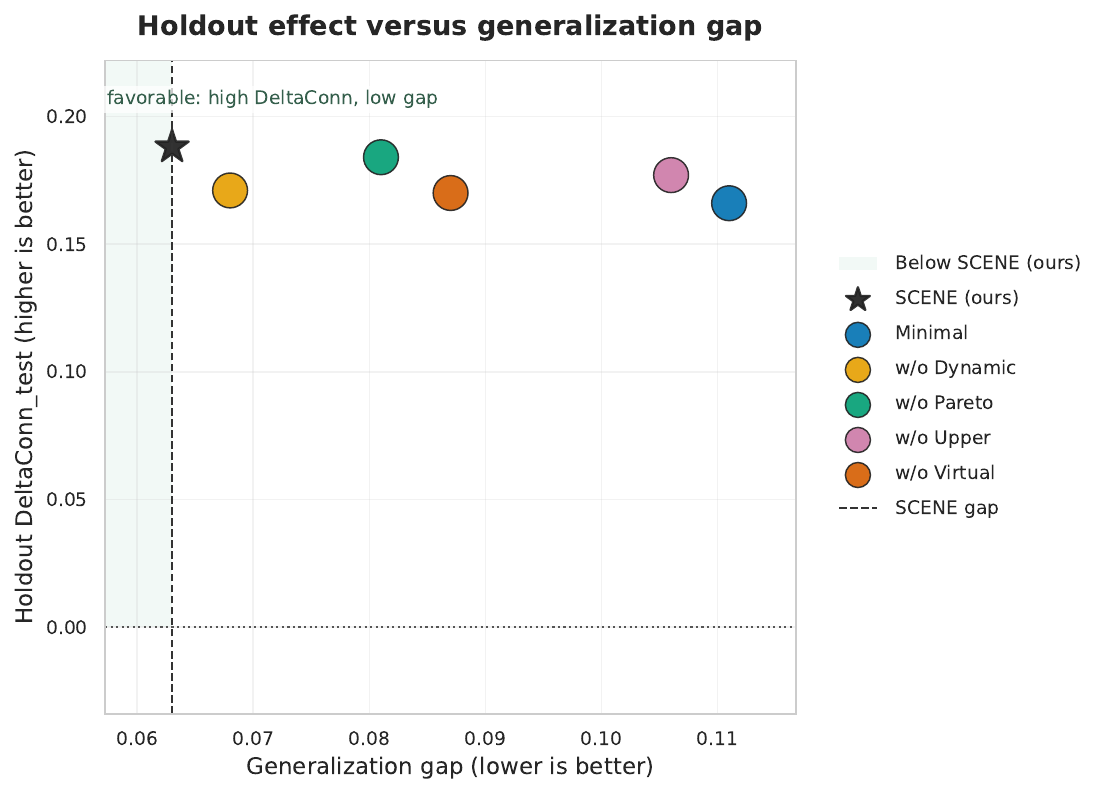}
\caption{L1000 holdout effect versus generalization gap. Each point is a run-aggregated ablation variant. Higher $\Delta$Conn indicates stronger holdout connectivity evidence, whereas lower gap indicates better train--holdout stability. The shaded upper-left region marks the favorable regime. SCENE (ours) combines high holdout effect with the smallest generalization gap among the displayed variants. The dashed vertical line marks the SCENE (ours) gap.}
\label{fig:s4c-effect-gap}
\end{figure}

Figure~\ref{fig:s4c-effect-gap} separates the two most important robustness axes: holdout connectivity effect and generalization gap. The favorable region is high DeltaConn and low gap. SCENE (ours) lies closest to this region in the current run summaries, whereas several ablations retain positive DeltaConn but incur larger generalization gaps.

Table~\ref{tab:s4-minimal-centered-profile} gives a minimal-baseline-centered
view of the same L1000 ablation evidence.  Instead of asking whether each
profile has one favorable metric in isolation, it asks which profiles move the
system beyond the minimal discovery regime along three core axes: held-out
connectivity effect, the strong-positive response tail, and train--holdout
stability.

\begin{table}[!htb]
\caption{Minimal-baseline-centered L1000 effect--stability gains.
Relative changes use the minimal baseline as the reference.  Higher values are
better in all three columns; for the gap column, a positive value means that
the train--holdout gap is reduced.  The table is descriptive and does not
introduce a composite score.}
\label{tab:s4-minimal-centered-profile}
\centering
\begingroup
\small
\setlength{\tabcolsep}{3pt}
\begin{tabularx}{\linewidth}{@{}lccc>{\raggedright\arraybackslash}X@{}}
\toprule
Profile & Effect lift \(\uparrow\) & Strong-tail lift \(\uparrow\) & Gap reduction \(\uparrow\) & Read-out \\
\midrule
Minimal & 0.0\% & 0.0\% & 0.0\% & Reference discovery regime. \\
w/o dynamic & +3.0\% & +3.3\% & +38.7\% & Small effect and tail gains; gap decreases. \\
w/o virtual & +2.4\% & -12.3\% & +21.6\% & Small effect lift with weaker strong-response tail. \\
w/o Pareto & +10.8\% & -31.9\% & +27.0\% & Effect lift without strong-tail concentration. \\
w/o upper & +6.6\% & +27.2\% & +4.5\% & Below SCENE (ours); limited gap reduction. \\
SCENE (ours) & \textbf{+13.3\%} & \textbf{+50.4\%} & \textbf{+43.2\%} & Largest balanced gains beyond minimal. \\
\bottomrule
\end{tabularx}
\endgroup
\end{table}

Table~\ref{tab:s4-minimal-centered-profile} clarifies the contribution of
upper-level planning in the L1000 setting.  Without the upper layer, the system
remains below SCENE (ours) and provides only a small train--holdout gap reduction,
suggesting that lower-level search alone is less stable across the holdout
boundary.  SCENE (ours) exhibits a different profile: it achieves the largest
effect lift, the strongest high-response tail, and the largest reduction
in the train--holdout gap relative to the minimal baseline.  These coupled
gains indicate that upper-level planning helps coordinate contextual proposal
generation, search, and validation so that the exported propositions preserve target-response signal across the holdout boundary.

\subsection{Evidence-Gated Lifecycle of Generated Features}
\label{app:s5-feature-lifecycle}

\begin{figure}[!htb]
    \centering
    \includegraphics[width=0.92\linewidth]{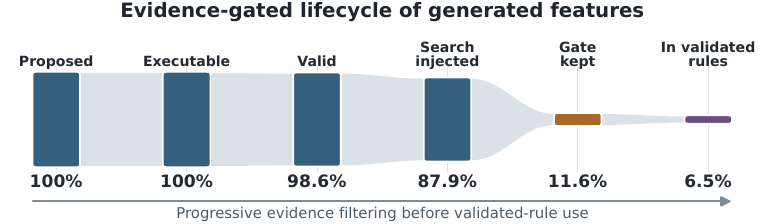}
    \caption{
    Evidence-gated lifecycle of SCENE-generated features. Percentages are measured relative to all proposed generated run-features. Generated features are first required to be executable and non-degenerate, then injected into search, filtered by an evidence gate, and finally used only if they appear in validated rules. The large drop after search injection indicates that SCENE treats generated features as candidate contextualizations rather than automatically reportable discoveries.
    }
    \label{fig:feature-lifecycle}
\end{figure}

SCENE allows the discovery process to introduce scenario-specific generated features, such as longitudinal summaries or perturbational signature summaries. This flexibility is useful for knowledge contextualization, but it also requires a strict execution boundary: generated features should not be treated as free-form model suggestions. They must be materialized from registered scenario tables, pass deterministic validity checks, enter the search only as executable feature objects, and survive evidence-based filtering before they can appear in reported rules.

Figure~\ref{fig:feature-lifecycle} audits this lifecycle over the completed discovery records. The denominator is the set of proposed generated run-features. Nearly all proposals are executable and non-degenerate after deterministic construction, and most are injected into the search frontier. The sharp reduction occurs at the evidence gate: only a small subset is retained after the search evaluates whether the generated feature provides useful frontier evidence. An even smaller subset appears in the final validated rules. This pattern is desirable for SCENE. It shows that the framework uses generated features to expand the grounding space, but does not allow this expanded space to directly determine reported propositions. Instead, generated features must first become valid scenario objects and then earn their place through the same evidence-gated search process as ordinary literals.

This audit supports the main claim that SCENE performs controlled knowledge contextualization rather than unconstrained feature invention. Broad biomedical cues can motivate new scenario-specific summaries, but the executable boundary is governed by the schema, construction grammar, and evidence gate. The final reported rules therefore reflect a conservative subset of generated features: features that are materialized, validated, searched, and retained with supporting evidence. The figure should be read as a protocol audit rather than an additional model-selection step; all quantities are computed from completed runs and are not used to revise the discovery process.

\subsection{Grounding Trace from Prior Knowledge to Executable Propositions}
\label{app:s6-grounding-trace}

\begin{figure}[!htb]
    \centering
    \includegraphics[width=\linewidth]{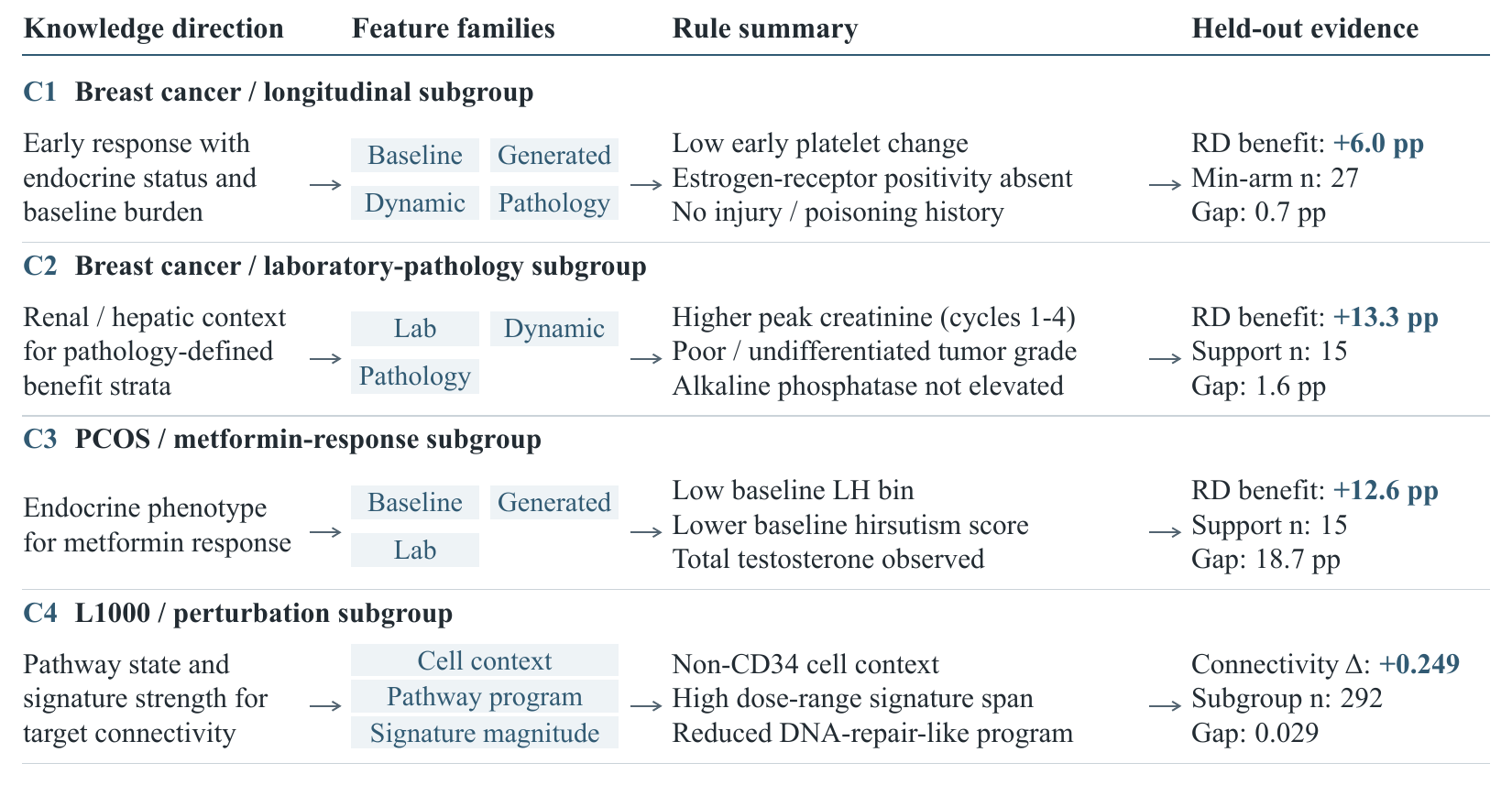}
    \caption{Illustrative grounding paths. Each row links a concise biomedical direction, feature families, a rule summary (all conditions apply), and archived validation evidence. Clinical risk-difference (RD) benefit is in percentage points (pp); L1000 reports a connectivity delta. Gap is the absolute training/discovery--validation effect difference. All four cases preserve effect direction. These are selected examples, not an additional model-selection procedure.}
    \label{fig:grounding-trace-cards}
\end{figure}

\begin{samepage}
The main quantitative results evaluate whether SCENE discovers useful rules or findings, but they do not by themselves show how a broad biomedical direction becomes a concrete scenario-level proposition. Figure~\ref{fig:grounding-trace-cards} illustrates this conversion through four selected examples. Each row connects a semantic direction, grounded feature families, a readable rule description, and the associated validation record. Trajectory examples use predeclared landmarked or trajectory-enhanced features and should be read as predictive subgroup hypotheses, not baseline-only treatment-decision rules.
\par\end{samepage}

The four examples were selected to be illustrative but evidence-bearing rather than purely anecdotal. The first three examples come from clinical-trial subgroup discovery and show breast-cancer and PCOS examples in which baseline descriptors, laboratory measurements, longitudinal summaries, generated feature bins, pathology descriptors, or endocrine markers are combined into executable subgroup rules. The last example comes from the L1000 perturbational setting and shows the same grounding idea in a different schema: a target-connectivity direction is instantiated through cell context, pathway-program state, and transcriptional signature magnitude. In all four cases, the output is not just a natural-language explanation. Each displayed rule description corresponds to an archived predicate over scenario-visible features, and the rightmost column reports post-discovery evidence on held-out or validation data.

This organization clarifies SCENE's knowledge-contextualization process. A static knowledge-injection approach can state that a biomedical concept is relevant, and a data-only rule miner can return a high-scoring rule, but neither necessarily records how the concept is grounded into the current schema. The figure illustrates the division of work in SCENE: a direction proposes the semantic intent, grounding maps it to feature families, and search turns those families into a concrete rule. The clinical examples report positive archived validation estimates of risk-difference benefit, while the L1000 example shows a positive connectivity delta for a context-specific perturbational subgroup. These examples therefore make the reported propositions inspectable.

The figure is not intended as an aggregate performance comparison and should not be interpreted as independent biomedical validation of the displayed mechanisms. Rather, it is a qualitative audit of the reporting boundary. It demonstrates that SCENE exports propositions with a visible grounding path, rather than reporting isolated rule strings or post-hoc textual rationales.

\paragraph{Fixed-input planner calls.}
On P-M-TR, seed 58, we compare the earliest three successful API calls in chronological order, with identical source data, split, LLM, temperature zero, and no fallback. All calls retain an insulin-resistance component, while the accompanying direction differs for Call 1 (Table~\ref{tab:planner-consistency}). Calls 2 and 3 produce identical structured groundings. Across the three call pairs, grounded-feature Jaccard averages 0.810 and ranges from 0.714 to 1.000. This small check distinguishes shared semantic focus from exact output determinism. Conditioning on successful calls does not measure API failure rates. The 0.097 split-level ARR standard deviation in Figure~\ref{fig:s1b-mean-sd-profile} describes the entire discovery pipeline across patient splits, not planner variance alone.

\paragraph{Exploratory report quality.}

We freeze one report for each method and clinical frame for SCENE and the baselines across six clinical frames (48 reports per evaluator). Two AI evaluators and one human expert rate all reports on the four dimensions in Table~\ref{tab:report-quality-audit}. Each cell averages six reports. Best non-SCENE selects the highest baseline mean separately for each evaluator and dimension.

\begin{table}[!htb]
\centering
\begin{minipage}[t]{0.49\linewidth}
\vspace{0pt}
\captionof{table}{Planner consistency.}
\label{tab:planner-consistency}
\centering
\small
\setlength{\tabcolsep}{0pt}
\renewcommand{\arraystretch}{1.15}
\begin{tabular*}{\linewidth}{@{\extracolsep{\fill}}lccc@{}}
\toprule
 & \multicolumn{2}{c}{Grounding} & \\
\cmidrule(lr){2-3}
Call & Direction & Feat. & $J$ \\
\midrule
1 & MR + IR & 12 & 0.714 \\
2 & HA + IR & 12 & 0.857 \\
3 & HA + IR & 12 & 0.857 \\
\bottomrule
\end{tabular*}
\end{minipage}\hfill
\begin{minipage}[t]{0.49\linewidth}
\vspace{0pt}
\captionof{table}{Report-quality ratings.}
\label{tab:report-quality-audit}
\centering
\small
\setlength{\tabcolsep}{0pt}
\renewcommand{\arraystretch}{1.15}
\begin{tabular*}{\linewidth}{@{\extracolsep{\fill}}lrrrrrrrr@{}}
\toprule
 & \multicolumn{4}{c}{SCENE} & \multicolumn{4}{c}{Best non-SCENE} \\
\cmidrule(lr){2-5}\cmidrule(l){6-9}
Rater & C & E & A & F & C & E & A & F \\
\midrule
AI-1 & 4.33 & 4.67 & 5.00 & 4.83 & 3.83 & 2.83 & 3.83 & 3.67 \\
AI-2 & 4.33 & 4.83 & 5.00 & 4.83 & 3.67 & 2.83 & 4.00 & 3.33 \\
Human & 4.50 & 5.00 & 5.00 & 5.00 & 4.00 & 3.33 & 4.17 & 3.67 \\
\bottomrule
\end{tabular*}
\end{minipage}
\par\smallskip
\begin{minipage}{\linewidth}
\small
MR, HA, and IR denote metabolic reserve, hyperandrogenism, and insulin-resistance magnitude. Feat. counts grounded features, and $J$ is the mean feature-set Jaccard with the other two calls. Rating dimensions are clinical/biological coherence (C), evidence sufficiency and stability (E), auditability and replayability (A), and readiness (F).
\end{minipage}
\end{table}

SCENE receives higher mean ratings on all four dimensions from each evaluator. This is exploratory evidence about the presentation and inspectability of the reports. A single human evaluator, without uncertainty estimates or a prospective workflow comparison, cannot establish general expert preference, biological validity, or readiness for clinical use. The audit does not compare SCENE with hypotheses generated by an expert-led discovery workflow.

\subsection{Knowledge-Guided Controls and Prior Sensitivity}
\label{app:additional-controls}
\label{app:prior-planner-sensitivity}

We test whether initial knowledge guidance or a single proposal pass can account for SCENE's gains, and whether explicit prior records contribute to them. The controls below complement Tables~\ref{tab:clinical_best1_benefit_revised} and~\ref{tab:l1000_ablation} with full-precision results and paired uncertainty. The preceding replay and component analyses retain their original comparison sets.

\paragraph{Paired uncertainty.}
The intervals below are 95\% percentile-bootstrap intervals obtained by resampling matched run units and recomputing the mean between-method difference. For the two MaxTE contrasts in Table~\ref{tab:paired-clinical-controls}, we pair identical patient split identifiers within each frame (50 splits). The L1000 comparison pairs episodes by target and seed (three targets, ten seeds each). These intervals describe variation under the fixed benchmark protocol. Repeated clinical splits share patients, and L1000 seeds reuse the same three targets.

\paragraph{Initial grounding for clinical rule search.}
As a targeted control for static initial knowledge guidance, \emph{MaxTE + Grounding} receives only the raw-feature set in SCENE's validated generation-0 grounding plan. Relative to unrestricted MaxTE, it keeps the patient split, endpoint, rule-length limit, MaxTE settings, and held-out evaluator fixed. It receives no SCENE thresholds, generated features, later-round feedback, or held-out outcomes. Table~\ref{tab:paired-clinical-controls} gives the full reported precision and both paired contrasts. Grounding alone improves MaxTE's point estimate on three frames and lowers it on three. SCENE has the highest point estimate on all six. The intervals for SCENE minus Grounded MaxTE exclude zero on the four PCOS frames and overlap zero on the two breast-cancer frames. This compares static grounding with the full pipeline.

\begin{table}[!htb]
\caption{Clinical control means and paired 95\% intervals for held-out bad-outcome ARR. S, M, and G denote SCENE, unrestricted MaxTE, and MaxTE + Grounding. Differences are in ARR units.}
\label{tab:paired-clinical-controls}
\centering
\begingroup
\small
\setlength{\tabcolsep}{0pt}
\renewcommand{\arraystretch}{1.15}
\begin{tabular*}{\linewidth}{@{\extracolsep{\fill}}lrrrcc@{}}
\toprule
Frame & SCENE & MaxTE & Grounded & $S-M$: 95\% CI & $S-G$: 95\% CI \\
\midrule
B-D-BL  & 0.0860 & 0.0760 & 0.0015 & $[-0.0385,\,0.0985]$ & $[-0.0193,\,0.1357]$ \\
B-D-TR  & 0.0950 & 0.0120 & 0.0265 & $[0.0008,\,0.2008]$ & $[-0.0079,\,0.1806]$ \\
P-M-BL  & 0.4410 & 0.2340 & 0.1704 & $[-0.0083,\,0.3210]$ & $[0.1532,\,0.3201]$ \\
P-M-TR  & 0.4390 & 0.1210 & 0.1721 & $[0.1360,\,0.4621]$ & $[0.1315,\,0.3130]$ \\
P-A-FPG & 0.2690 & 0.1170 & 0.1210 & $[0.0783,\,0.2347]$ & $[0.0648,\,0.1994]$ \\
P-A-AUC & 0.2700 & 0.1320 & 0.1009 & $[0.0678,\,0.2543]$ & $[0.0933,\,0.2040]$ \\
\bottomrule
\end{tabular*}
\endgroup
\end{table}

Grounded MaxTE has P-ARR 0.03506, U-Supp. 50.67\%, and Best-1 direction consistency 66.33\%, versus 0.093, 84.7\%, and 100.0\% for SCENE. These diagnostics follow the metric definitions and Best-1 selection and held-out replay protocol used in Table~\ref{tab:clinical_best1_benefit_revised}. Paired intervals are provided for the two MaxTE contrasts, not for every original baseline.

\paragraph{Source-documented prior records.}
We remove only the explicit source-documented prior records while keeping the schema, task context, LLM, search budget, split, and evaluator fixed. Table~\ref{tab:source-prior-sensitivity} reports lower ARR on the two tested frames, with positive paired intervals for Full minus No Sources. This supports a contribution from the supplied records in these frames. The control retains the LLM's parametric knowledge and does not measure sensitivity to all possible priors or across the full benchmark.

\begin{table}[!htb]
\caption{Targeted removal of source-documented prior records. No Sources removes the explicit records only. $\Delta$ is Full minus No Sources in held-out bad-outcome ARR.}
\label{tab:source-prior-sensitivity}
\centering
\begingroup
\small
\setlength{\tabcolsep}{0pt}
\renewcommand{\arraystretch}{1.15}
\begin{tabular*}{\linewidth}{@{\extracolsep{\fill}}lrrrc@{}}
\toprule
Frame & Full & No Sources & $\Delta$ & 95\% paired CI \\
\midrule
B-D-BL & 0.0860 & 0.0787 & 0.0073 & $[0.0008,\,0.0217]$ \\
P-M-TR & 0.4390 & 0.3637 & 0.0753 & $[0.0432,\,0.1521]$ \\
\bottomrule
\end{tabular*}
\endgroup
\end{table}

\paragraph{One-shot L1000 proposals.}
The One-shot LLM Proposal control uses the same source priors, schema, initial candidate budget, split, output type, and Top-$K$ evaluator as SCENE, on RPS6, MYC, and AURKB with ten paired seeds per target. It performs one proposal/grounding pass, validates and ranks the candidates using training-side evidence, then freezes and replays them. It has no later evolution or evidence feedback. The initial proposal budget is matched, but total adaptive computation is not.

\begin{table}[!htb]
\caption{One-shot L1000 control and paired 95\% intervals. The contrast is SCENE minus One-shot except for Gap, where it is One-shot minus SCENE. Positive differences favor SCENE. Intervals for percentage metrics are in percentage points (pp).}
\label{tab:paired-l1000-control}
\centering
\begingroup
\small
\setlength{\tabcolsep}{0pt}
\renewcommand{\arraystretch}{1.15}
\begin{tabular*}{\linewidth}{@{\extracolsep{\fill}}lrrl@{}}
\toprule
Metric & SCENE & One-shot & 95\% CI \\
\midrule
Conn. $\uparrow$ & 0.188 & 0.107 & $[0.0173,\,0.1364]$ \\
Strong-Pos. $\uparrow$ & 41.5\% & 24.42\% & $[3.20,\,25.01]$ pp \\
AUPRC $\uparrow$ & 0.268 & 0.2532 & $[-0.0209,\,0.0181]$ \\
U-Supp. $\uparrow$ & 31.4\% & 38.31\% & $[-10.1211,\,8.3101]$ pp \\
Gap $\downarrow$ & 0.063 & 0.129 & $[0.0042,\,0.2179]$ \\
\bottomrule
\end{tabular*}
\endgroup
\end{table}

Table~\ref{tab:paired-l1000-control} supports gains in connectivity, strong-response concentration, and stability from training to holdout under this protocol. The AUPRC and U-Supp. intervals include zero. The one-shot control has the higher U-Supp. point estimate. These results do not establish superiority on every metric or replace a comparison with specialized perturbational retrieval methods.

\section{Limitations}
\label{app:limitations}

SCENE is designed for biomedical hypothesis generation. The clinical analyses identify scenario-grounded subgroups with held-out outcome contrasts, but these propositions are not diagnostic criteria, treatment recommendations, or patient-level decision rules without independent validation. Live language-model calls can propose different directions or grounding cues across providers, model versions, and decoding settings. Similarly, L1000 replay provides perturbational assay evidence for prioritizing follow-up, not wet-lab mechanism confirmation or drug-efficacy evidence. The experiments cover the clinical task frames and L1000 target programs studied in this paper; applications to new endpoints, cohorts, assays, or disease areas should be accompanied by task-specific validation and domain review. These results should therefore be interpreted as evidence that broad biomedical knowledge can be contextualized into auditable propositions, rather than as a substitute for prospective clinical or experimental studies.

\section{Broader Impacts}
\label{app:broader-impacts}

SCENE may improve biomedical discovery workflows by making hypothesis generation more transparent: outputs are expressed as executable rules with source-traceable search directions, support/effect summaries, and replay diagnostics that can be inspected by domain experts. This can help researchers organize large clinical or perturbational datasets and prioritize candidates for follow-up analysis. More broadly, SCENE points toward auditable auto-research workflows in which
agents propose executable hypotheses rather than ungrounded claims;
however, such automation can also create automation bias if concise rules are
trusted without expert review. The main risk is overinterpretation of concise rules as actionable medical or therapeutic claims. Responsible use therefore requires expert review, independent validation before downstream decisions, and appropriate governance for data privacy, security, and fairness when patient data are involved. Clear reporting of validation scope and uncertainty diagnostics is important to keep the resulting propositions in their intended exploratory role. Finally, multi-agent discovery has computational and operational costs. Live LLM calls, evolutionary search, and repeated validation can increase compute use and complicate reproducibility. The implementation choices in this paper emphasize typed role outputs, deterministic validation, and reusable manifests so that future users can limit unnecessary calls, replay completed runs, and audit the provenance of reported propositions.


\end{document}